\documentclass[10pt,twocolumn,letterpaper]{article}

\usepackage{wacv}
\usepackage{time}
\usepackage{epsfig}
\usepackage{adjustbox}
\usepackage{graphicx}
\usepackage{amsmath}
\usepackage{amssymb}
\usepackage{subcaption}

\usepackage{cuted}
\usepackage{dsfont}
\usepackage{multirow}
\usepackage{array}
\usepackage[accsupp]{axessibility}
\usepackage{appendix}
\newcommand{\PreserveBackslash}[1]{\let\temp=\\#1\let\\=\temp}
\newcommand*\ruleline[2]{\par\noindent\raisebox{.8ex}{\makebox[{#1}]{\hrulefill\hspace{1ex}\raisebox{-.8ex}{#2}\hspace{1ex}\hrulefill}}}
\newcolumntype{C}[1]{>{\PreserveBackslash\centering}p{#1}}
\newcolumntype{R}[1]{>{\PreserveBackslash\raggedleft}p{#1}}
\newcolumntype{L}[1]{>{\PreserveBackslash\raggedright}p{#1}}

\DeclareMathOperator*{\argmin}{arg\,min}

\def\wacvPaperID{346} 

\wacvfinalcopy

\ifwacvfinal
\def\assignedStartPage{1} 
\fi
\ifwacvfinal
\usepackage[breaklinks=true,bookmarks=false]{hyperref}
\else
\usepackage[pagebackref=true,breaklinks=true,colorlinks,bookmarks=false]{hyperref}
\fi

\ifwacvfinal
\else
\fi

\begin{document}

\title{StyleMC: Multi-Channel Based Fast Text-Guided Image Generation and Manipulation}

\author{Umut Kocasarı\quad Alara Dirik\quad Mert Tiftikci\quad Pinar Yanardag\\
Boğaziçi University\\
Istanbul, Turkey\\
{\tt\small \{umut.kocasari, alara.dirik, mert.tiftikci\}@boun.edu.tr, yanardag.pinar@gmail.com}
}

\maketitle

\begin{strip}\centering
 \begin{minipage}{.45\textwidth}
        \centering
        \includegraphics[width=\textwidth]{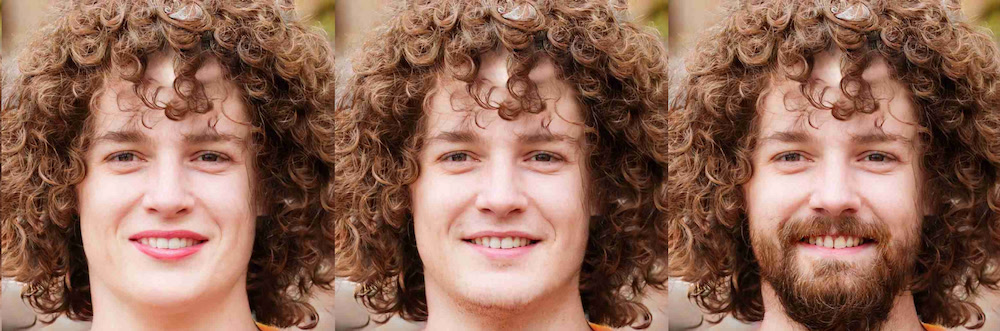}
        \vspace{-0.75cm}\captionof*{figure}{`Beard' on FFHQ. }
\end{minipage}
\begin{minipage}{0.45\textwidth}
        \centering
        \includegraphics[width=\textwidth]{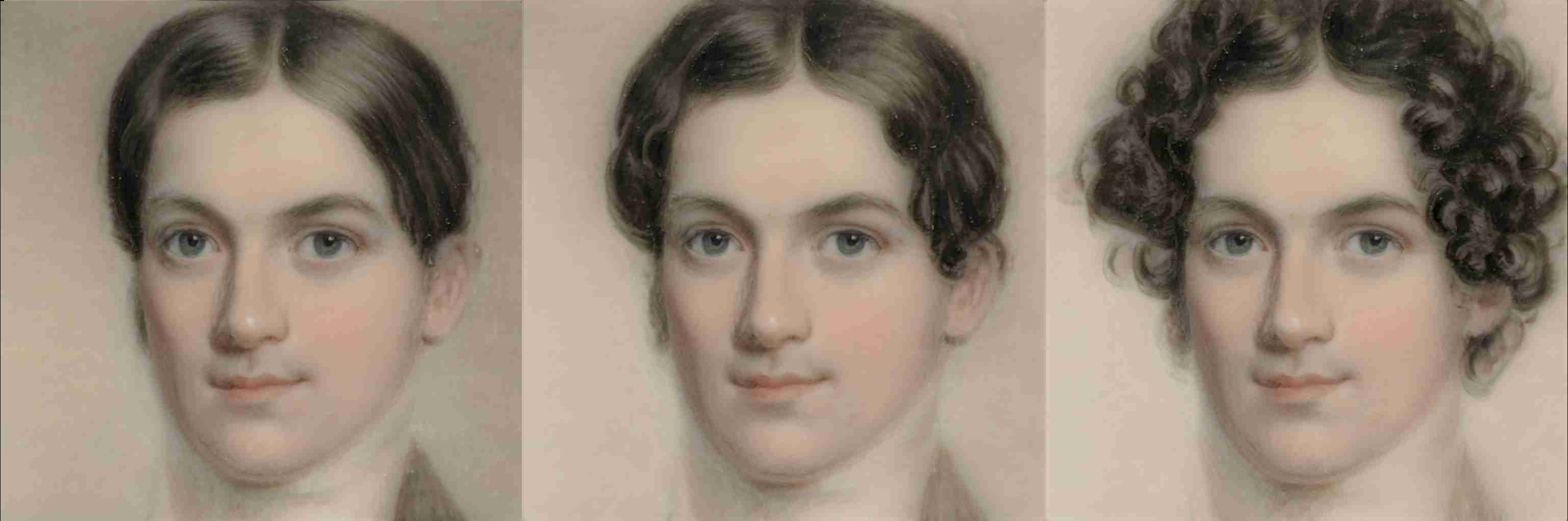}
        \vspace{-0.75cm}\captionof*{figure}{`Curly hair' on MetFaces.}
\end{minipage}
\begin{minipage}{.45\textwidth}
        \centering
        \includegraphics[width=\textwidth]{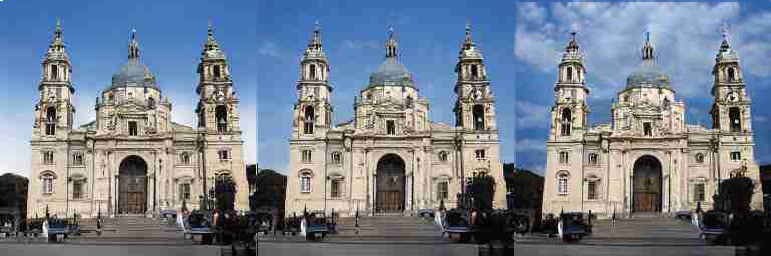}
        \vspace{-0.75cm}\captionof*{figure}{`Cloud' on LSUN Church.}
\end{minipage}
\begin{minipage}{0.45\textwidth}
        \centering
        \includegraphics[width=\textwidth]{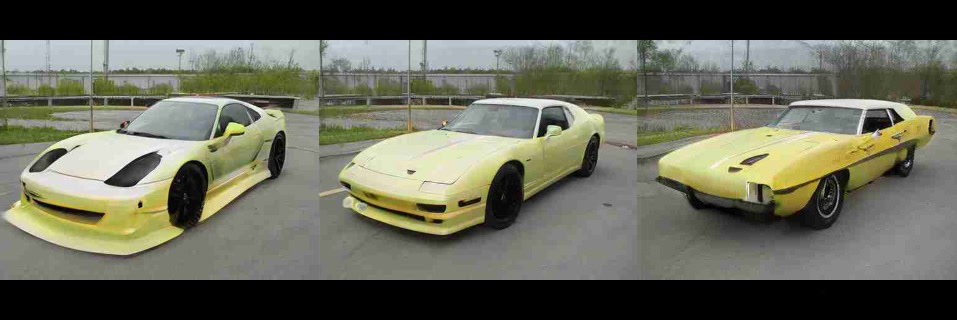}
        \vspace{-0.75cm}\captionof*{figure}{`Classic car' on LSUN Car.}
\end{minipage}
\begin{minipage}{.45\textwidth}
        \centering
        \includegraphics[width=\textwidth]{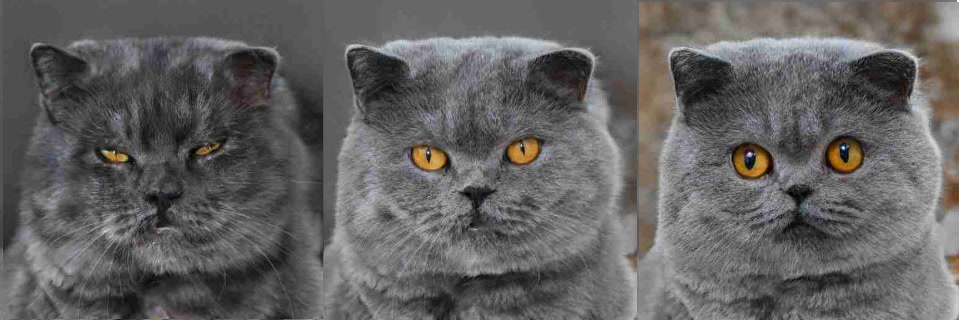}
        \vspace{-0.75cm}\captionof*{figure}{`Cute cat' on AFHQ Cat.}
\end{minipage}
    \begin{minipage}{0.45\textwidth}
        \centering
        \includegraphics[width=\textwidth]{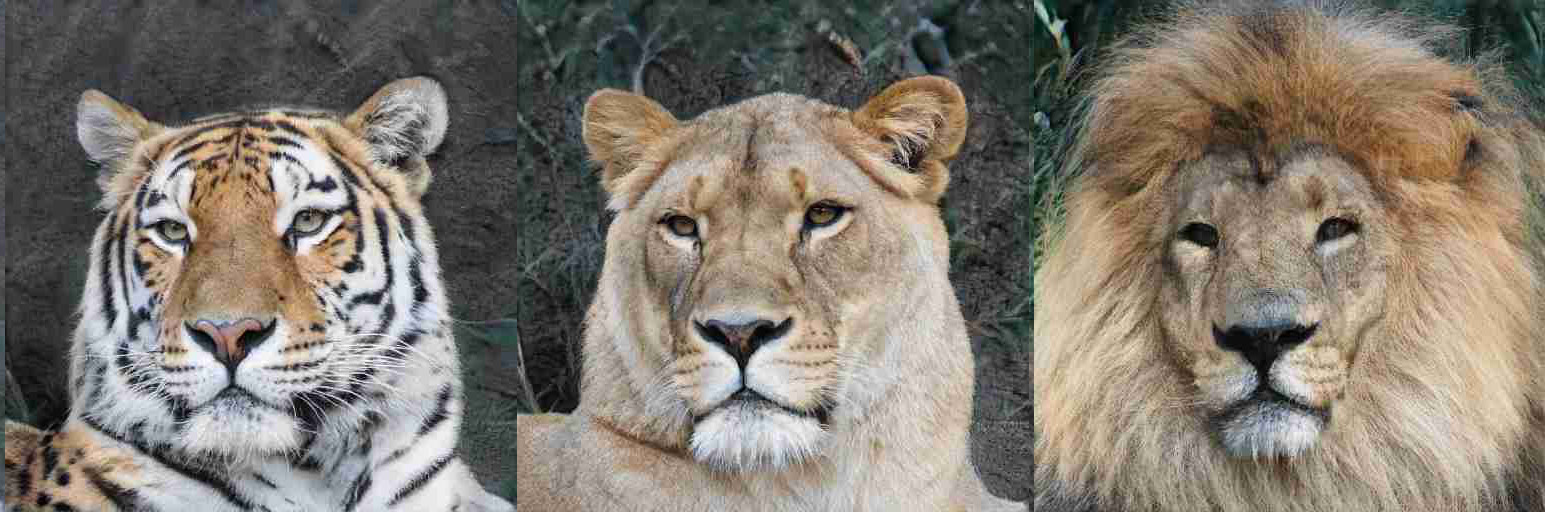}
        \vspace{-0.75cm}\captionof*{figure}{`Wild Lion' on AFHQ Wild.}
\end{minipage}
\vspace*{-0.3cm}
\captionof{figure}{Examples of text-driven manipulations using our method. For each example, the middle column shows the input images, left and right columns show manipulated results toward negative and positive directions, respectively.
\label{fig:teaser}}
\end{strip}

\vspace*{-0.8cm}

\begin{abstract}
Discovering meaningful directions in the latent space of GANs to manipulate semantic attributes typically requires large amounts of labeled data. Recent work aims to overcome this limitation by leveraging the power of Contrastive Language-Image Pre-training (CLIP), a joint text-image model. While promising, these methods require several hours of preprocessing or training to achieve the desired manipulations. In this paper, we present StyleMC, a fast and efficient method for text-driven image generation and manipulation. StyleMC uses a CLIP-based loss and an identity loss to manipulate images via a single text prompt without significantly affecting other attributes. Unlike prior work, StyleMC requires only a few seconds of training per text prompt to find stable global directions, does not require prompt engineering and can be used with any pre-trained StyleGAN2 model. We demonstrate the effectiveness of our method and compare it to state-of-the-art methods. Our code can be found at \url{http://catlab-team.github.io/stylemc}.
\end{abstract} 

\section{Introduction}
Generative Adversarial Networks (GANs) \cite{NIPS2014_5423} have revolutionized generative modeling in computer vision since their emergence. Due to their powerful image synthesis capabilities, they are widely used for various visual tasks including image generation \cite{DBLP:journals/corr/abs-1710-10916}, image manipulation \cite{wang2017highresolution}, super-resolution \cite{Sun_2020}, and domain translation \cite{CycleGAN}. 
 
Despite their success, how to control the results generated by GANs remains an active research question. Previous work on controlled generation has shown that it is possible to generate images that belong to certain categories or have certain attributes by training conditional models \cite{mirza2014conditional}. However, conditional GANs require large amounts of labeled data for each target attribute. InfoGAN \cite{chen2016infogan} is an approach that aims to develop models that generate a disentangled latent space in which each latent dimension controls a particular attribute. However, these approaches provide only limited control, depending on the granularity of available supervised information.

Recent research addressing these problems and aiming at controllable generation includes simple methods such as modifying the latent codes of images \cite{radford2015unsupervised} and more complex methods such as searching for directions and interpolating latent vectors within pre-trained GAN models such as StyleGAN \cite{StyleGAN}. Another branch of work aims at finding disentangled directions in the latent space of GANs in a more principled way. Most of this work discovers domain-independent and interpretable directions such as \textit{zoom-in, rotation}, and \textit{translation} \cite{plumerault2020controlling,harkonen2020ganspace,voynov2020unsupervised}, while other frameworks propose to find a set of domain-specific directions such as \textit{hair color} or \textit{gender} on face images \cite{shen2020interfacegan} or cognitive features \cite{goetschalckx2019ganalyze}. The directions found are then used to modify a generated image by controlling the latent code by a certain amount to enhance or negate the target attribute in the generated image. Other work uses \textit{style space} of StyleGAN2  to discover disentangled attributes and manipulate images for both coarse (e.g., gender, identity) and fine (e.g., hairstyle, eyes) visual features \cite{wu2020stylespace}. Recent work has also shown that these image manipulation methods can also be applied to real images by finding a latent code that accurately reconstructs the input image \cite{Richardson2020EncodingIS}. The latent code of the inverted real image can then be fed as input to GAN to perform processing operations directly on real images \cite{abdal2019image2stylegan}.

Recent works such as StyleCLIP and Paint by Word \cite{bau2021paint} uses CLIP \cite{Radford2021LearningTV}  to manipulate real-world images via user-specified text prompts. However, both methods require hours of preprocessing or training to find stable directions. In this work, we propose a method to find image-independent manipulation directions in the latent space of pre-trained StyleGAN2 models using user-specified text prompts.  Our method takes advantage of the joint representational power of CLIP and the generative power of StyleGAN2 while benefiting from the following key observations:

\begin{itemize}
\item We use style space of StyleGAN2 which is shown to be its most disentangled latent space \cite{wu2020stylespace}. Using the style space, we find multiple style channels to compute a global direction that can perform complex manipulations.
\item Unlike previous work, our method finds directions using only layers up to $256\times256$ resolution within StyleGAN2, providing a significant speedup. We then use the found directions to apply manipulations and generate images at high resolutions such as $1024\times1024$.
\item Our method uses only 128 randomly generated images to find stable and global manipulation directions regardless of the given text prompt. 
\end{itemize}

Unlike previous work, such as StyleCLIP, our approach requires only a few seconds of training to find stable directions and does not require prompt engineering. In addition, our method is input agnostic, and can be applied to inverted real images as well as randomly generated images. We demonstrate the manipulation capabilities and efficiency of our method on a variety of datasets. Our results show that the discovered directions can successfully perform the desired processing while operating significantly faster.

The rest of this paper is organized as follows. Section \ref{sec:related_work} discusses related work on latent space manipulation. Section \ref{sec:methodology} presents our framework and Section \ref{sec:experiments} discusses quantitative and qualitative results. Section \ref{sec:limitations} discusses the limitations and implications of our work and Section \ref{sec:conclusion} concludes the paper.

\section{Related Work}
\label{sec:related_work}
\subsection{Generative Adversarial Networks} 

Generative Adversarial Networks (GANs) are two-part networks consisting of a generator and a discriminator \cite{NIPS2014_5423} trained simultaneously in an adversarial manner.  StyleGAN \cite{StyleGAN} and StyleGAN2 are among the popular GAN approaches that generate high-quality images. They map the input latent code to an intermediate latent space using a mapping network. BigGAN \cite{BigGAN} is another large-scale model that uses \textit{skip-z} connections, as well as a class vector. In this work, we work with pre-trained StyleGAN2 models.

\subsection{Latent Space Manipulation}

Several methods have been proposed to exploit the latent space of GANs for image manipulation, which can be divided into two broad categories: supervised and unsupervised methods. Supervised approaches typically benefit from pre-trained attribute classifiers that guide the optimization process to discover meaningful directions in the latent space, or use labeled data to train new classifiers that directly aim to learn directions of interest  \cite{goetschalckx2019ganalyze,shen2020interfacegan}.  Other work shows that it is possible to find meaningful directions in latent space in an unsupervised way \cite{voynov2020unsupervised,jahanian2019steerability,upchurch2017deep}. GANSpace \cite{harkonen2020ganspace} proposes to apply Principal Component Analysis (PCA) \cite{wold1987principal} to randomly sampled latent vectors of the intermediate layers of BigGAN and StyleGAN models.  A similar approach is used in SeFA \cite{shen2020closed}, where they directly optimize the intermediate weight matrix of the GAN model in closed form.  LatentCLR \cite{yuksel2021latentclr} proposes a contrastive learning approach to find unsupervised directions that are transferable to different classes.

\subsection{Text-Based Image Manipulation}
Text-based image manipulation methods aim to generate images that contain visual attributes corresponding to the given text input without changing irrelevant attributes \cite{Dong2017SemanticIS,Zhang2017StackGANTT, Nam2018TextAdaptiveGA,Li2020ManiGANTI,li2020lightweight}. One of the recent works that uses image-text matching methods is TediGAN \cite{Xia2020TediGANTD}, which inverts real images using the inversion module of StyleGAN and then learns the correspondences between visual and linguistic attributes.  

Other recent work uses image-text matching methods such as CLIP, to harness the power of joint image-text representations. CLIP is a multimodal contrastive learning framework with two encoder modules that aim to map the image and text pairs to the same embedding space. To achieve this, it maximizes the similarity between the embeddings of the matched image and text instances while minimizing that of the unmatched instance, resulting in a powerful bidirectional mapper. Recent works such as StyleCLIP and Paint by Word use CLIP to provide feedback to the generated images. StyleCLIP provides three different methods, namely \textit{Latent Optimization, Latent Mapper} and \textit{Global Directions}, which we refer to as StyleCLIP-LO, StyleCLIP-LM and StyleCLIP-GD for the rest of this paper. StyleCLIP-LO directly optimizes the latent code given an image and a text prompt. StyleCLIP-LM uses a latent residual mapper trained on a particular text prompt. StyleCLIP-GD maps a text prompt in an input-independent global direction. Paint by Word uses user-specified masks to perform manipulations within a specified region. However, both approaches require hours of preprocessing and training time. Our method aims to overcome this limitation by proposing a more efficient way to find stable and accurate directions to introduce or emphasize desired attributes in images.

\section{Methodology}
\label{sec:methodology}

\subsection{Background on Style Space}
\label{sec:stylegan}

The generation process of StyleGAN2 consists of several latent spaces, namely $\mathcal{Z}$, $\mathcal{W}$, $\mathcal{W+}$ and $\mathcal{S}$. More formally, let $\mathcal{G}$ denote a generator acting as a mapping function $\mathcal{G}: \mathcal{Z} \to \mathcal{X}$ where $\mathcal{X}$ is the target image domain. The latent code $\mathbf{z} \in \mathcal{Z}$ is drawn from a prior distribution $p(\mathbf{z})$, typically chosen to be Gaussian. The $\mathbf{z}$ vectors are transformed into an intermediate latent space $\mathcal{W}$ using a mapper function consisting of 8 fully connected layers. The latent vectors $\mathbf{w} \in \mathcal{W}$ are then transformed into channel-wise style parameters, forming the \textit{style space}, denoted $\mathcal{S}$, which is the latent space that determines the style parameters of the image. The $\mathcal{W+}$ space is an extended version of $\mathcal{W}$ that uses a different intermediate latent vector on each layer of the synthesis network. It has been shown that the $\mathcal{W}+$ space better reflects the disentangled nature of the latent space than the $\mathcal{W}$ space and is therefore more commonly used for image inversion \cite{Tov2021DesigningAE}.

The synthesis network of the generator in StyleGAN2 consists of several blocks, each block having two convolutional layers for synthesizing feature maps. Each main block has an additional $1\times1$ convolutional layer that maps the output feature tensor to RGB colors, referred to as \textit{tRGB}. The three different style code vectors are denoted as $\mathbf{s}_{i1}$, $\mathbf{s}_{i2}$, and $\mathbf{s}_{itRGB}$, where $i$ indicates the block number. Given a block $i$, the style vectors $\mathbf{s}_{i1}$ and $\mathbf{s}_{i2}$ of each block consist of style channels that control disentangled visual attributes. The style vectors of each layer are obtained from the intermediate latent vectors $\mathbf{w} \in \mathcal{W}$ of the same layer by three affine transformations, $\mathbf{w}_{i1} \rightarrow \mathbf{s}_{i1}, \mathbf{w}_{i2} \rightarrow \mathbf{s}_{i2}, \mathbf{w}_{i2} \rightarrow \mathbf{s}_{it R G B}$.

In our work, we use the style space $\mathcal{S}$ to perform manipulations, as it is shown to be the most disentangled, complete and informative space \cite{wu2020stylespace} compared to $\mathcal{W}$ and $\mathcal{W+}$.

\begin{figure}[t!]
\begin{center}
\centerline{\includegraphics[width=1\columnwidth]{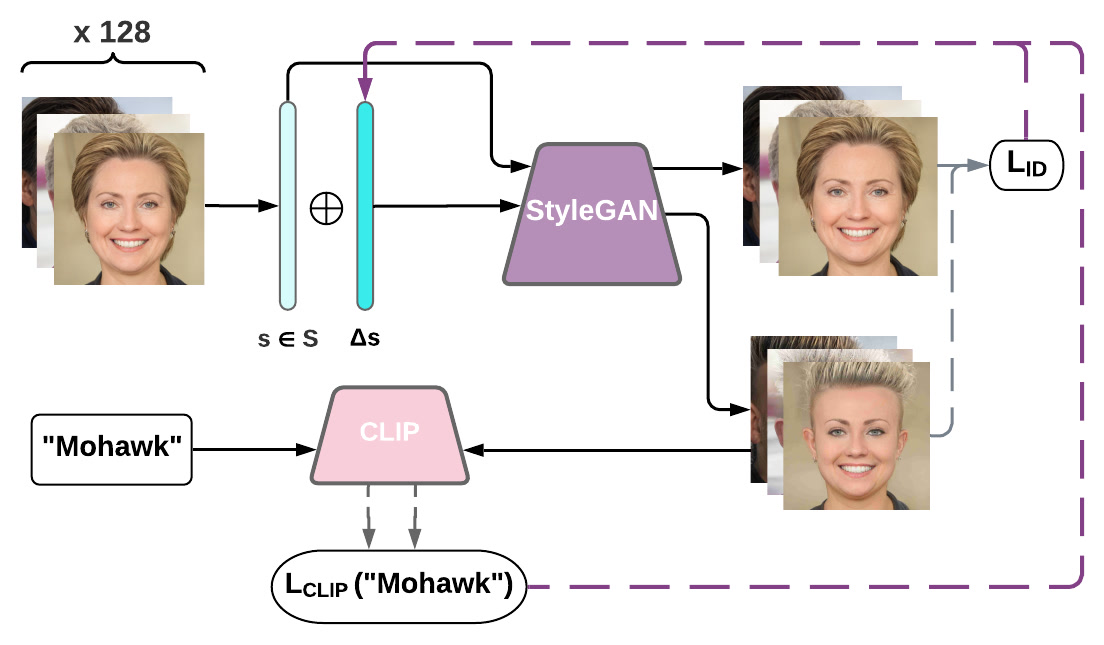}}
\caption{StyleMC framework (using the text prompt \textit{`Mohawk'} as an example).The latent code $\mathbf{s}$ and $\Delta\mathbf{s} +\mathbf{s}$ are passed through the generator.The global manipulation direction $\Delta \mathbf{s}$ corresponding to the text prompt is optimized by minimizing CLIP loss and identity loss.}
\label{fig:framework}
\end{center}
\vskip -0.3in
\end{figure}
\subsection{StyleMC: Style Space based Multi-Channel Directions}
\label{sec:multi-channel}
Given a pre-trained StyleGAN2 generator $\mathcal{G}$ and a style code $\mathbf{s} \in \mathcal{S}$, let $\mathcal{G}(\mathbf{s})$ represent the generated image. Our method takes a text prompt $t$ such as '\textit{A woman wearing makeup}' as input and finds a manipulation direction $\Delta \mathbf{s}$ such that $G(\mathbf{s} + \Delta \mathbf{s})$ generates a manipulated image in which the target attribute specified by $t$ is present or enhanced, while other attributes remain mostly unaffected. A diagram of our method is shown in Figure \ref{fig:framework}.

We use a combination of a CLIP loss and an identity loss, taking advantage of the disentangled nature of the style space $\mathcal{S}$. More specifically, our method attempts to find a global manipulation direction $\Delta s$ that controls the target attribute $t$ by iteratively training over a batch of randomly generated images. The CLIP-based loss term $\mathcal{L}_{ CLIP }$ minimizes the cosine distance between CLIP embeddings of the generated image and the text prompt $t$ as follows:
\begin{equation}
\begin{split}
 \mathcal{L}_{CLIP} = D_{CLIP}(G(s + \Delta s), t) 
\end{split}
\end{equation}

 where $D_{ CLIP }$ is the cosine distance between CLIP embeddings. We also use an identity loss that minimizes the distance between the input images and the generated images:

\begin{equation}
\mathcal{L}_{ID}  = 1- \left \langle R(G(s)), R(G(s + \Delta s)) \right \rangle 
\end{equation}

 where $R$ is an identity network, such as ArcFace \cite{Deng2019ArcFaceAA} in the case of face recognition, and $\langle \cdot, \cdot \rangle$ computes cosine similarity. Identity loss prevents changes to irrelevant attributes\footnote{See the ablation study in Appendix \ref{appdx:identity_loss}.}. The loss of our network is formulated as follows:
 
\begin{equation}
 \argmin_{\mathbf{\Delta \mathbf{s}} \in \mathcal{S}} \lambda_{C} \mathcal{L}_{CLIP} + \lambda_{ID} \mathcal{L}_{ID}
 \label{eq:final}
\end{equation}

where $\lambda_{C}$ and $\lambda_{ ID }$ are the loss coefficients of $\mathcal{L}_{ CLIP }$ and $\mathcal{L}_{ ID }$, respectively. We initialize $\Delta \mathbf{s}$ as a vector of zeros and minimize Eq.\ref{eq:final} to find $\Delta \textbf{s}$. We then apply the found direction $\Delta s$ to the generated images using $\mathcal{G}(\mathbf{s} + \alpha \Delta \mathbf{s})$, where $\alpha$ is a parameter indicating the strength. The directions can be applied either to randomly generated images from StyleGAN2 or to real images inverted by a StyleGAN2 encoder such as e4e \cite{Tov2021DesigningAE}. Note that we do not optimize the latent code as in the StyleCLIP-LO method \cite{patashnik2021styleclip}. Instead, we find a global direction for a given text prompt $t$ that can be applied to any image.
Our method works significantly faster than the other methods by taking advantage of the following insights:

\begin{figure*}[!htp]
\begin{center}
Original  \hspace{0.3cm}  Beard  \hspace{0.25cm}   Long hair \hspace{0.2cm}   Happy  \hspace{0.15cm}   Curly hair\hspace{0.15cm}     Frowning   \hspace{0.05cm}  Blonde hair   \hspace{0.05cm}  Tanned  \hspace{0.15cm}   Relieved   \hspace{0.15cm}  Excited
\includegraphics[width=2\columnwidth]{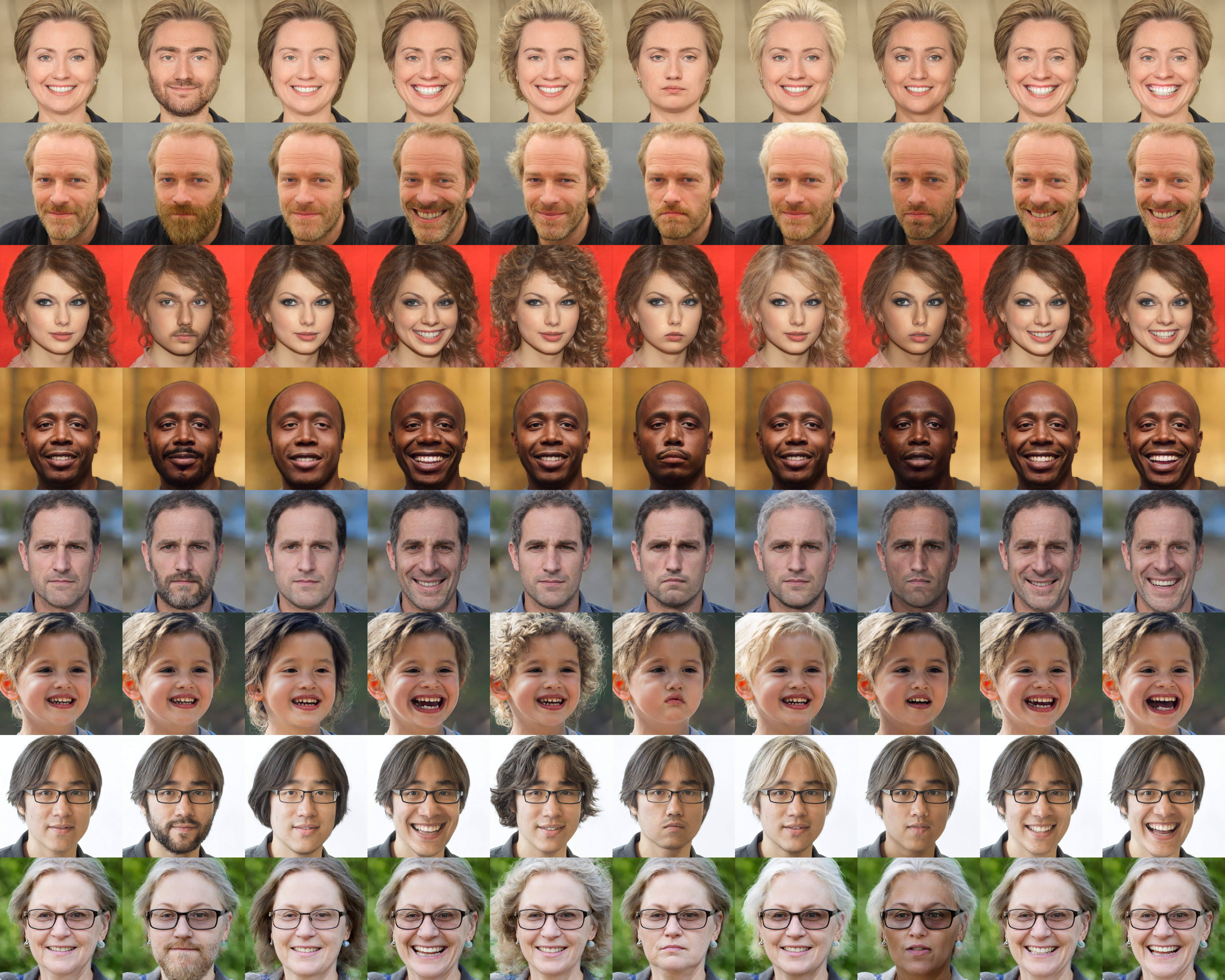}
\caption{A variety of manipulations on StyleGAN2 FFHQ model. Rows 1-4 shows inverted real images and Rows 5-8 shows randomly generated images. The text prompt used for the manipulation is above each column.}
\label{fig:final}
\end{center}
\vskip -0.2in
\end{figure*}

\begin{itemize} 
\item \textbf{Low-resolution layers.} The training process of the StyleGAN2 model starts by focusing on low-resolution features and progressively focuses on higher resolutions, shifting its attention to finer details. As shown in \cite{karras2020analyzing}, the low-resolution images are not significantly affected by the higher-resolution layers. Inspired by this observation, we only use layers up to $256\times256$ to find a manipulation direction within the $\mathcal{S}$ space. This strategy eliminates the additional time required for computation within the style blocks corresponding to $512\times512$ and $1024\times1024$ resolutions. 

\item \textbf{Small batch of images.} We compute our loss function using only a small batch of randomly generated images to find the direction $\Delta \mathbf{s}$. Regardless of the text prompt, we find that using a batch of $128$ images is sufficient to find stable and generalizable directions. The reason for this strategy is that the manipulation effect caused by the style channels remains consistent across all images \cite{wu2020stylespace}. 

\item \textbf{Operating on $\mathcal{S}$.} We operate directly on $\mathcal{S}$-space, which is shown to be more disentangled than $\mathcal{W}$ and $\mathcal{W+}$ spaces \cite{wu2020stylespace}. Our method is significantly faster than other methods at finding the desired direction $\Delta \mathbf{s}$ due to the disentangled nature of the $\mathcal{S}$ space. In addition, our framework identifies and uses multiple style channels to compute directions and hence, can perform more complex edits. As shown in previous work \cite{karras2020analyzing,wu2020stylespace}, while single channel manipulations can achieve edits such as changing the \textit{hair color} or \textit{gender} of a face image, they fail at more complex manipulations such as changing \textit{age}, which typically requires a combination of multiple channels, such as \textit{wrinkle, grey hair} and \textit{eyeglasses}. Therefore, our method incorporates multiple style channels and captures complex manipulations such as \textit{age} or change of \textit{personal identity} while finding the desired direction.

\end{itemize} 

Refer to the ablation study in Section \ref{sec:ablation} to see how these insights affect manipulation and computation time.

\section{Experiments}
\label{sec:experiments}

We evaluate the proposed method on StyleGAN2 on a variety of datasets, including FFHQ \cite{StyleGAN}, LSUN Car, Church, Horse \cite{yu2015lsun}, AFHQ Cat, Dog, Wild \cite{Choi2020StarGANVD} and MetFaces \cite{Karras2020TrainingGA} datasets. We also compare our method to the state-of-the-art text-based manipulation methods StyleCLIP and TediGAN, and the unsupervised methods GANspace, SeFa, and LatentCLR. Next, we discuss our experimental setup and then present results for several StyleGAN2 models.

\begin{figure*}[t!]
\begin{center}
\includegraphics[width=\textwidth]{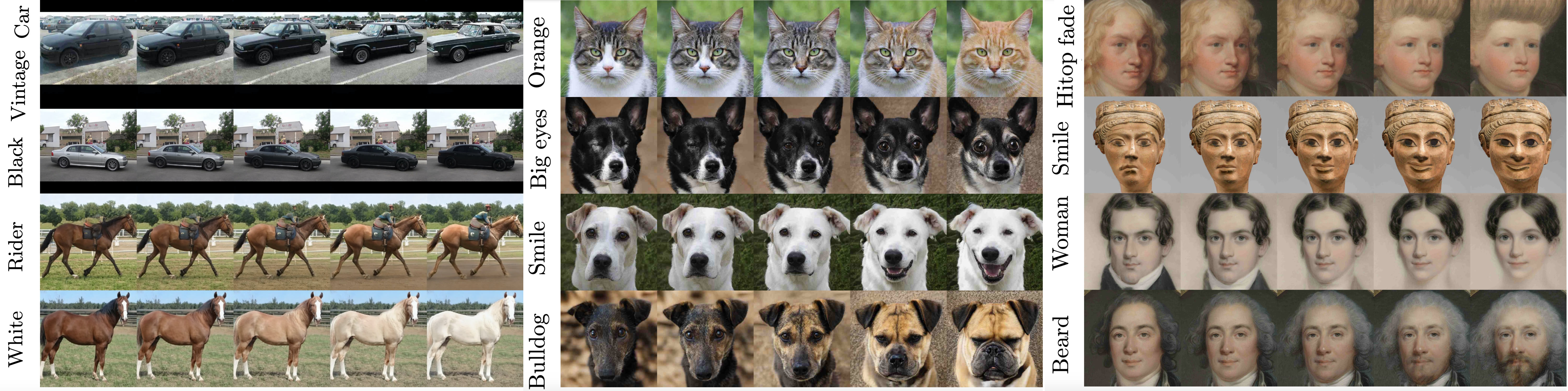}
(a)  \hspace{5cm}  (b)  \hspace{5cm}   (c) 
\caption{(a) Edits on the LSUN Car and Horse. (b) Edits on the AFHQ Cat and Dog. (c) Edits on the MetFaces. The input images are in the middle, and left and right show the manipulation in the positive and negative directions.}
\label{fig:faces_animals_met}
\end{center}
\vskip -0.25in
\end{figure*}

\subsection{Experimental Setup}
\label{sec:implementation}
For manipulation experiments on real images, we use the e4e method and map their latent codes from the $\mathcal{W+}$ space to the $\mathcal{S}$ space. The dimension of the $\textbf{s}$ vectors we use to generate images is 9088. Following \cite{wu2020stylespace}, we exclude $s_{tRGB}$ layers since they cause entangled manipulations and transform the entire image. Additionally, we exclude style channel parameters of the last 4 blocks when finding $\Delta \mathbf{s}$ as they represent very fine-grained features and are difficult to be used for editing tasks. For all experiments, we set the coefficient of the CLIP loss term to 1, while the coefficient of the identity loss term takes values between 0.1 and 6. Since it is desirable to retain features other than the target attribute, e.g., facial identity for FFHQ, the identity loss is increased so that the found direction is not affected by common model biases such as making target face younger as applying makeup. However, for manipulations that involve a complete change in identity, such as Donald Trump, the identity loss coefficient should be set to a low value. In most of our experiments, we used a coefficient value of 0.1, 0.5, or 2, depending on the complexity of the manipulation. We used a single Titan RTX GPU for our experiments. For StyleCLIP, reported times are taken directly from \cite{patashnik2021styleclip}. For TediGAN and StyleCLIP, we use the official Pytorch implementations.\footnote{\url{https://github.com/weihaox/TediGAN}, \url{https://github.com/orpatashnik/StyleCLIP}}

\begin{figure}[t!]
\begin{center}
\centerline{\includegraphics[width=1\columnwidth]{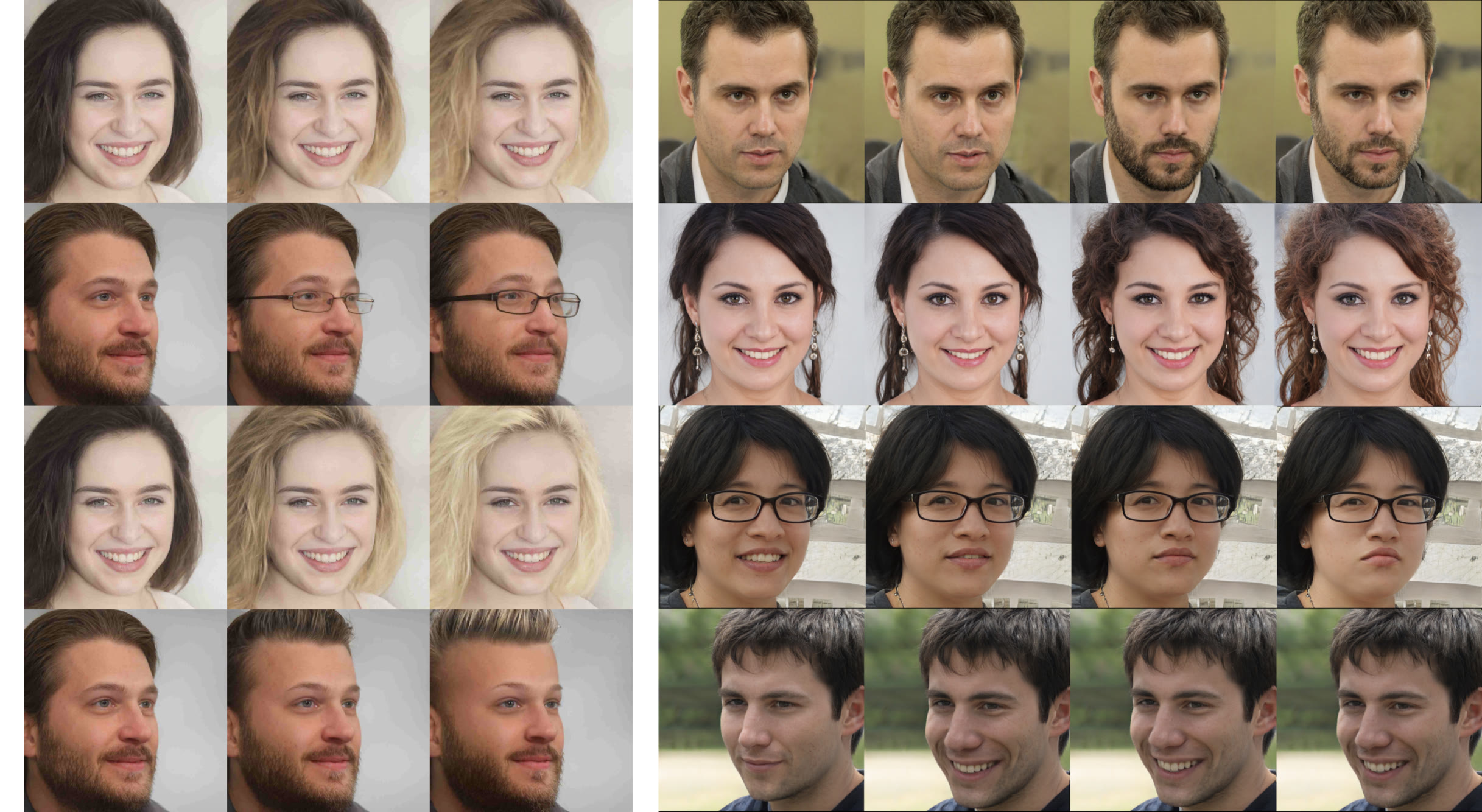}} 
 (a)  \hspace{.2\textwidth} (b)
\caption{(a) Comparison of single (top two rows) and multi-channel methods (bottom two rows) using Blonde and Mohawk prompts. (b) Manipulations with increasing batch sizes $32$, $128$, and $1024$ (top two rows) and increasing resolution $128$, $256$, $1024$ (bottom two rows) using Beard, Curly Hair, Frown and Smile prompts. }
\label{fig:combined_singlemulti}
\end{center}
\vskip -0.3in
\end{figure}

\begin{figure*}[]
	\setlength{\tabcolsep}{1pt}
	\begin{center}
	{\footnotesize
	\begin{tabular}{ccccccccccccccc}
		  \tiny{Input} & \tiny{TediGAN} & \tiny{sCLIP-GD} & \tiny{sCLIP-LM} & \tiny{Ours} & \tiny{Input} & \tiny{TediGAN} & \tiny{sCLIP-GD} & \tiny{sCLIP-LM} & \tiny{Ours}  & \tiny{Input} & \tiny{TediGAN} & \tiny{sCLIP-GD} & \tiny{sCLIP-LM} & \tiny{Ours} \\
        
		\includegraphics[width=0.06\textwidth]{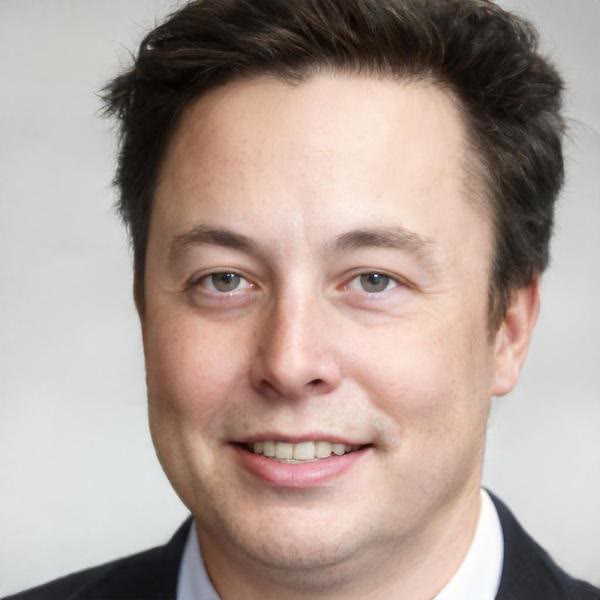} &
		\includegraphics[width=0.06\textwidth]{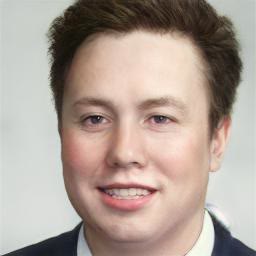} &
		\includegraphics[width=0.06\textwidth]{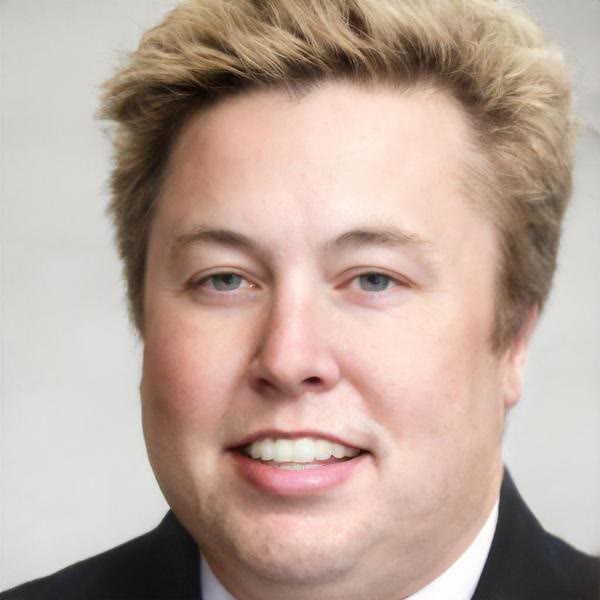} &
		\includegraphics[width=0.06\textwidth]{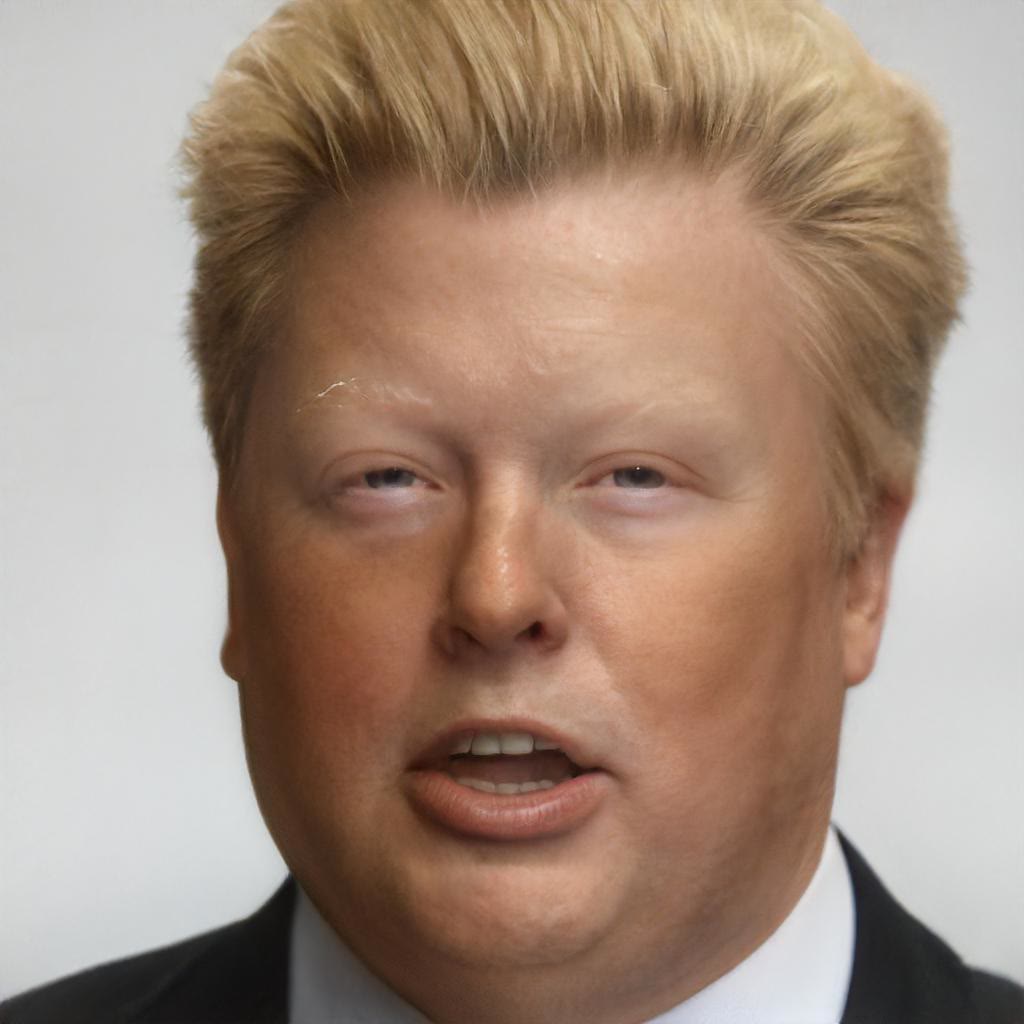} &
		\includegraphics[width=0.06\textwidth]{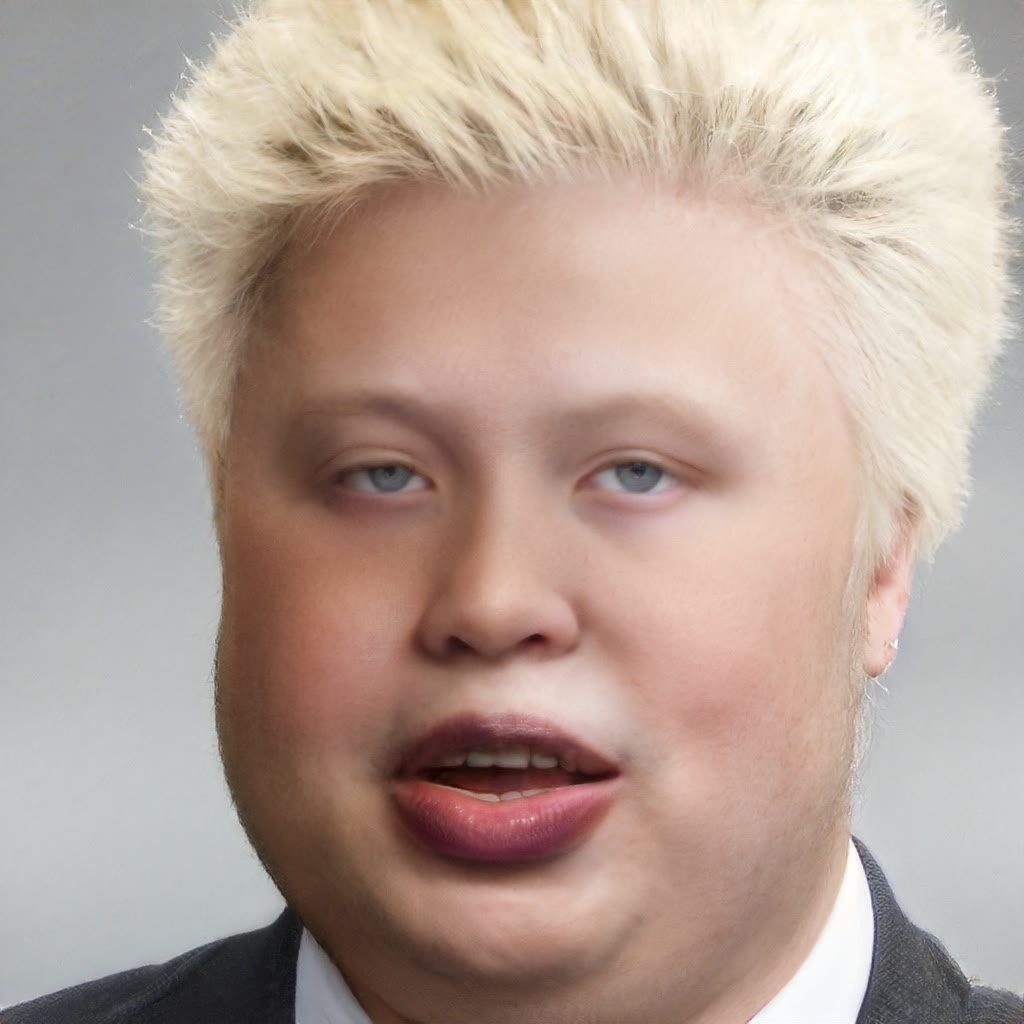} &
        
		\includegraphics[width=0.06\textwidth]{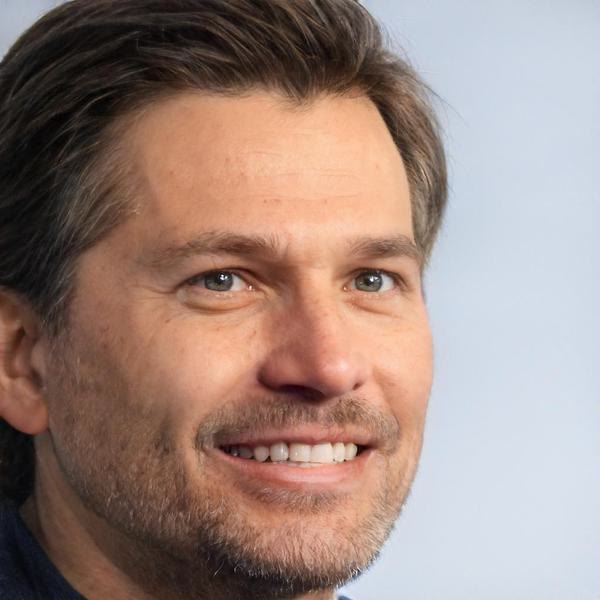} &
		\includegraphics[width=0.06\textwidth]{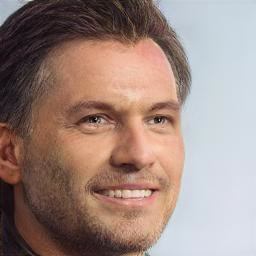} &
		\includegraphics[width=0.06\textwidth]{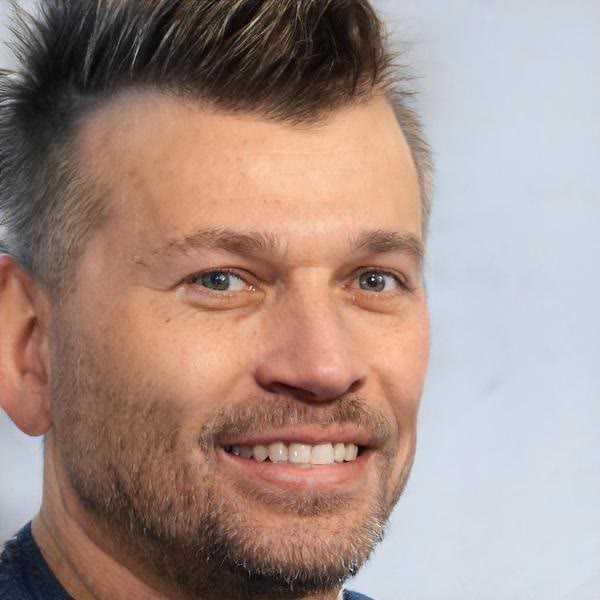} &
		\includegraphics[width=0.06\textwidth]{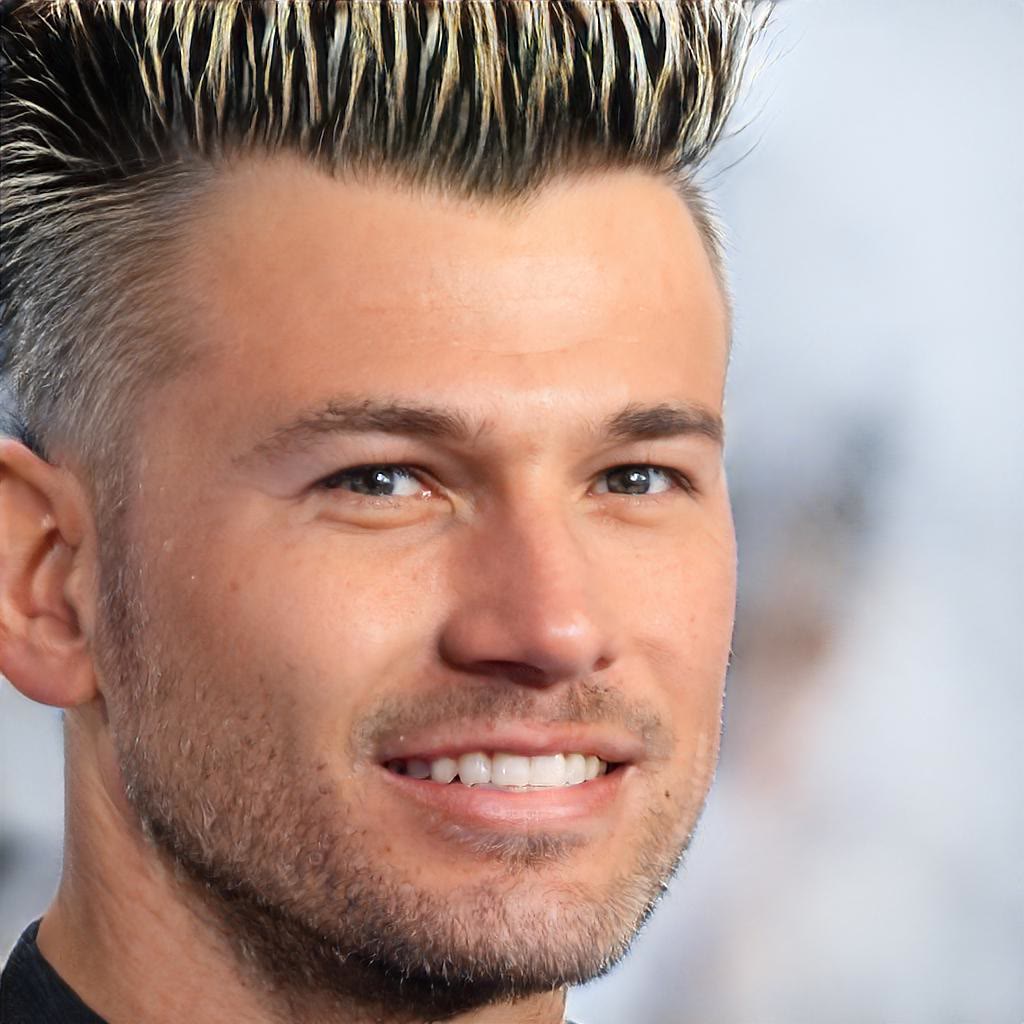} &
		\includegraphics[width=0.06\textwidth]{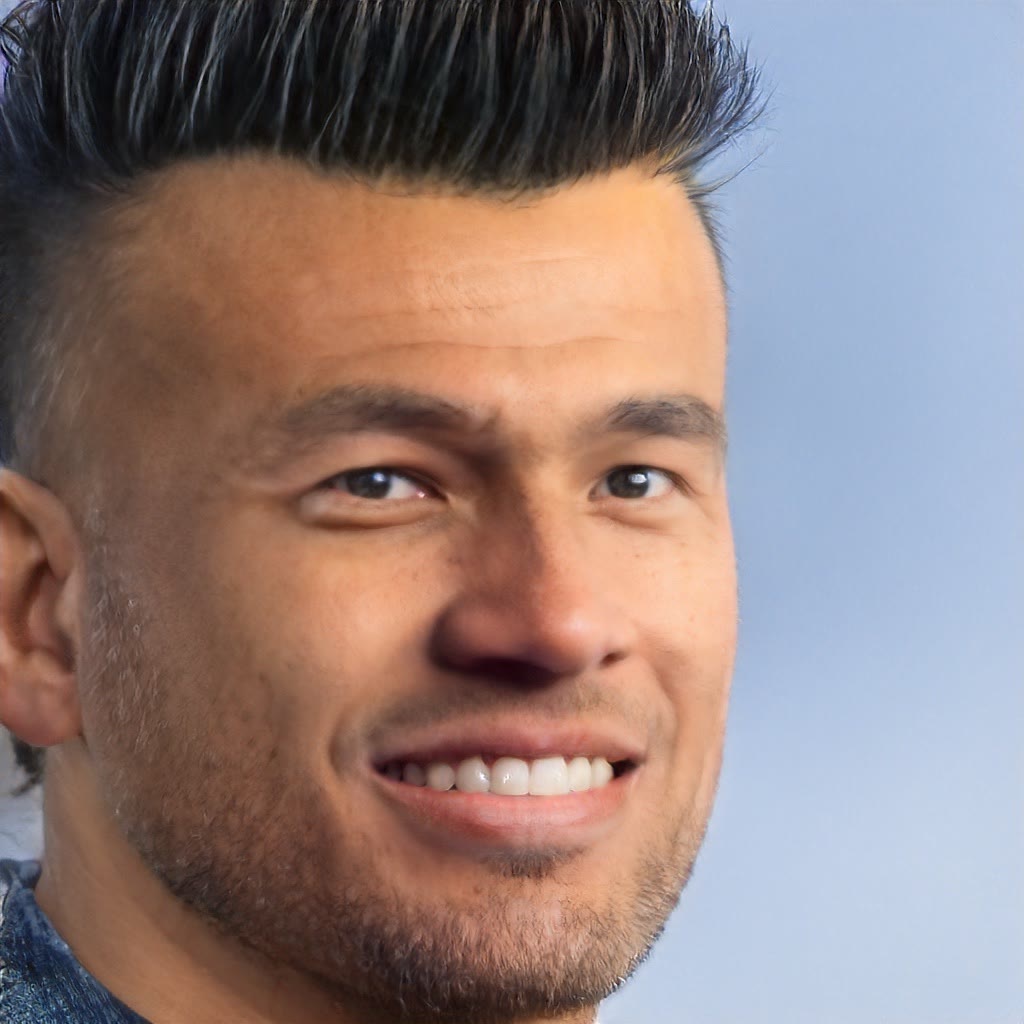} &
		
		\includegraphics[width=0.06\textwidth]{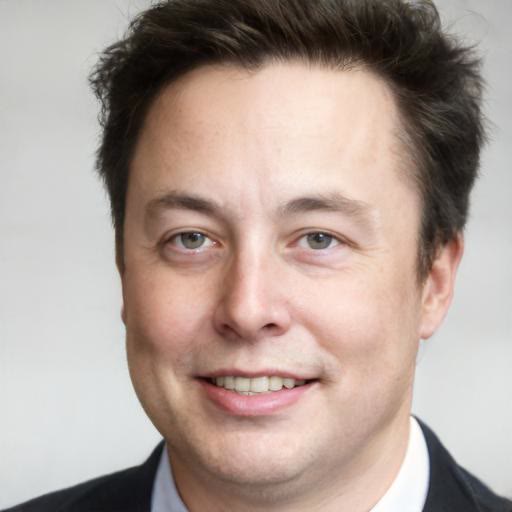} &
		\includegraphics[width=0.06\textwidth]{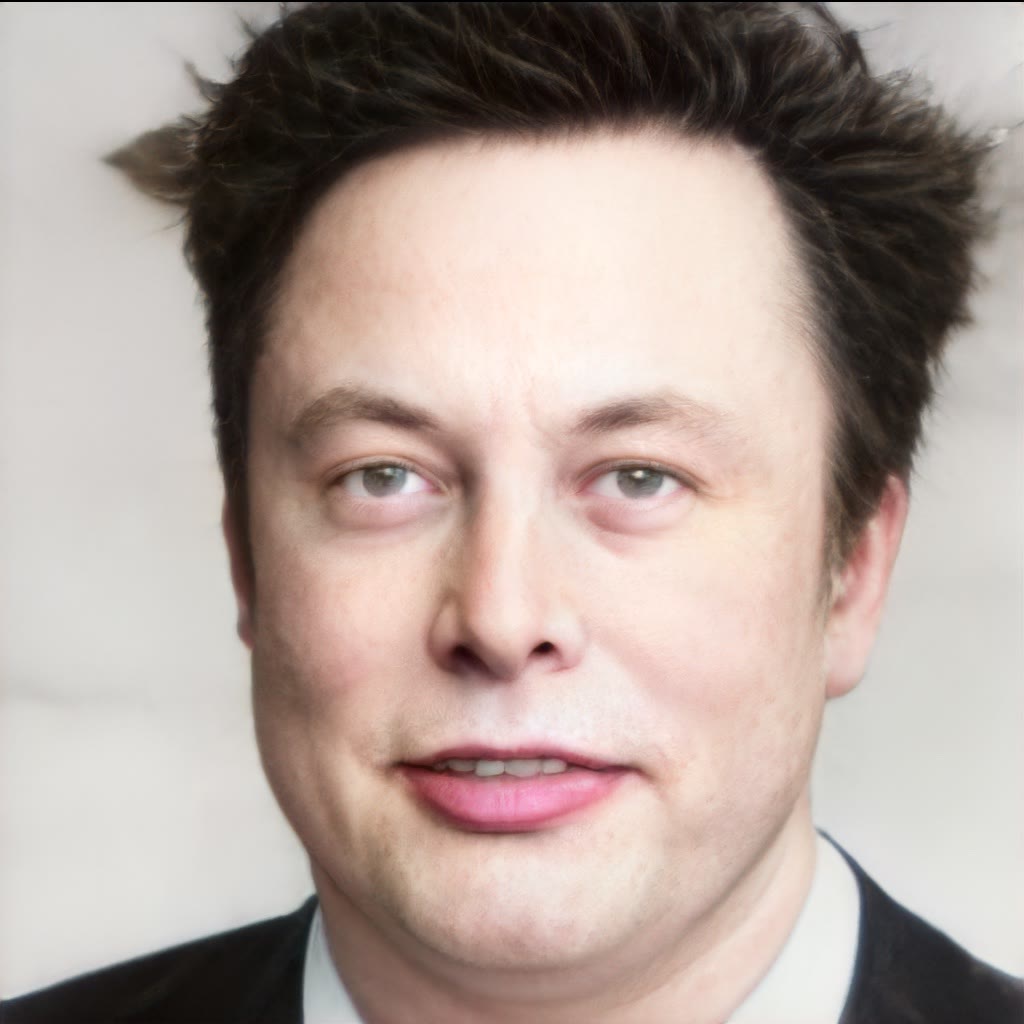} &
		\includegraphics[width=0.06\textwidth]{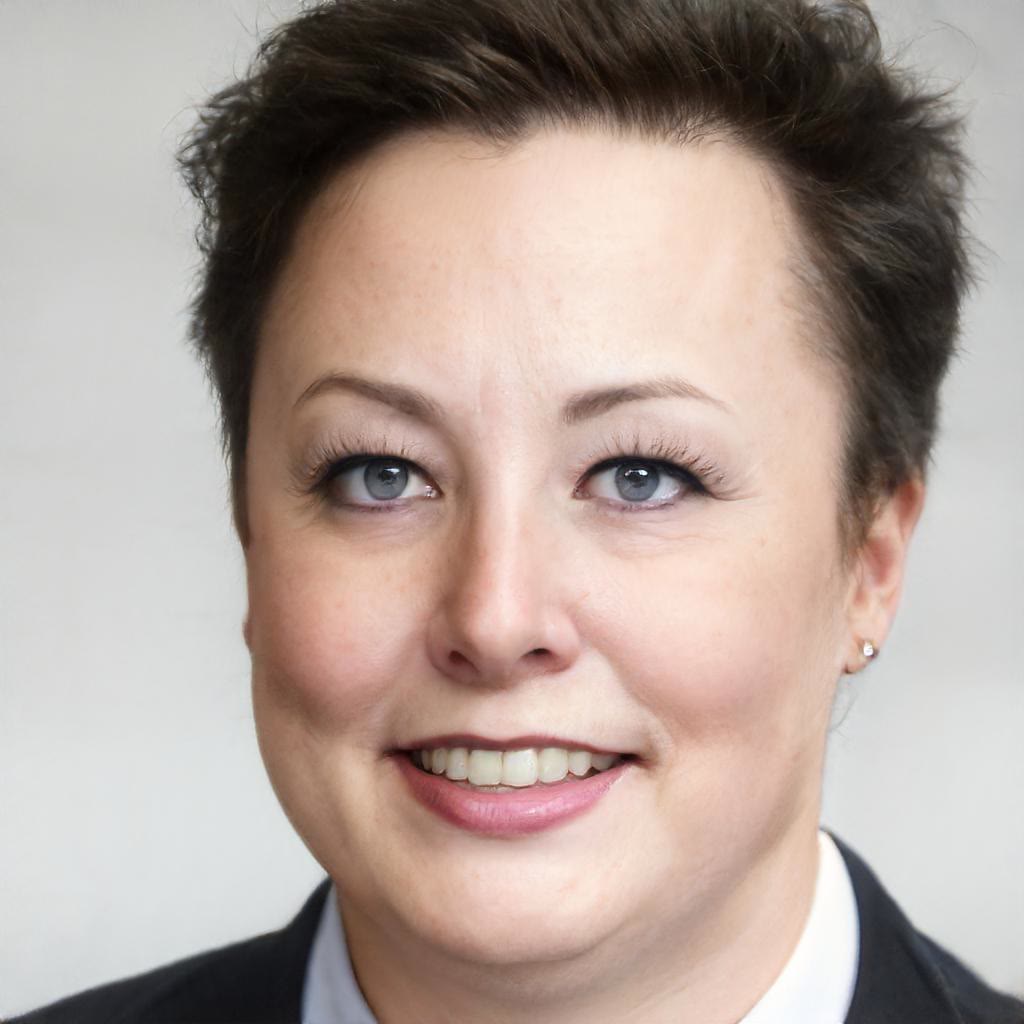} &
		\includegraphics[width=0.06\textwidth]{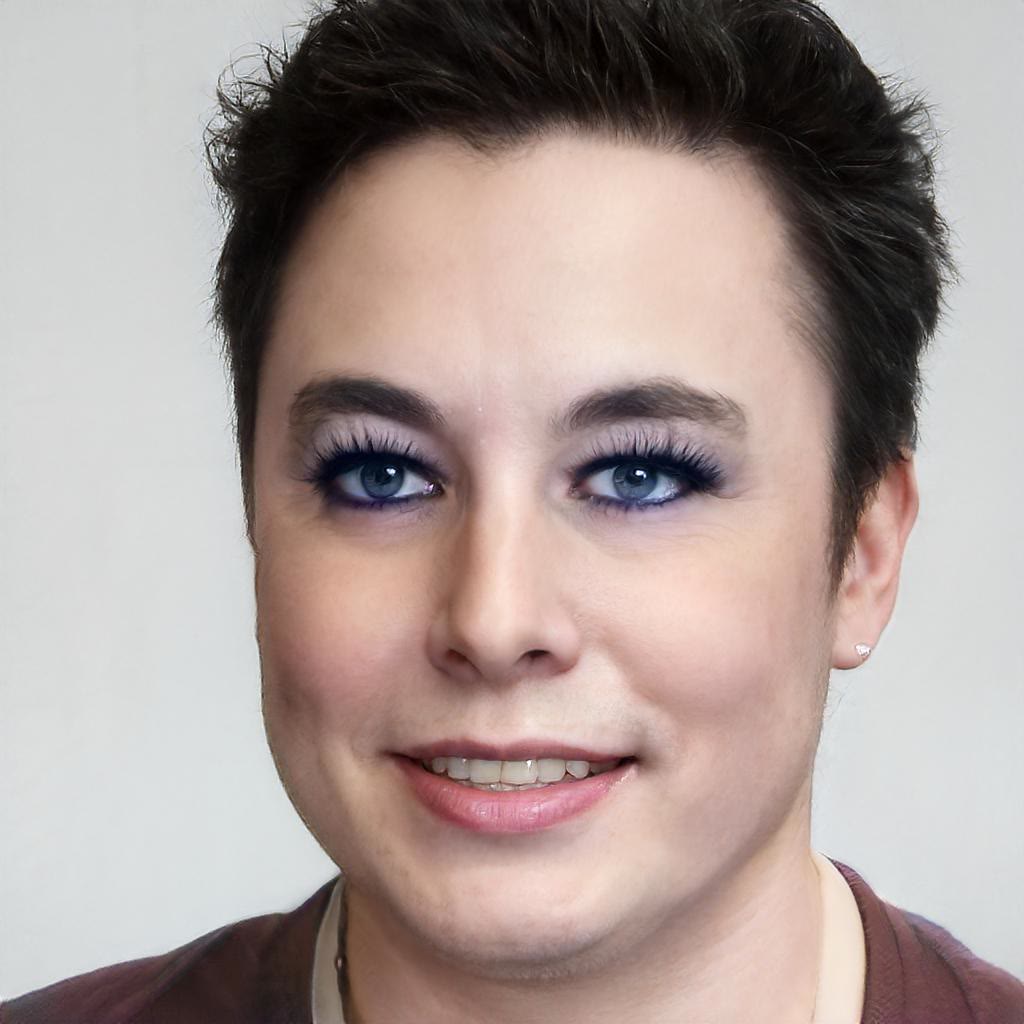} &
		\includegraphics[width=0.06\textwidth]{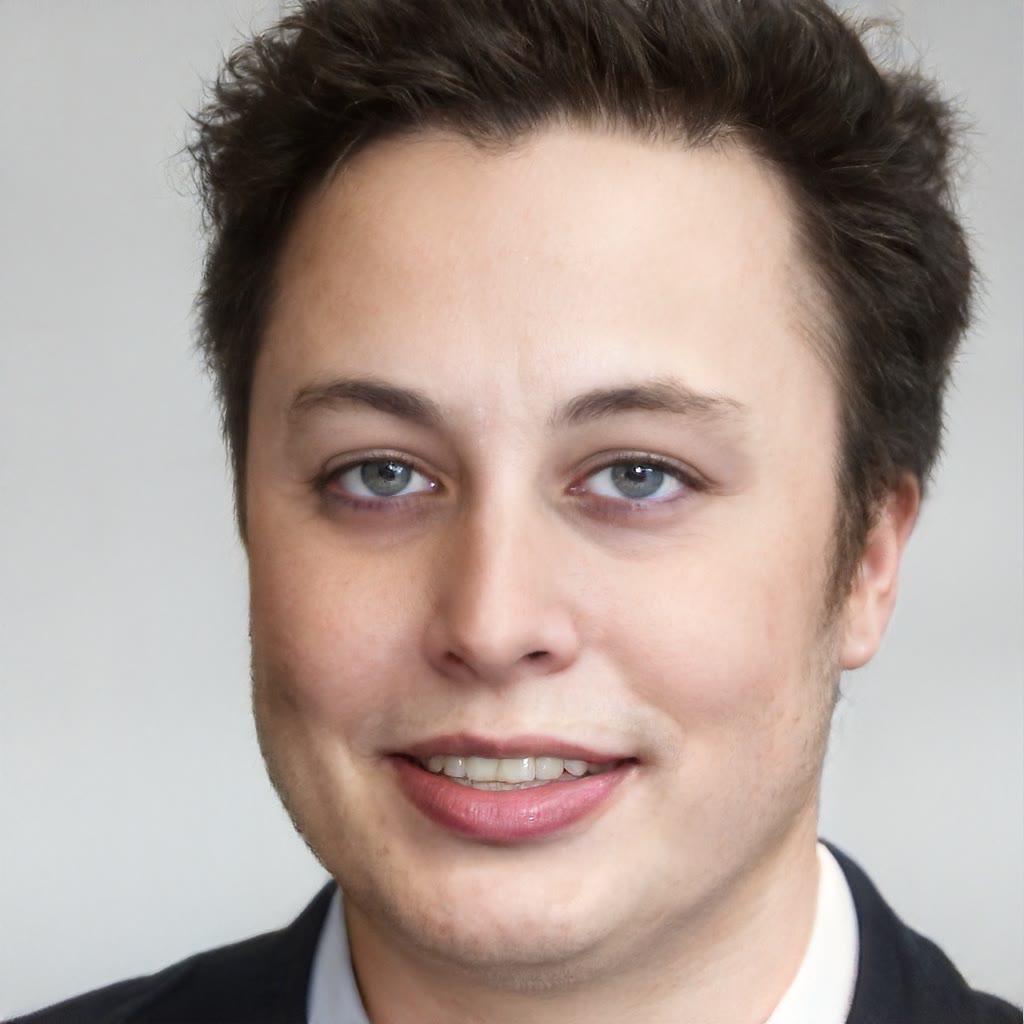} \\

		\includegraphics[width=0.06\textwidth]{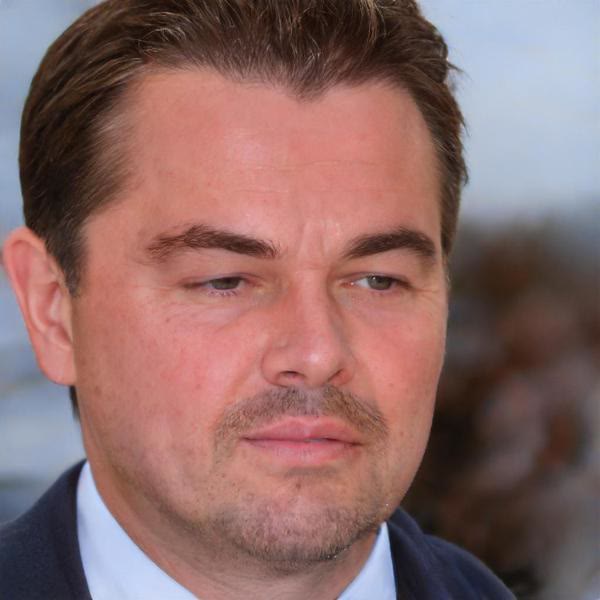} &
		\includegraphics[width=0.06\textwidth]{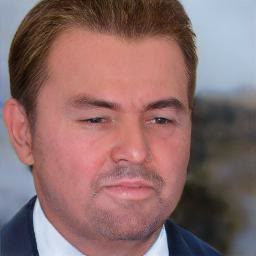} &
		\includegraphics[width=0.06\textwidth]{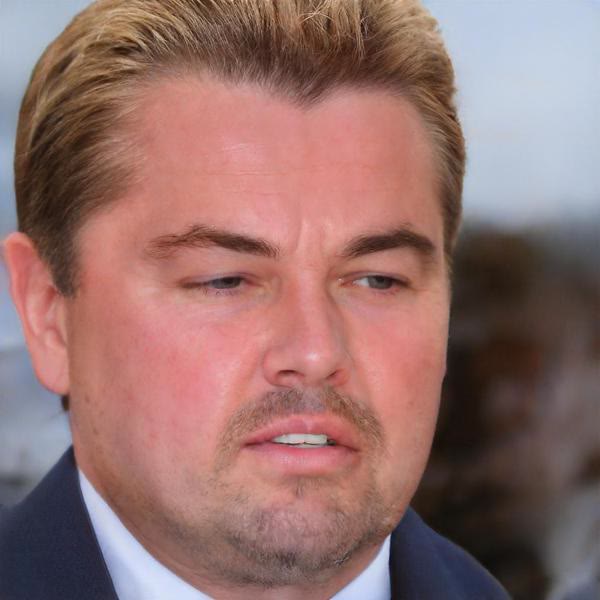} &
		\includegraphics[width=0.06\textwidth]{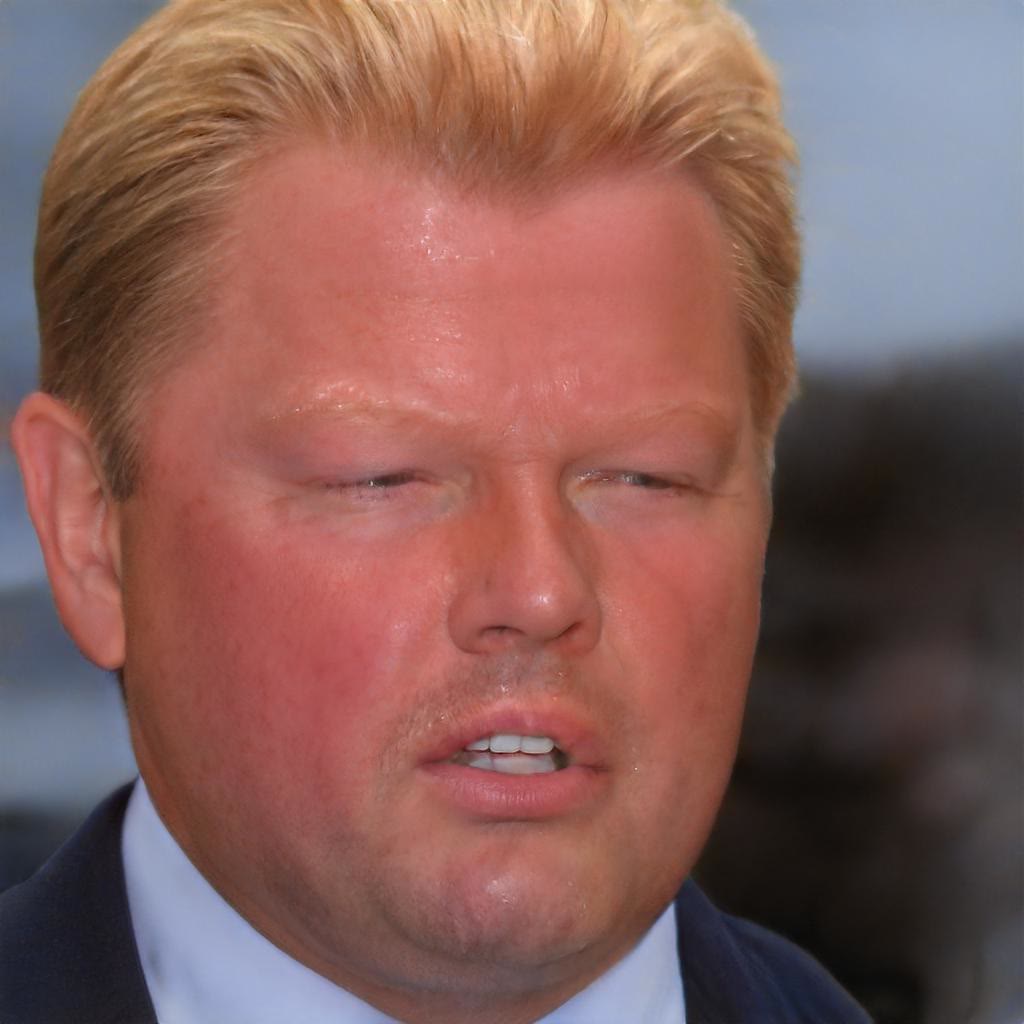} &
		\includegraphics[width=0.06\textwidth]{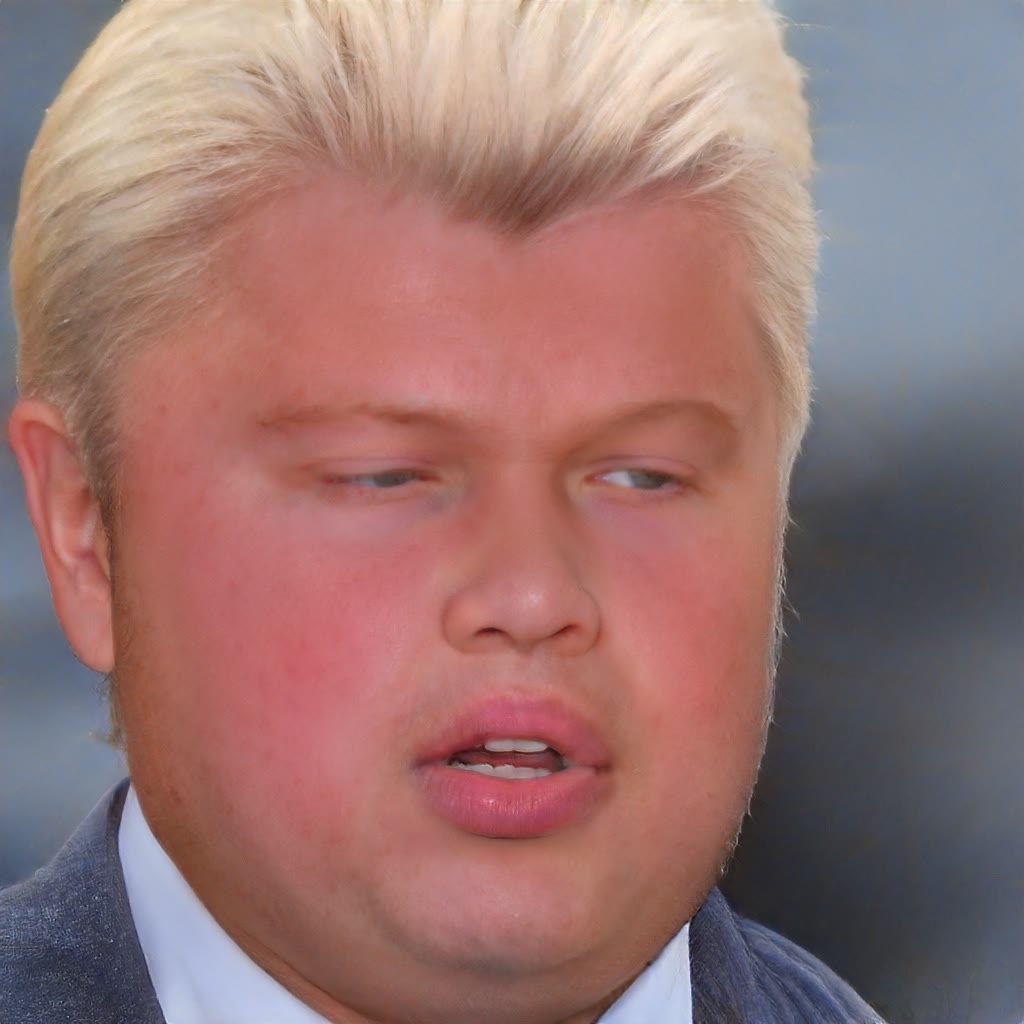} &
		
		\includegraphics[width=0.06\textwidth]{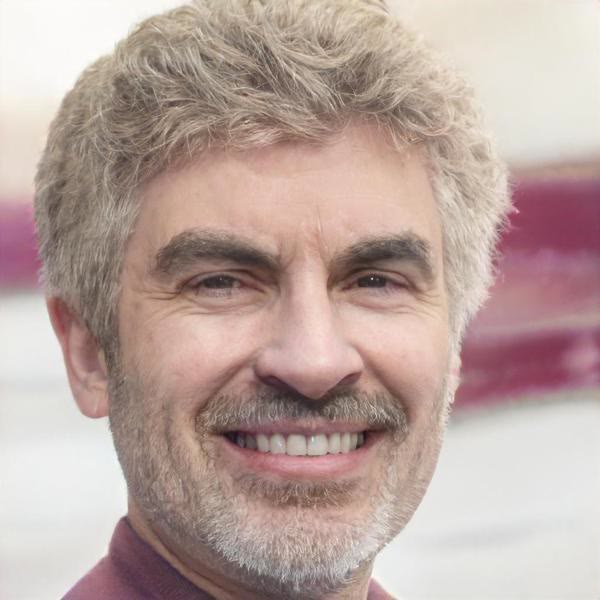} &
		\includegraphics[width=0.06\textwidth]{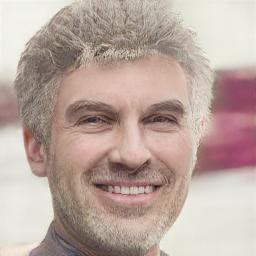} &
		\includegraphics[width=0.06\textwidth]{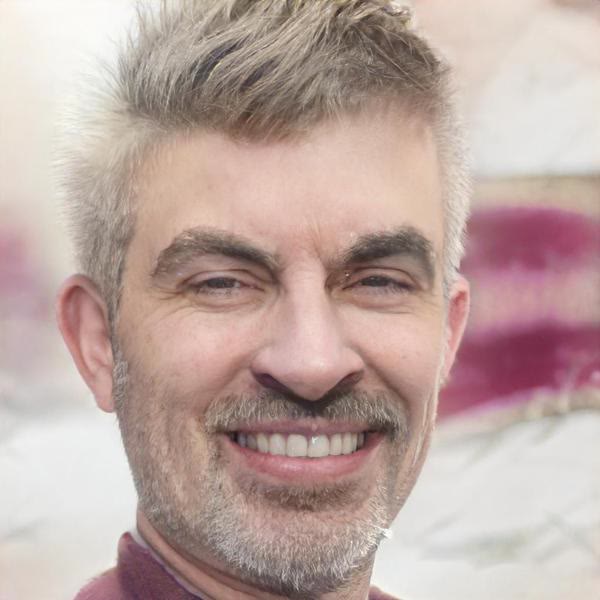} &
		\includegraphics[width=0.06\textwidth]{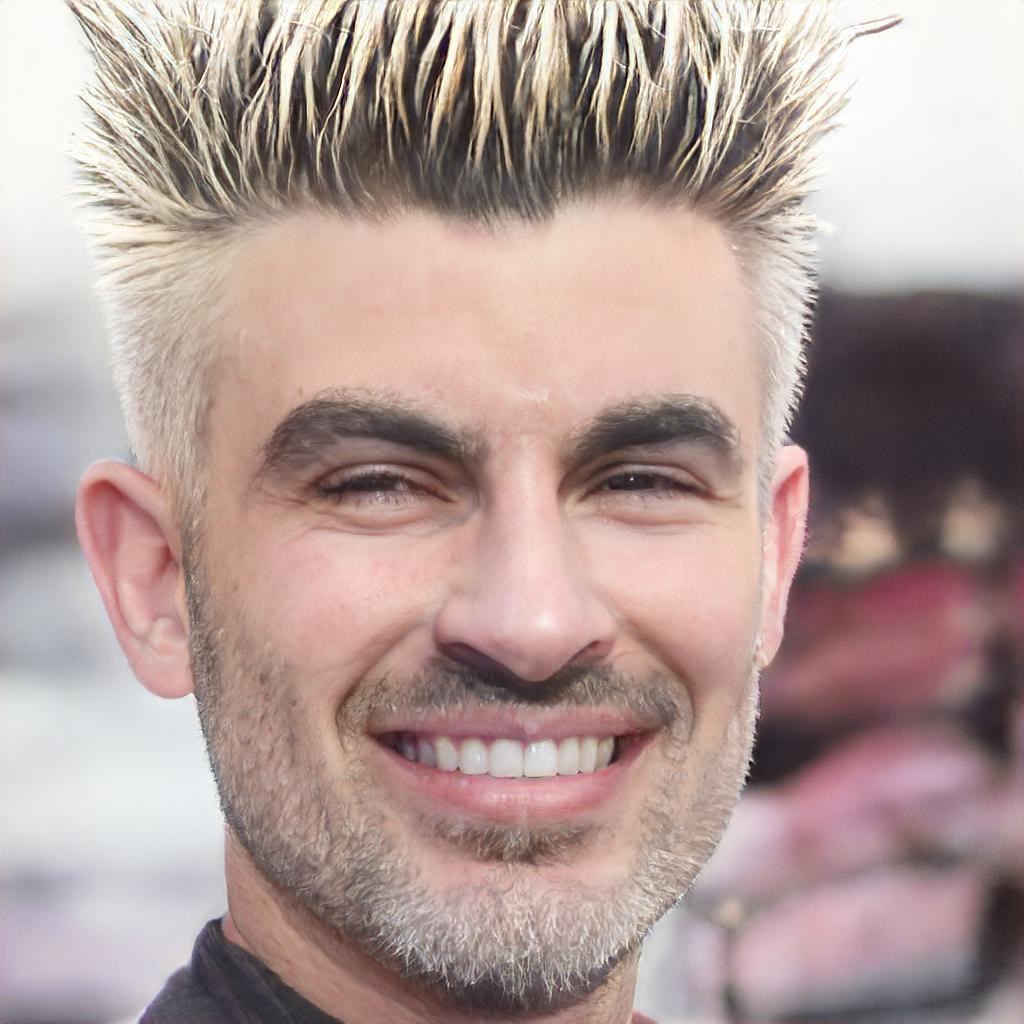} &
		\includegraphics[width=0.06\textwidth]{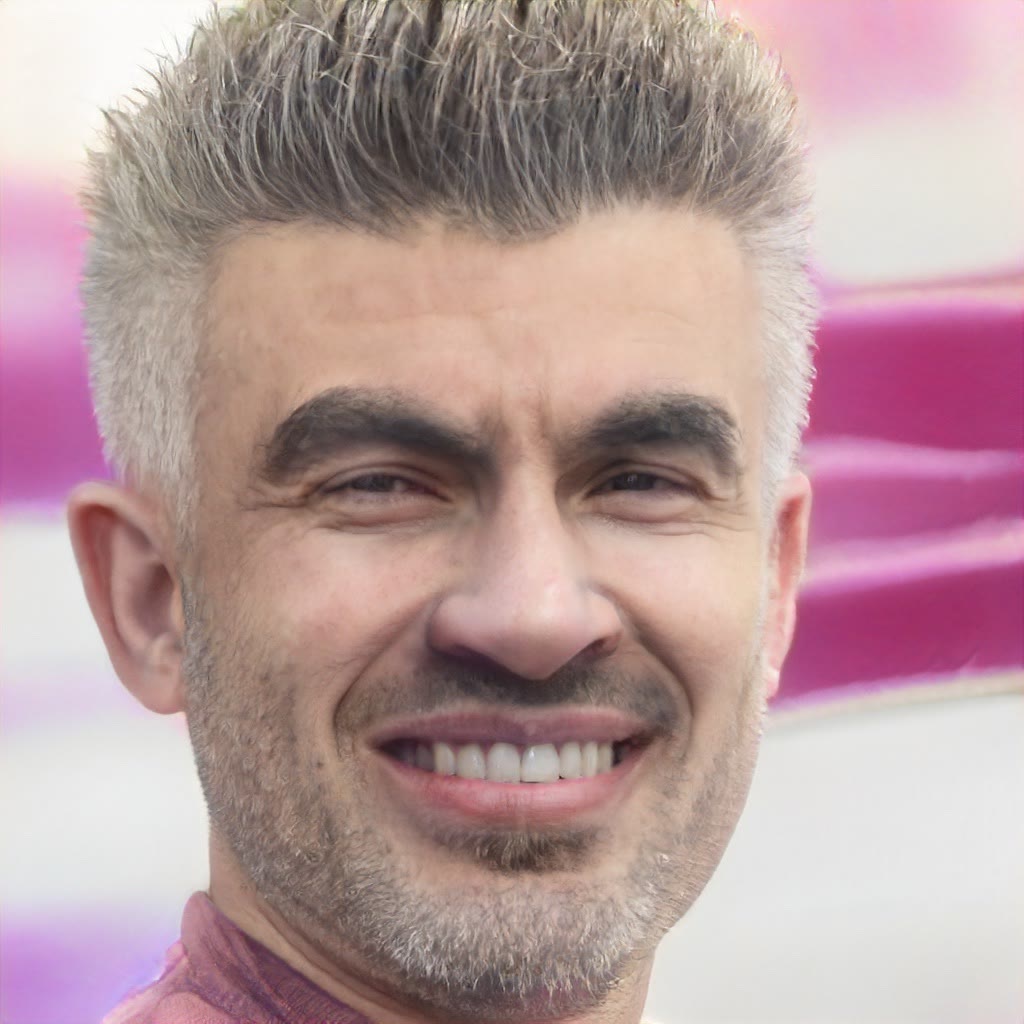} &
		
		\includegraphics[width=0.06\textwidth]{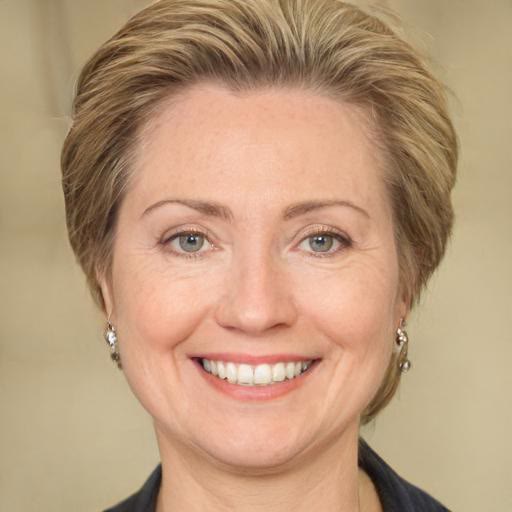} &
		\includegraphics[width=0.06\textwidth]{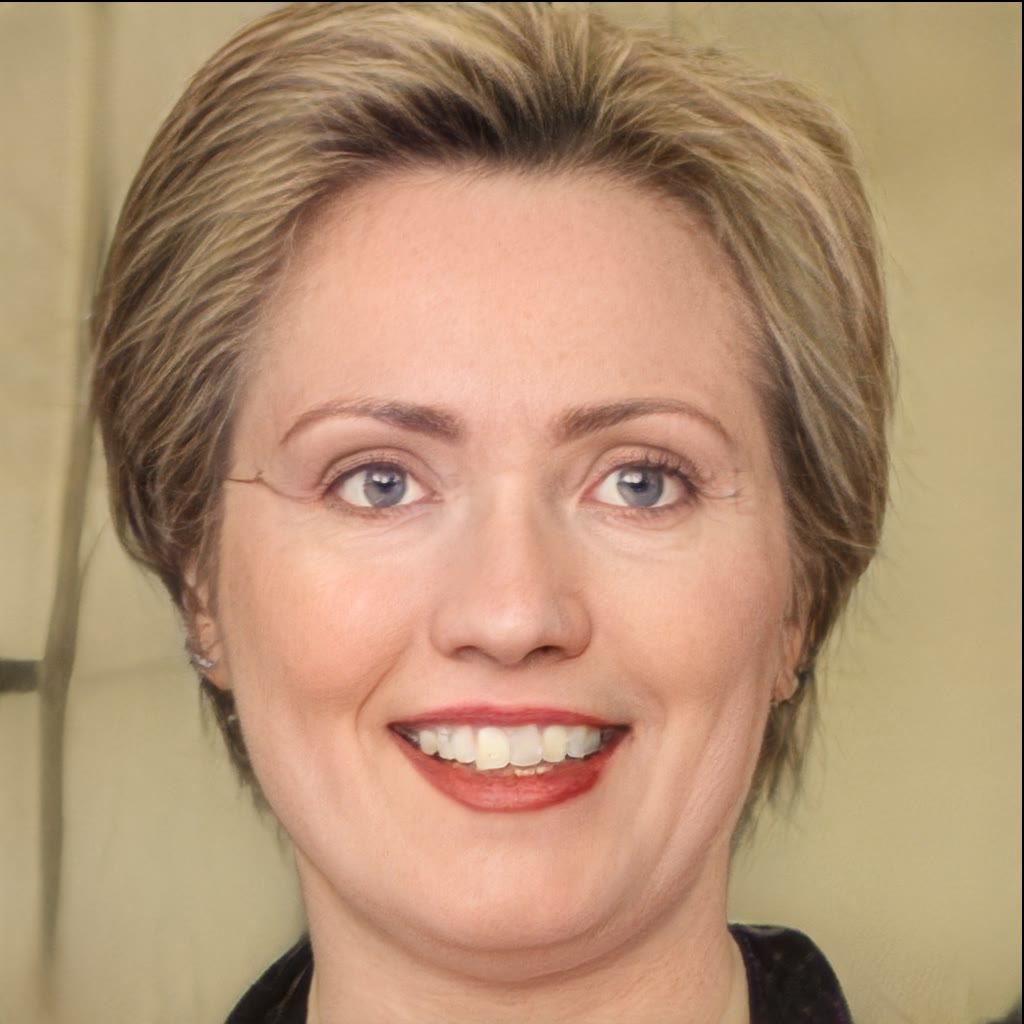} &
		\includegraphics[width=0.06\textwidth]{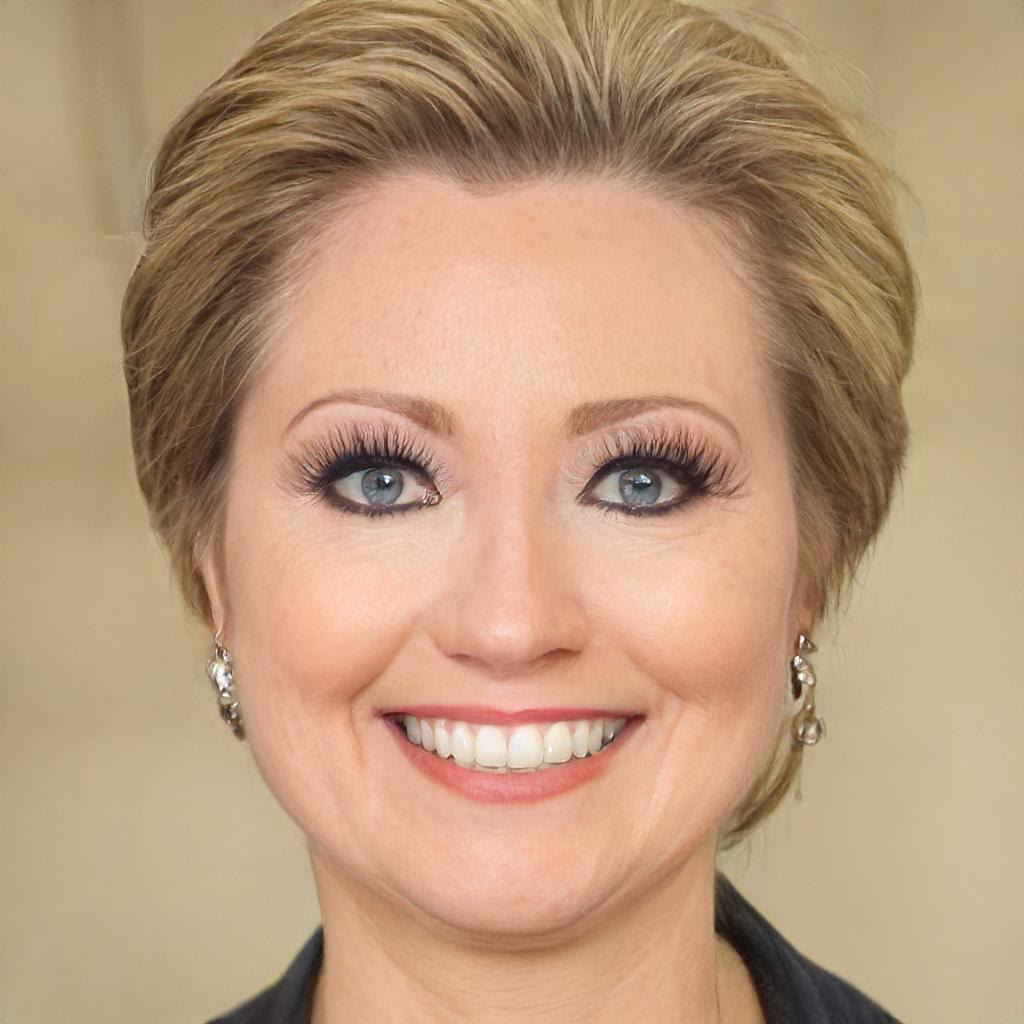} &
		\includegraphics[width=0.06\textwidth]{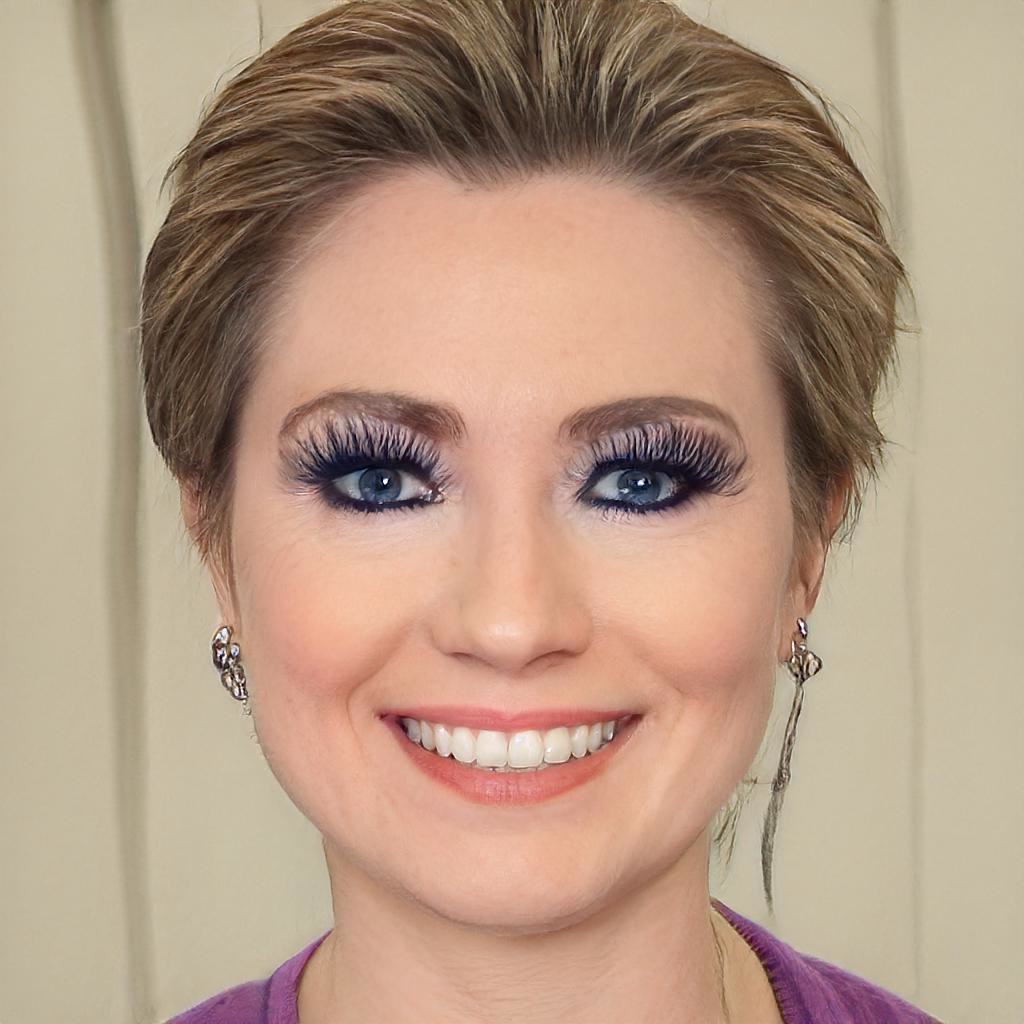} &
		\includegraphics[width=0.06\textwidth]{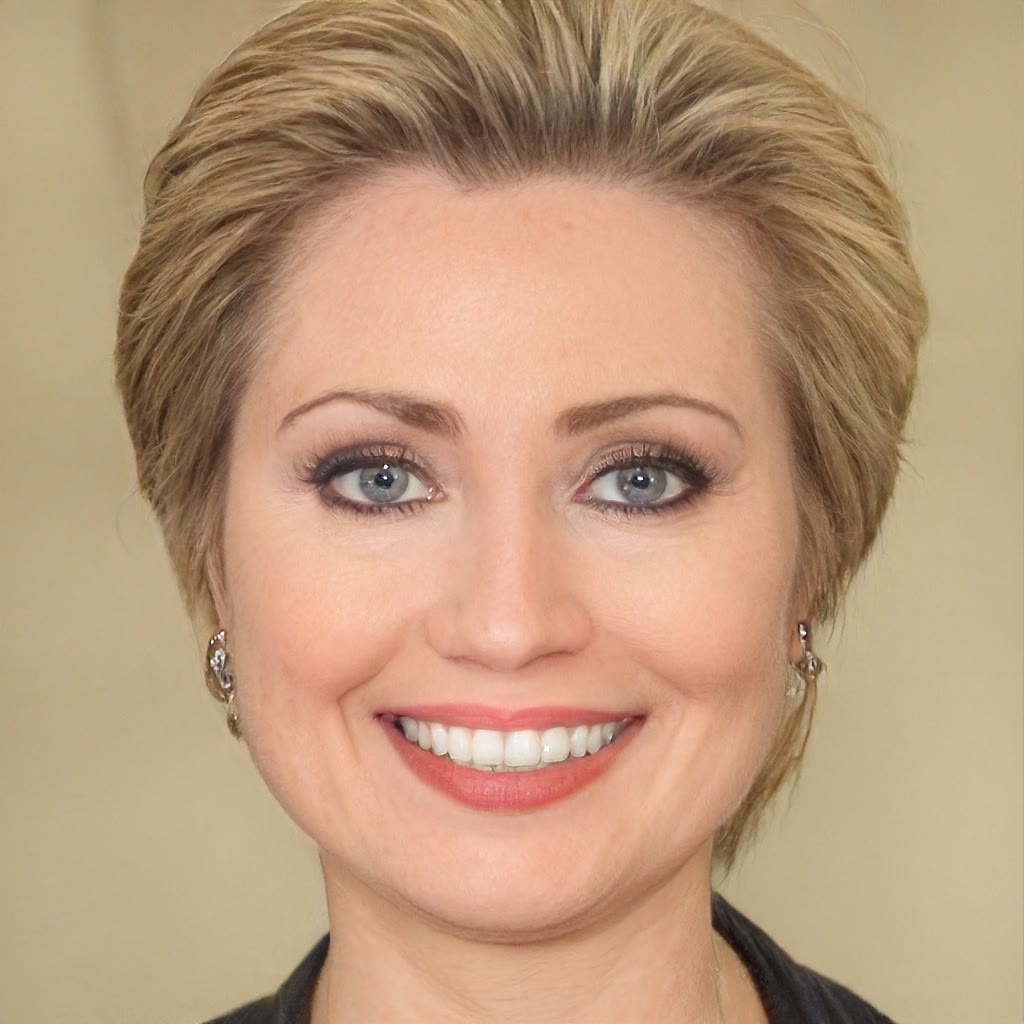} \\
		
		\includegraphics[width=0.06\textwidth]{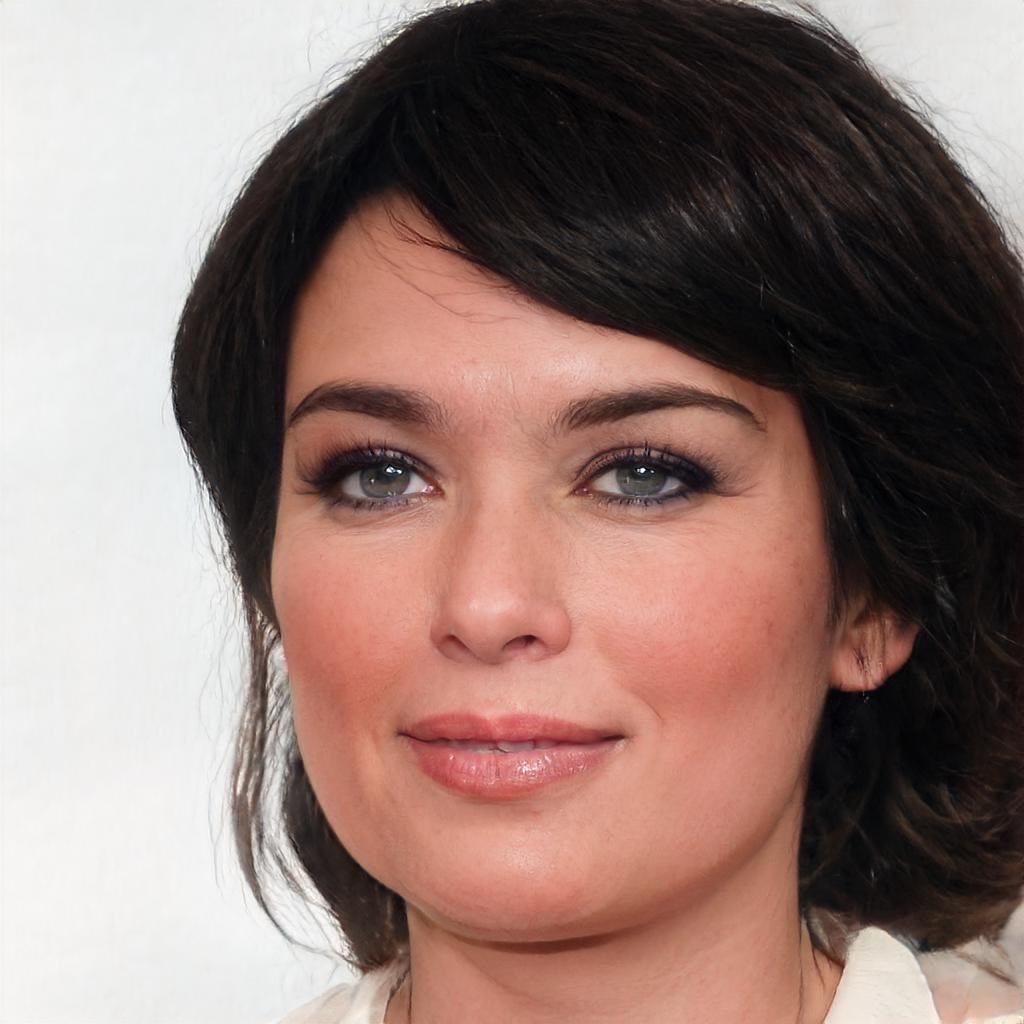} &
		\includegraphics[width=0.06\textwidth]{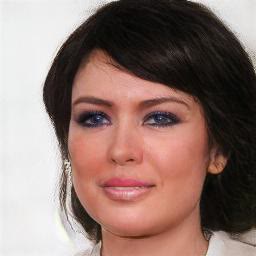} &
		\includegraphics[width=0.06\textwidth]{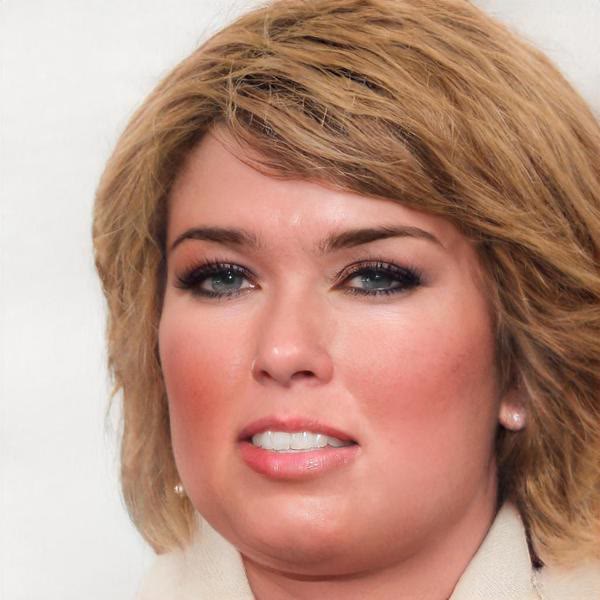} &
		\includegraphics[width=0.06\textwidth]{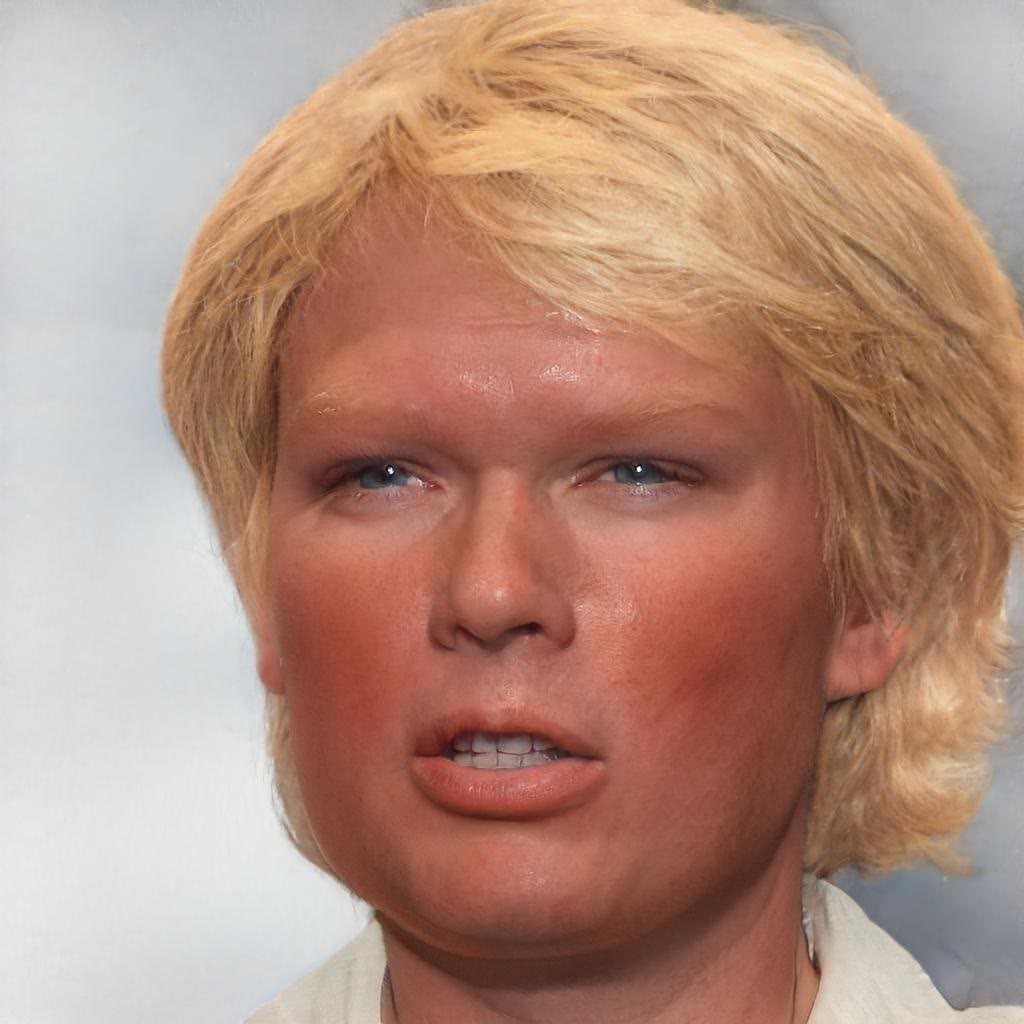} &
		\includegraphics[width=0.06\textwidth]{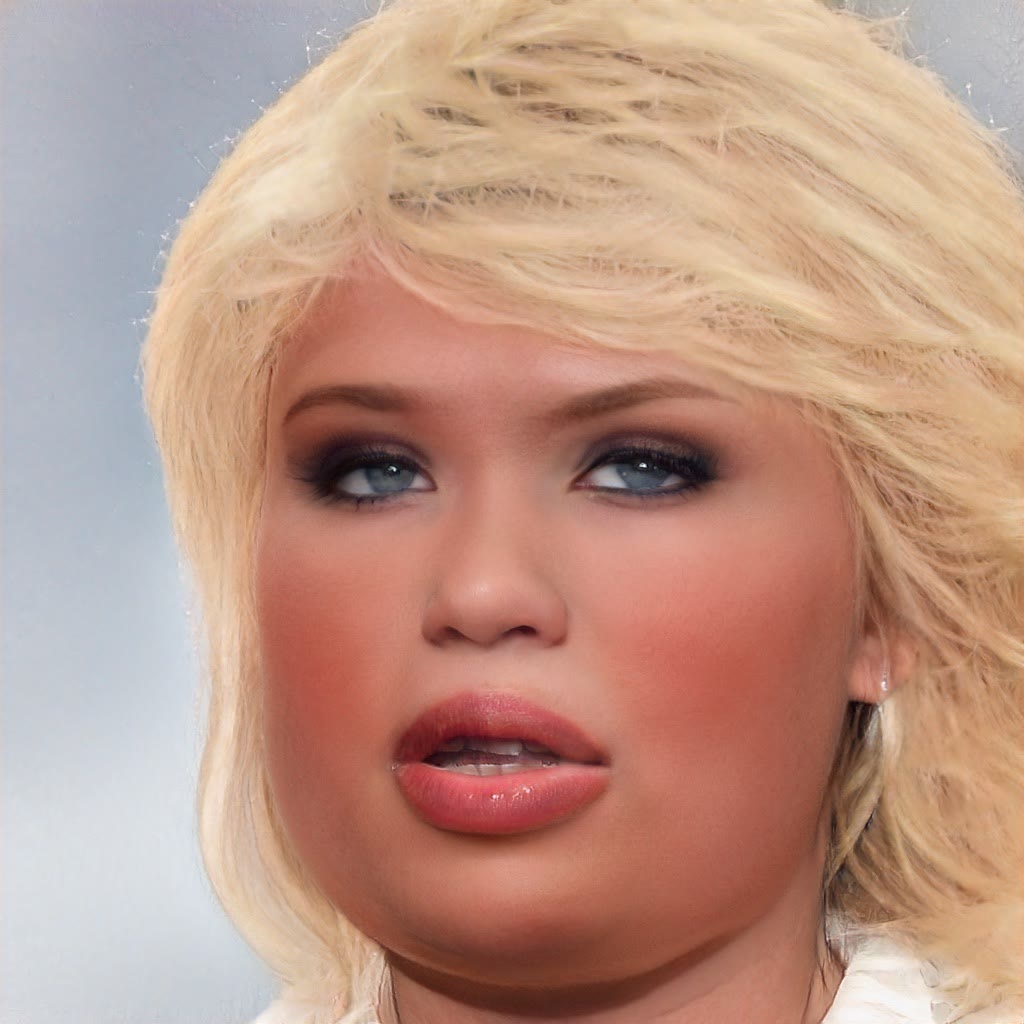} &
		
		\includegraphics[width=0.06\textwidth]{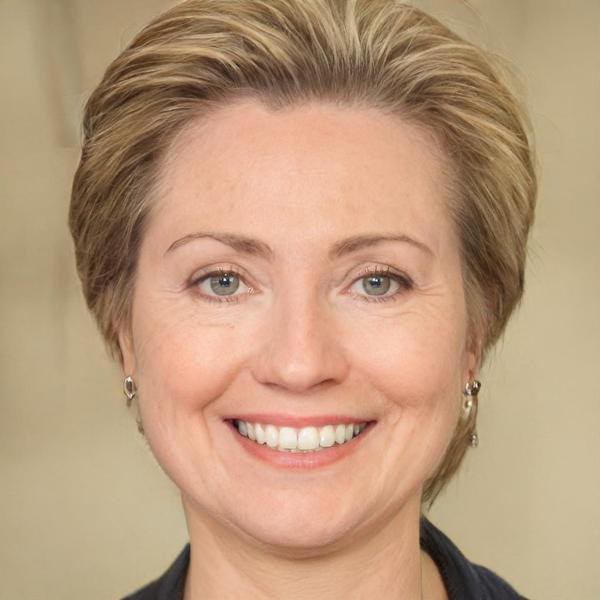} &
		\includegraphics[width=0.06\textwidth]{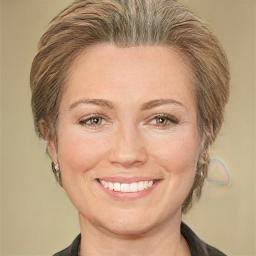} &
		\includegraphics[width=0.06\textwidth]{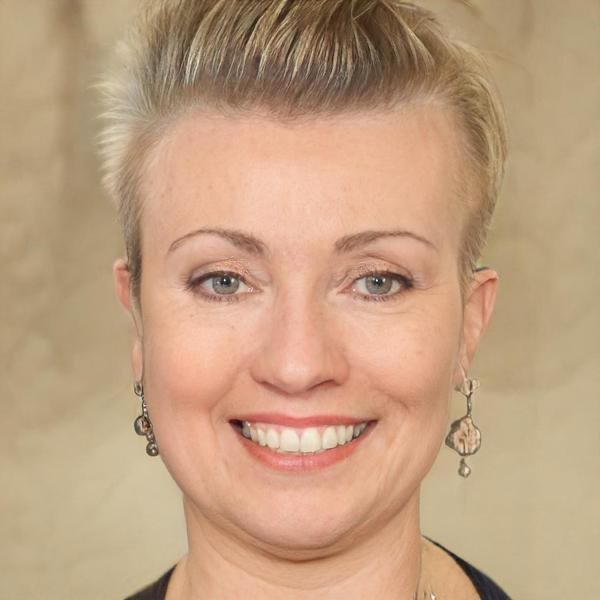} &
		\includegraphics[width=0.06\textwidth]{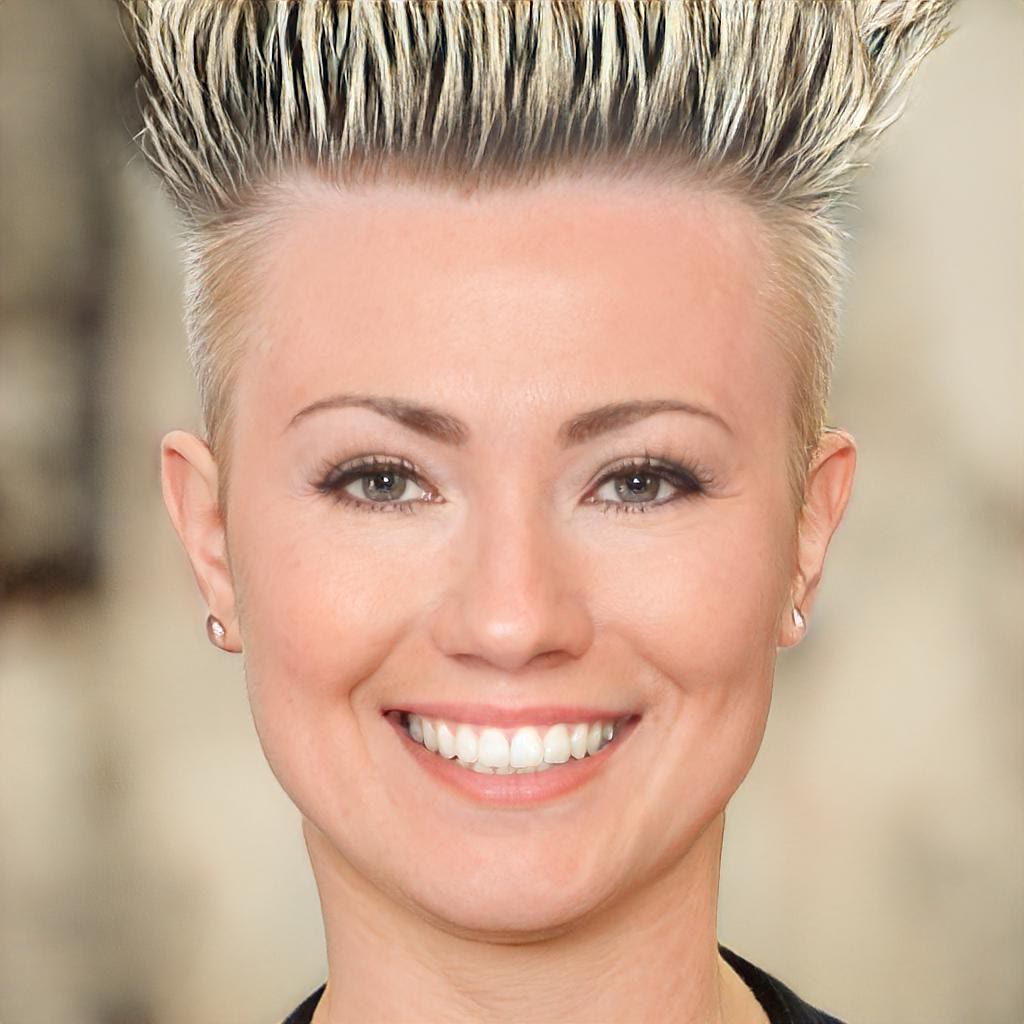} &
		\includegraphics[width=0.06\textwidth]{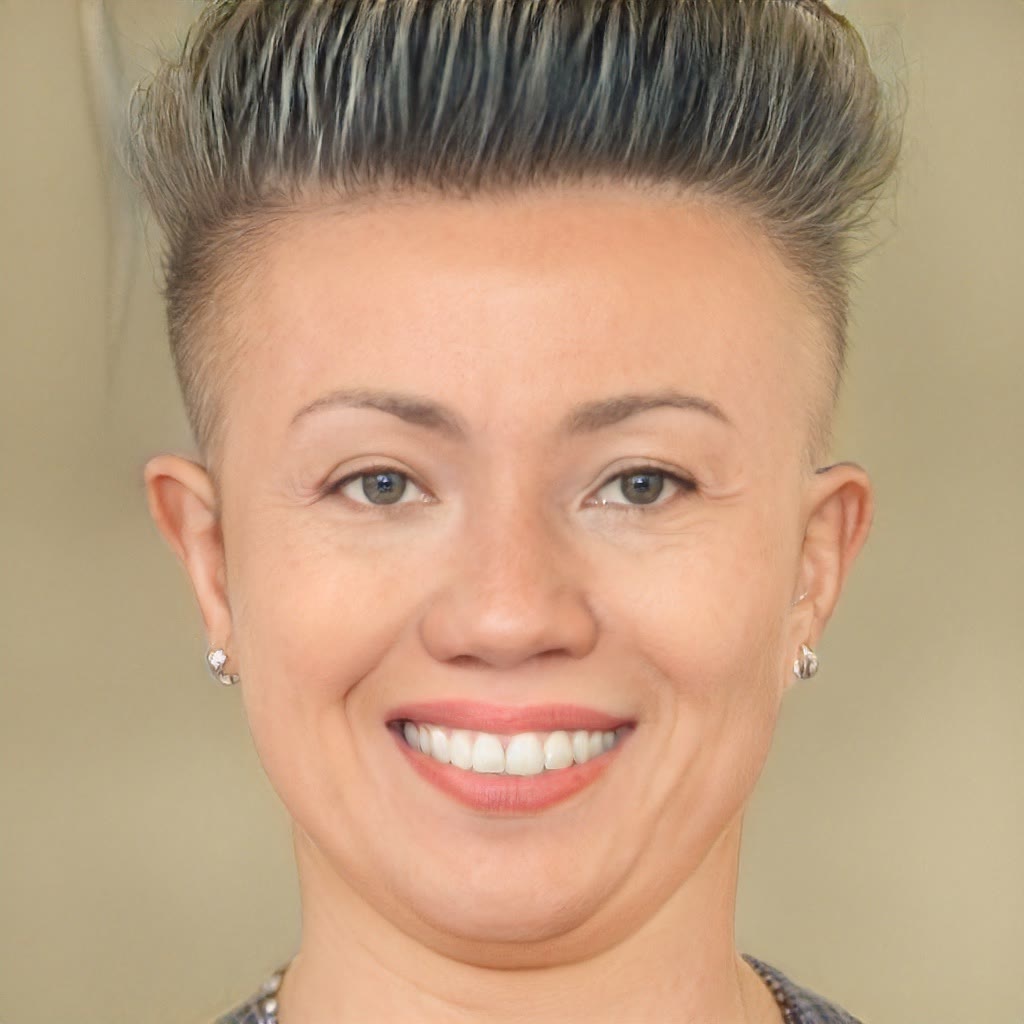} &
		
		\includegraphics[width=0.06\textwidth]{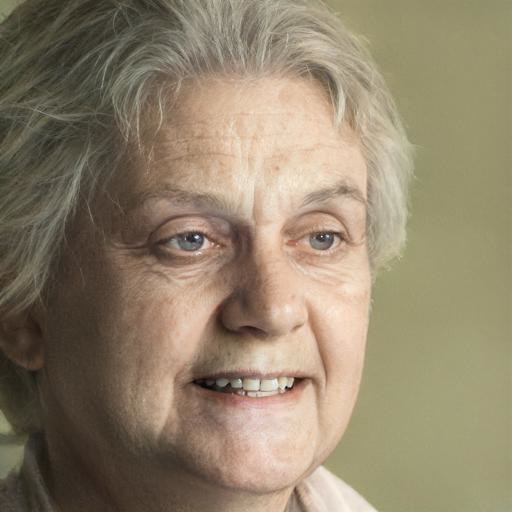} &
		\includegraphics[width=0.06\textwidth]{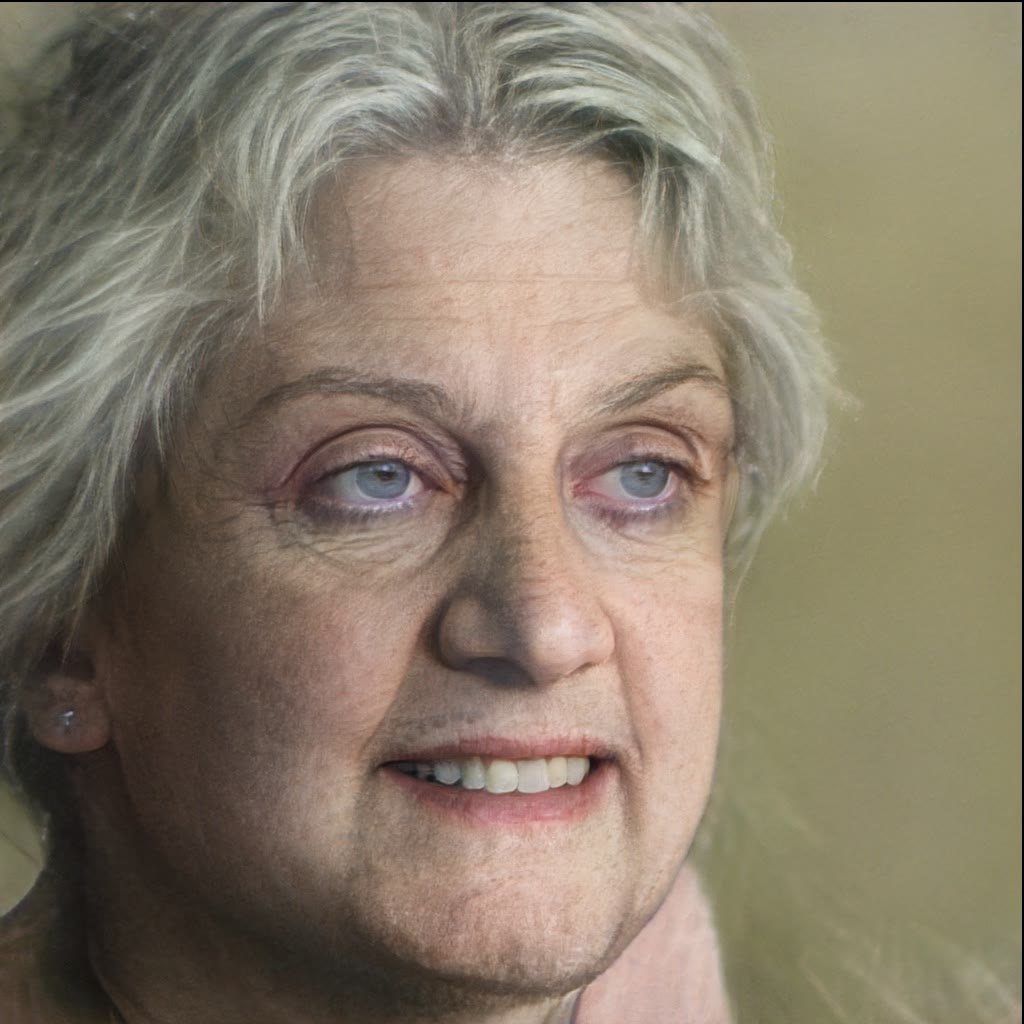} &
		\includegraphics[width=0.06\textwidth]{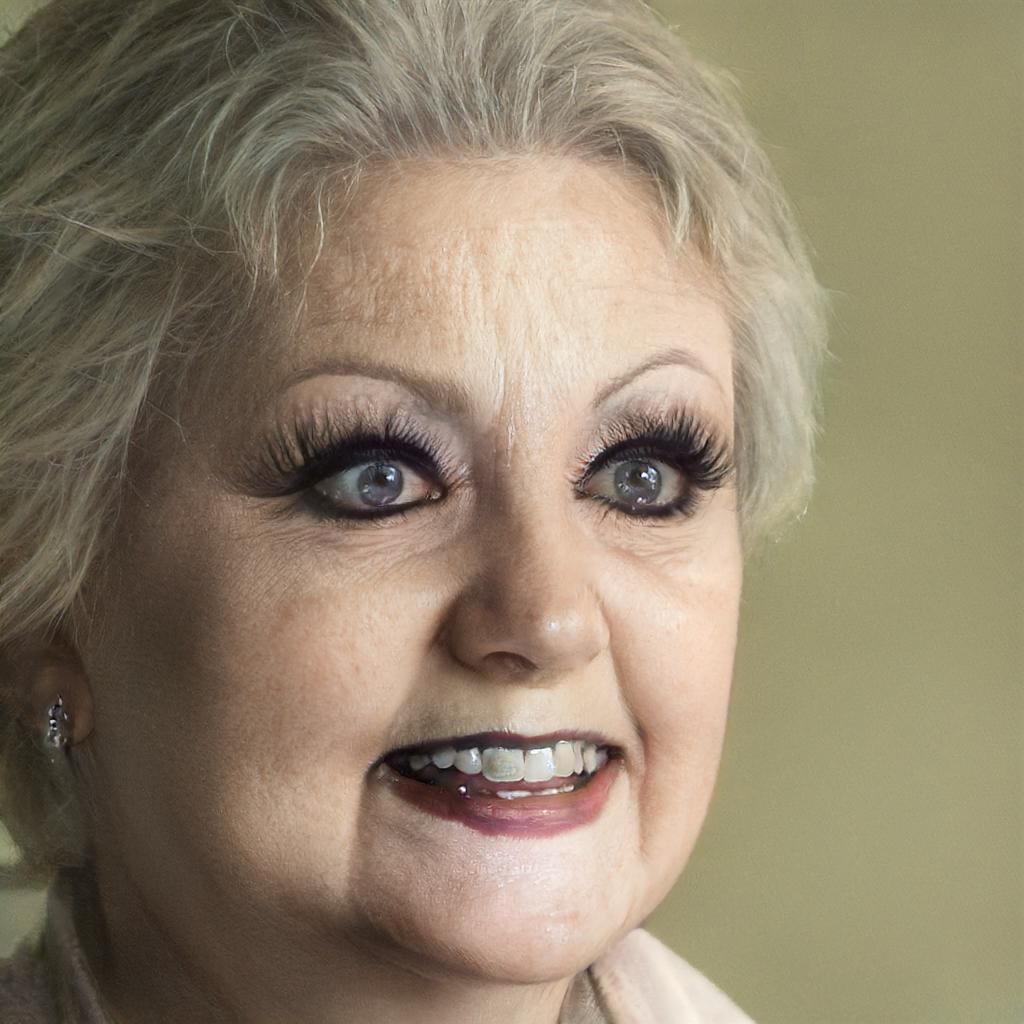} &
		\includegraphics[width=0.06\textwidth]{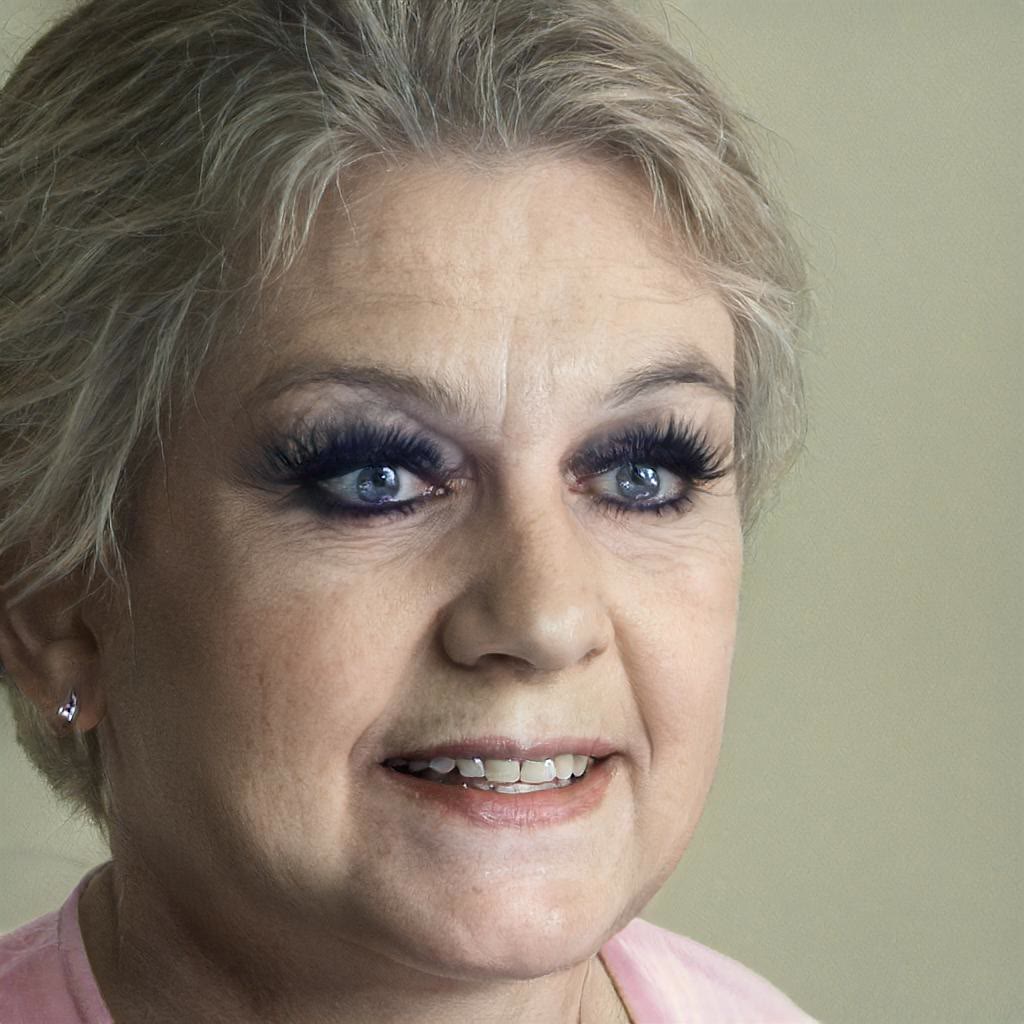} &
		\includegraphics[width=0.06\textwidth]{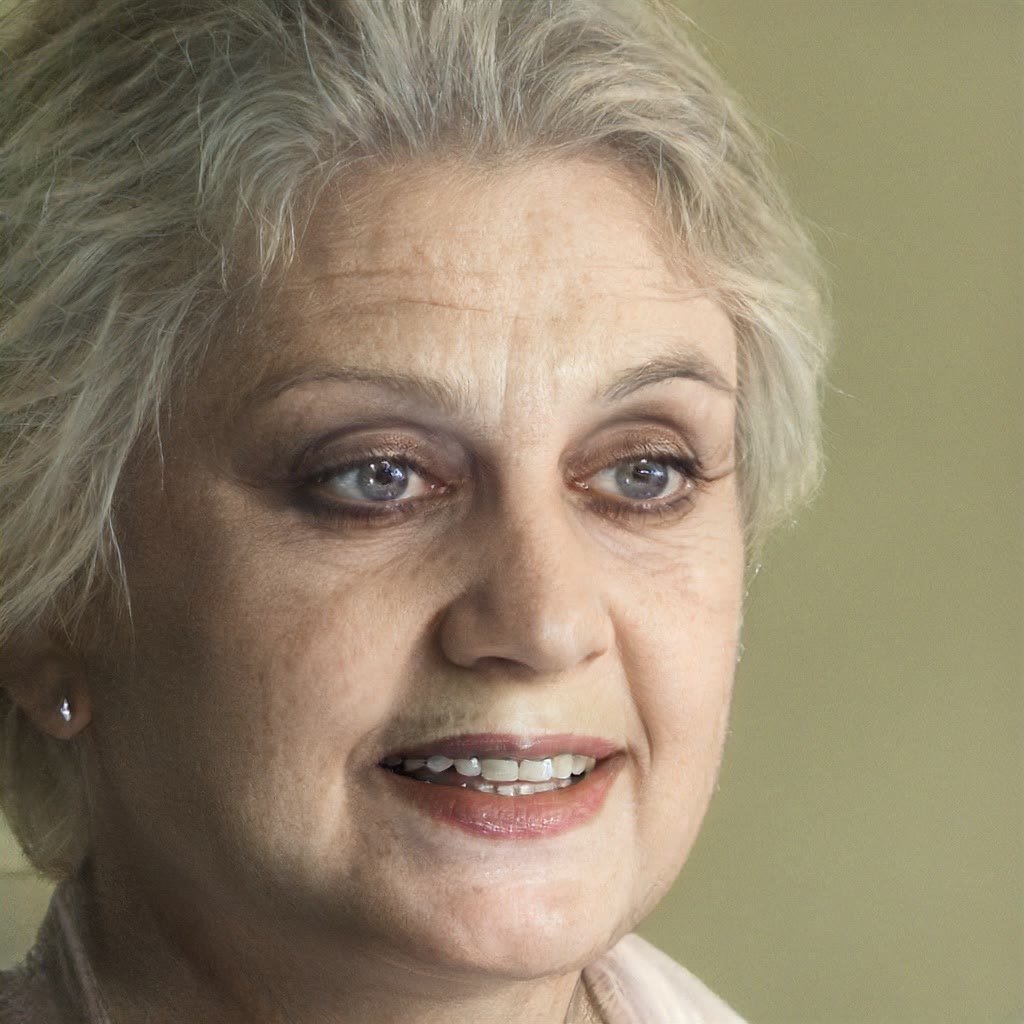} \\
		
    \multicolumn{5}{c}{\ruleline{0.3\textwidth}{Trump}} &\multicolumn{5}{c}{\ruleline{0.3\textwidth}{Mohawk}} &\multicolumn{5}{c}{\ruleline{0.3\textwidth}{Makeup}} 
	\end{tabular}}
	\end{center}
	\caption{\label{fig:global-vs-mapper} Comparison of our method with TediGAN, StyleCLIP-GD and StyleCLIP-LM methods.}
\end{figure*}
 
\subsection{Qualitative Results}
Our method is capable of performing edits in a variety of domains (see Figure \ref{fig:teaser}). Moreover, Figure \ref{fig:final} shows that our method can perform complex edits on the FFHQ dataset, ranging from stylistic edits such as \textit{curly hair} or \textit{beard} to emotional edits such as \textit{relieved} or \textit{excited}. As Figure \ref{fig:faces_animals_met} (a) shows, our method can successfully perform a variety of simple and complicated edits such as \textit{white horse} and \textit{vintage car} with models trained on the LSUN Car and LSUN Horse datasets. Furthermore, Figure \ref{fig:faces_animals_met} (b) shows that our method can handle a variety of simple (e.g., \textit{fur color}, \textit{eyes}) and complex (e.g., \textit{species, expression}) manipulations with models trained on the AFHQ Cat and AFHQ Dog datasets. Finally, Figure \ref{fig:faces_animals_met} (c) shows manipulations with a model trained on the MetFaces dataset.

\subsection{Ablation Study}
\label{sec:ablation}
Our method benefits from several important insights, such as using a small batch size and low-resolution layers to find manipulation directions for speedup. We also  find multiple channels and use identity loss for manipulation quality and effectiveness. In this section, we perform ablation studies to understand the contribution of each component.

\textbf{Low-resolution layers} Our method is able to find directions at low-resolution layers to speedup the process. To test the effectiveness of this approach, we find directions at resolutions $128\times128$, $256\times256$ and $1024\times1024$ (see Figure \ref{fig:combined_singlemulti} (b)). Our results show that $128\times128$ does not encapsulate enough signals to find suitable directions, while $256\times256$ and $1024\times1024$ achieve  comparable manipulations. In terms of computation time, working with  resolution $128\times128$ requires $3s$, while $256\times256$ and $1024\times1024$ require $5s$ and $18s$, respectively. Therefore, we use only the layers up to $256\times256$ resolution to find the desired directions while achieving a  significant speedup, and then use the found directions to apply manipulations at high resolutions such as $512\times512$ or $1024\times1024$.

\textbf{Small batch size} Our method uses only $128$ images to find directions. To understand how the results and computation time change with different batch sizes, we apply our method to a batch of $32$, $128$, and $1024$ images (see Figure \ref{fig:combined_singlemulti} (b)). Our experiment shows that using a batch of $32$ images leads to slightly unnatural manipulations, while a batch of $128$ images can compete with the results obtained with $1024$ images. In terms of computation time, determining direction with $32$ images requires $1.2s$, while $128$ and $1024$ require $5s$ and $35s$, respectively. Thus, using a batch of $128$ images leads to the desired manipulation performance while providing significant speedup. We also note that the directions are not sensitive to the attributes of the images in the batch, and that using the same set of images for any given text prompt works in practice.

\textbf{Single-channel vs. Multi-channel} Next, we perform an ablation study by manipulating images using a single channel. Similar to our multi-channel method, we use a CLIP-based loss:
\begin{align*}
\begin{split}
 \mathcal{L}_{S} = & 1- \left \langle CLIP (G(s + \Delta s)), CLIP (t_1)- CLIP (t_2) \right \rangle 
\end{split} 
\end{align*}
where $s$ corresponds to the latent code of the image and $t_1$ and $t_2$ are the user-specified text inputs. Our experiments show that single-channel yields less stable directions for a single text prompt $t$. To mitigate this effect, we use two text prompts $t_1$ and $t_2$, where $t_1$ contains the target/positive attributes (e.g. \textit{`A man with Mohawk hairstyle'} and $t_2$ the neutral/negative version of the target attribute (e.g. \textit{`A man with hair'}). We also exclude identity loss, since the single-channel based changes are disentangled and affect only a single attribute without changing the identity of the image.

Figure \ref{fig:combined_singlemulti} (a) shows a comparison between single-channel and multi-channel manipulations. We note that the single-channel approach can successfully handle simple manipulations such as \textit{blonde} without affecting other attributes, while it fails at more complex manipulations such as \textit{mohawk}. This is because complex style changes such as \textit{mohawk} cannot be performed by manipulating a single channel, and therefore result in unrelated changes such as adding \textit{eyeglasses}. On the other hand, our multi-channel method can successfully handle the complex \textit{mohawk} manipulation and is able to find a more prominent \textit{blonde} style.

\begin{figure*}[!htp]
\vskip -0.2in
\hspace{1cm} Input \hspace{0.4cm}  StyleMC$\downarrow$ \hspace{0.1cm} StyleMC$\uparrow$  \hspace{0.1cm} GANspace$\downarrow$ \hspace{0.1cm} GANspace$\uparrow$ \hspace{0.2cm} SeFa$\downarrow$ \hspace{0.4cm} SeFa$\uparrow$\hspace{0.35cm} LatentCLR$\downarrow$ \hspace{0.1cm} LatentCLR$\uparrow$
\begin{center}
    \begin{tabular}{ C{3pt} C{.95\textwidth} }
        \rule{0pt}{44pt}\rotatebox{90}{\quad Smile } & \multirow{3}{*}[44pt]{\includegraphics[width=0.95\textwidth]{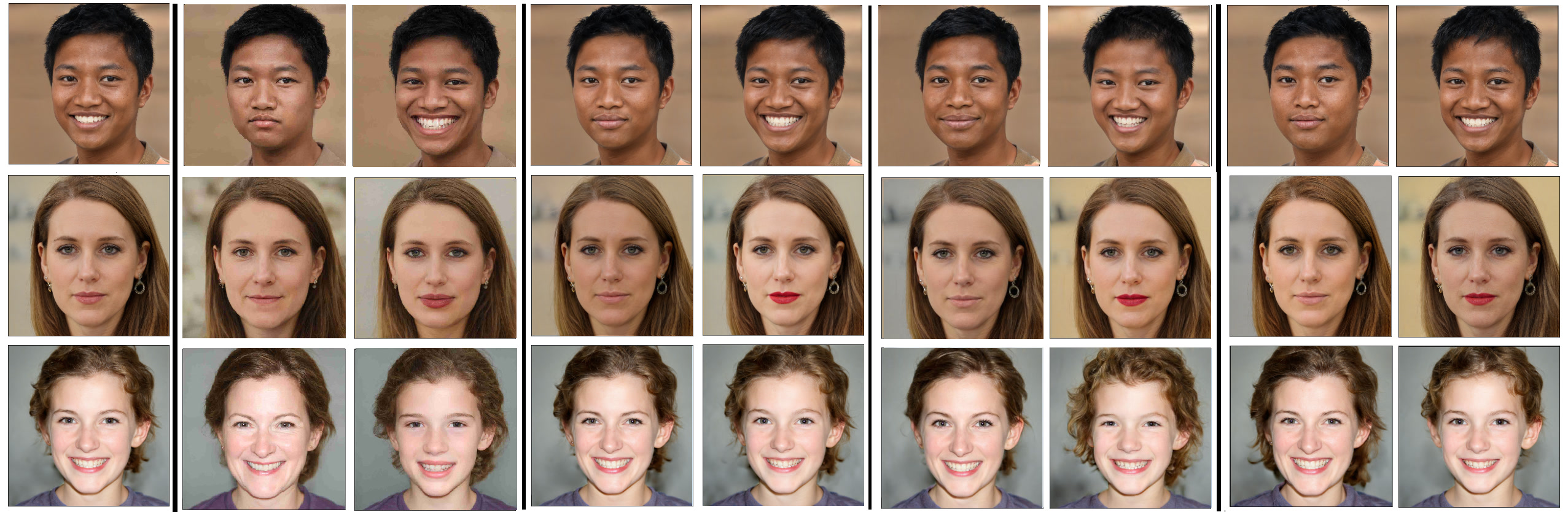}} \\
        \rule{0pt}{44pt}\rotatebox[origin=lc]{90}{Lipstick } & \\
        \rule{0pt}{44pt}\rotatebox[origin=lc]{90}{Young} & \\ 
    \end{tabular}
    \vskip 0.2in
\caption{Comparison with GANspace, SeFa and LatentCLR methods. The leftmost image represents the input image, while images denoted with $\uparrow$ and $\downarrow$ represent edits in the positive or negative direction.}
\vskip -0.1in
\label{fig:full}
\end{center}
\end{figure*}

\subsection{Comparison with Text-Guided Methods}
We compare our method to state-of-the-art text-driven manipulation methods, StyleCLIP-LM, StyleCLIP-GD and TediGAN for manipulating real face images with text prompts. Similar to \cite{patashnik2021styleclip}, we excluded StyleCLIP-LO as it suffers from self-reported unstability issues. For comparison, we used the StyleGAN2 model trained on FFHQ and used a set of complex text prompts such as \textit{`Trump'}, \textit{`Mohawk'} and \textit{`Makeup'} (see Figure \ref{fig:global-vs-mapper}). Most of these prompts require multiple attributes to be changed simultaneously, such as \textit{hair color}, \textit{eyes}, \textit{mouth}, \textit{facial  expression}, and \textit{facial structure}. For the \textit{`Donald Trump'} prompt, StyleCLIP-GD leads to some visual style changes but fails to capture the identity of \textit{Trump}, while the TediGAN method leads to insignificant style changes. In contrast, our method and the StyleCLIP-LM method capture important features specific to the target identity, such as \textit{puffy face}, \textit{blonde hair}, \textit{pink skin}, and \textit{squinty eyes}. For the \textit{`Mohawk'} prompt, StyleCLIP-GD and TediGAN produce minimal changes, while our method produces competitive results with StyleCLIP-LM. For the \textit{`Makeup'} prompt, all methods achieve the target manipulation to some degree.

\begin{table}[t!]
\begin{center}
\resizebox{\columnwidth}{!}{
\begin{tabular}{|c|c|c|c|c|c|}
\hline
Method & Pre- & Training & Inference & Input & Latent\\  & proc & Time & Time & Agnostic &  Space \\
\hline
StyleCLIP-LO & - & - & 98s & no & $\mathcal{W+}$ \\
StyleCLIP-LM & - & 10-12h & 75ms & no  & $\mathcal{W+}$\\
StyleCLIP-GD & 4h & - & 72ms & yes  & $\mathcal{S}$\\
TediGAN & - & 12h+ & 21s & yes & $\mathcal{W+}$ \\
Ours & - & 5s & 65ms & yes & $\mathcal{S}$ \\
\hline
\end{tabular}
} 
\end{center}
\caption{Time comparison of our method with three StyleCLIP methods and TediGAN. }

\label{tab-time}
\end{table}

We compare the computation time for finding the manipulation directions and editing the images in Table \ref{tab-time}. Compared to the other methods, our method does not require prompt engineering and takes significantly less time to find and perform the manipulations. More specifically, the StyleCLIP-LO method solves an optimization problem in $\mathcal{W+}$ space to optimize the latent code directly, and requires several minutes of optimization to edit a single image. Although there is no preprocessing or training time, it is input-dependent and leads to unstable results. The StyleCLIP-LM method operates in $\mathcal{W+}$ space and trains a mapper network for a given text prompt. However, this method is also input dependent and requires 10-12 hours of training. StyleCLIP-GD finds an input-independent global direction instantaneously for any text prompt, but only after 4 hours of preprocessing. The disadvantage of this technique is that it requires hours of computation before any manipulation can be done, and it does not work well for complex and specific attributes, as can be seen in Figure \ref{fig:global-vs-mapper}. The TediGAN method, on the other hand, requires $12+$ hours of preprocessing as it encodes both the image and the text into the latent space and trains an image inversion and encoding module. Nevertheless, TediGAN yields unfavorable results compared to our method and StyleCLIP methods. In contrast, our method is independent of the input and requires 5 seconds\footnote{The time is reported as an average of 100 trainings.} of training to find stable and global manipulation directions.

\subsection{Comparison with Unsupervised Methods}
Next, we compare how the directions found on FFHQ differ across state-of-the-art unsupervised methods such as GANspace, SeFA and LatentCLR. Figure \ref{fig:full}  shows the visual comparison between directions that are commonly found by all methods, including \textit{Smile, Lipstick} and \textit{Young} directions. As can be seen from the visuals, all methods perform similarly and are able to manipulate the images towards the desired attributes.

\section{Limitations and Broader Impact}
\label{sec:limitations}
Our method is based on pre-trained StyleGAN2 and
CLIP models, so the manipulation capabilities strongly depend on the datasets they were trained on. We note that while the joint representation capabilities of CLIP are powerful, they are still limited and may be biased towards certain attributes. Our framework has similar concerns as any other image synthesis tool that can be used for malicious purposes, as discussed in \cite{korshunov2018deepfakes}.

\section{Conclusion}
\label{sec:conclusion}
We introduced a fast and efficient method for text-guided image generation and manipulation. Unlike previous work that requires hours of preprocessing or training, our method requires only a few seconds per text prompt to find a global manipulation direction and generates results that are on par with state-of-the-art models such as StyleCLIP in terms of quality. In our experiments, we have shown that our method can be used to apply a variety of manipulations ranging from simple style changes such as \textit{hair color} to complex changes such as \textit{gender, personal identity, species} and provide control over the strength and direction of the manipulations.

\noindent\textbf{Acknowledgments}
This publication has been produced benefiting from the 2232 International Fellowship for Outstanding Researchers Program of TUBITAK (Project No: 118c321). We also acknowledge the support of NVIDIA  with the donation of the TITAN RTX GPU and GCP credits from Google. 
\newpage

{\small
\bibliographystyle{ieee_fullname}
\bibliography{wacv}
}

\appendix
\begin{appendices}
\section{Identity Loss}
\label{appdx:identity_loss}
\begin{figure}[!h]
\begin{center}
\centerline{\includegraphics[width=\columnwidth]{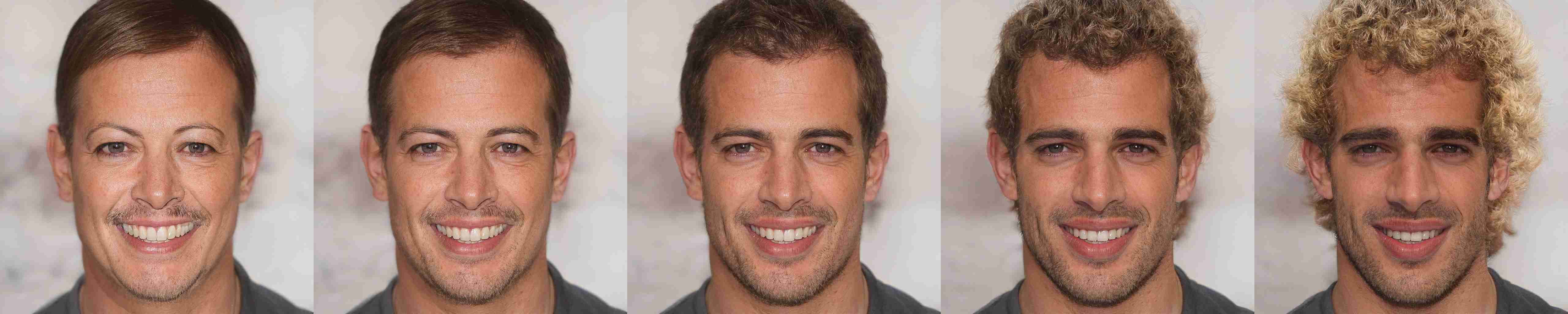}}
\centerline{\includegraphics[width=\columnwidth]{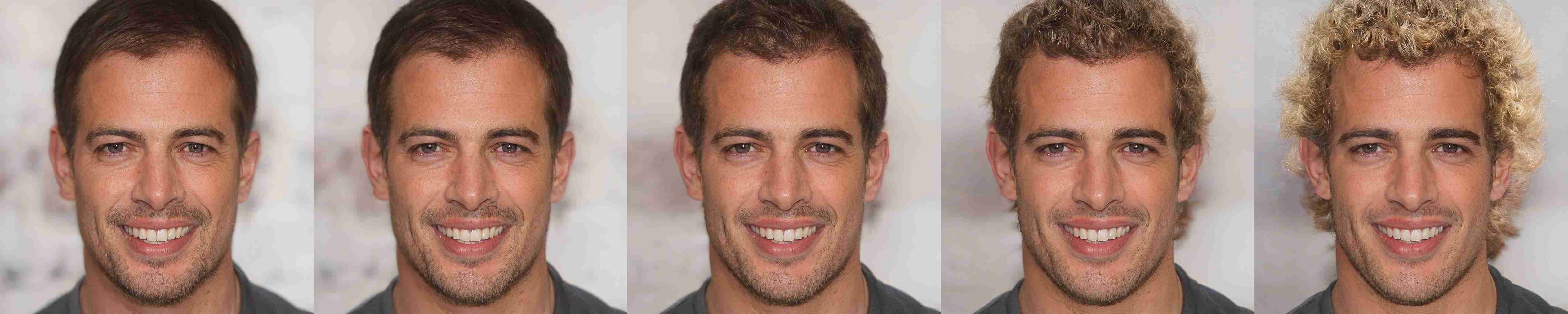}}
\centerline{\includegraphics[width=\columnwidth]{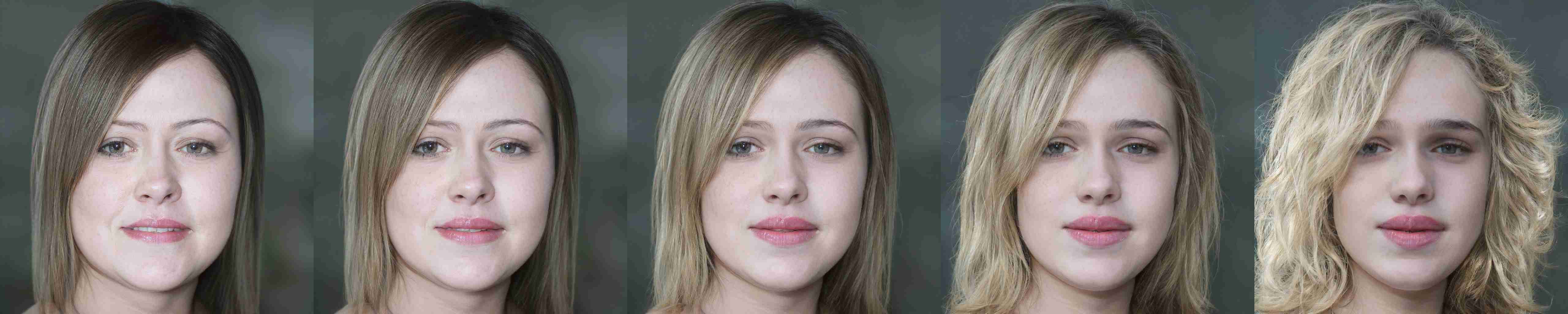}}
\centerline{\includegraphics[width=\columnwidth]{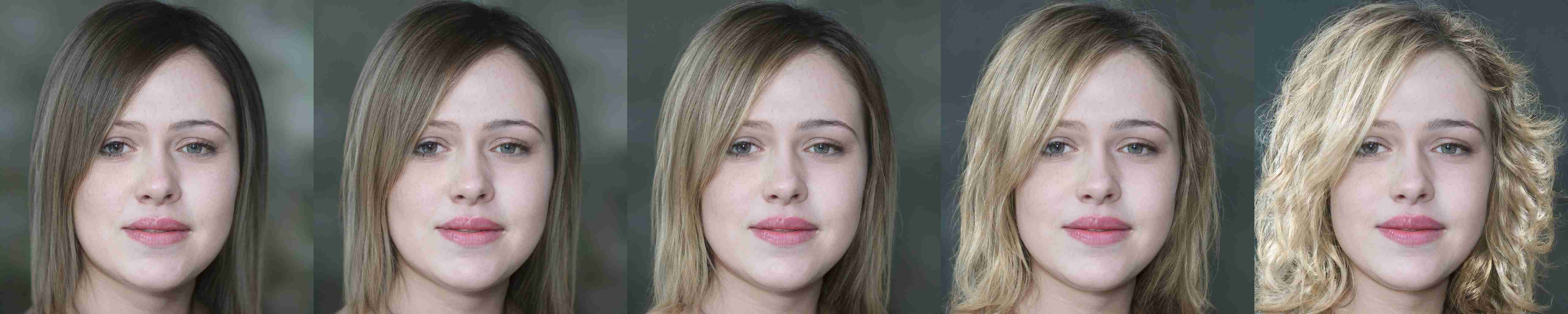}}
\centerline{\includegraphics[width=\columnwidth]{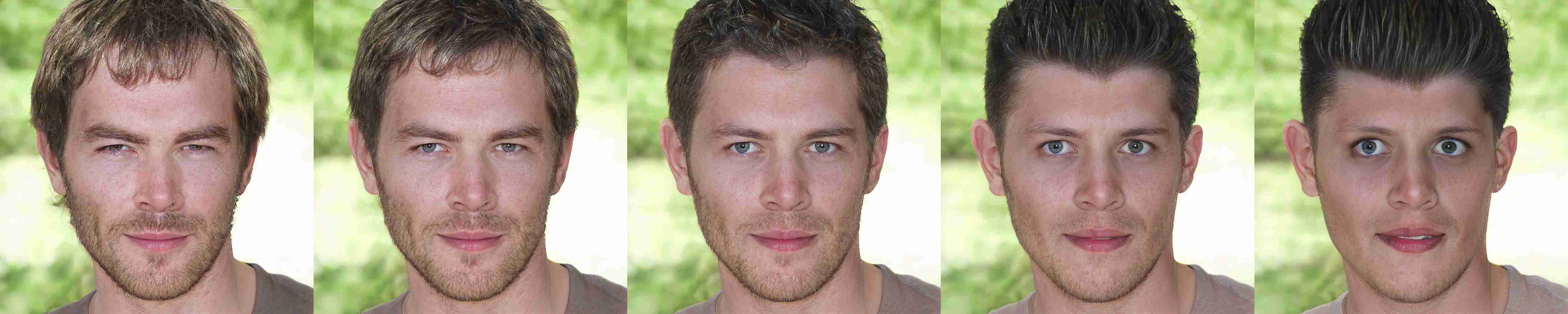}}
\centerline{\includegraphics[width=\columnwidth]{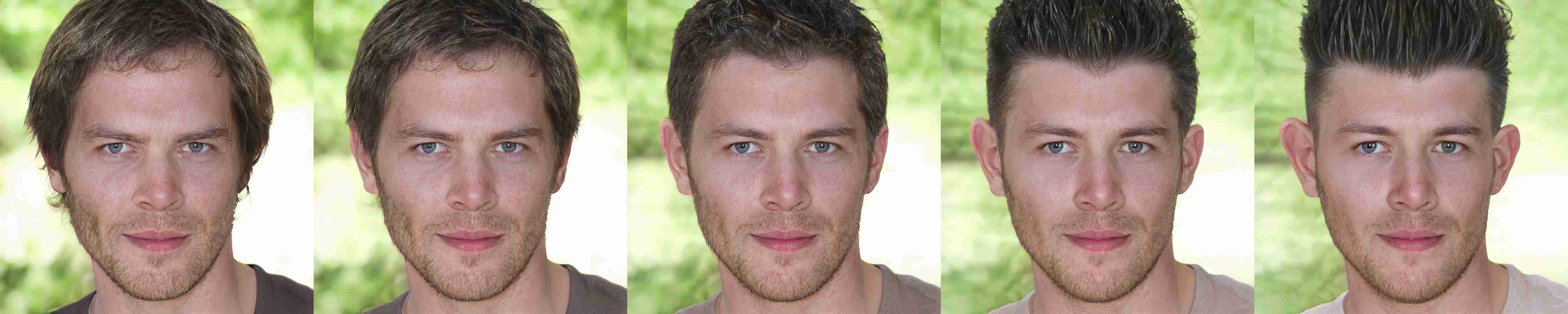}}
\centerline{\includegraphics[width=\columnwidth]{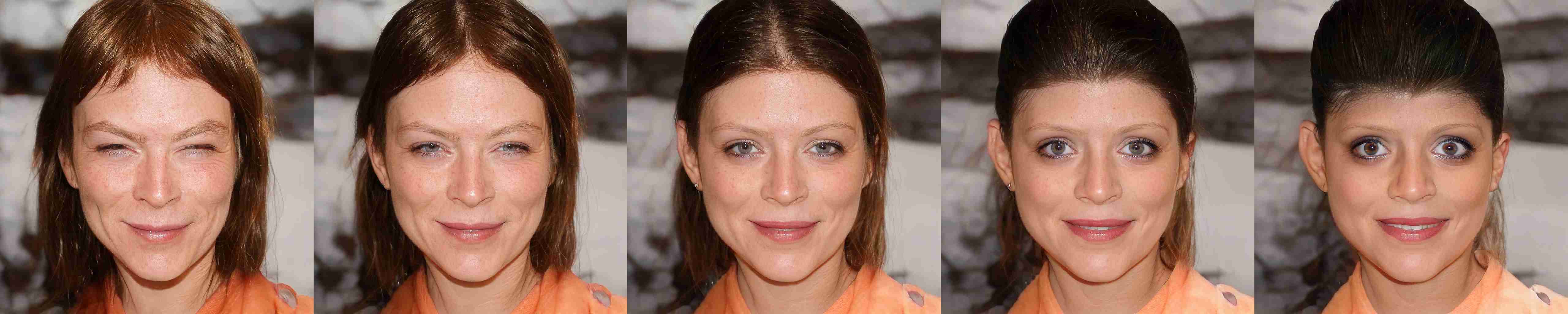}}
\centerline{\includegraphics[width=\columnwidth]{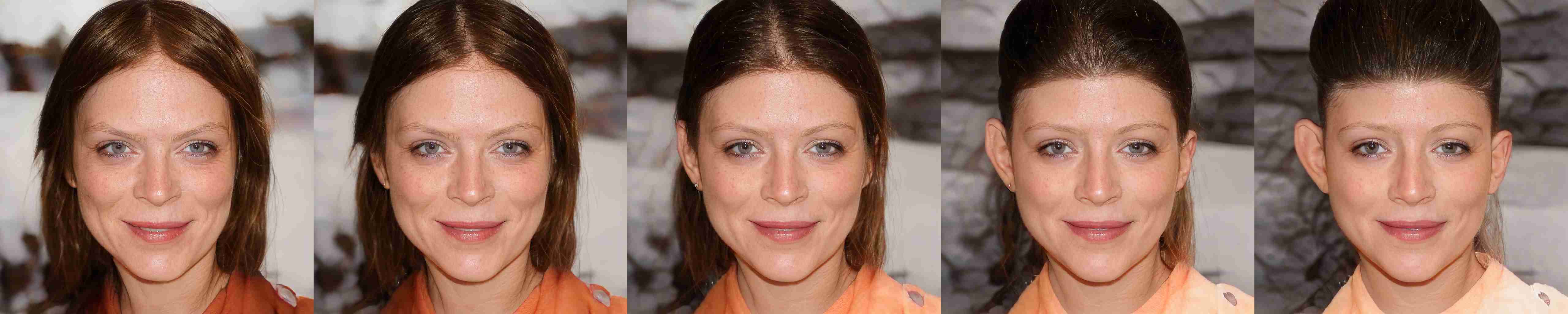}}
\caption{Identity loss ablation study for our method. Results without and with identity loss for “curly hair” and “hitop fade hair” manipulations are given in alternating rows. As can be seen, the identity of individual in the input image is not preserved when identity loss is not used.}
\label{fig:identity}
\end{center}
\vskip -1in
\end{figure}

To show the significance of the identity loss for preserving the identity of the person in the input image, we perform the following experiment. We generate random images using StyleGAN2 \cite{StyleGAN} trained on the FFHQ dataset. For each image, we perform a manipulation with and without the identity loss, where the coefficient of $\mathcal{L_{ID}}$ is set to $ \lambda_{ID} = 10$ and $\lambda_{ID}= 0$ respectively. We observe that our method fails to preserve the identity of the person when the identity loss is omitted. Moreover, we observe that identity loss is highly effective at preventing changes to irrelevant features. The results are shown in Figure \ref{fig:identity}.

\newpage

\begin{strip}\centering
\begin{minipage}{0.4\columnwidth}
\centering
\footnotesize{
A bird with \textbf{black eye rings} and a \textbf{black bill}, with a \textbf{brown crown} and a \textbf{brown belly}.
}
\end{minipage}
\begin{minipage}{0.4\columnwidth}
\centering
\footnotesize{
A bird with a \textbf{white belly}, a \textbf{white crown}, and \textbf{white wings}.
}
\end{minipage}
\begin{minipage}{0.4\columnwidth}
\centering
\footnotesize{
A bird is \textbf{orange} and \textbf{black} in colour, with a \textbf{blue crown} and \textbf{black eye rings}.
}
\end{minipage}
\begin{minipage}{0.4\columnwidth}
\centering
\footnotesize{
The bird has a \textbf{black head} and a \textbf{yellow belly}.
}
\end{minipage}
\begin{minipage}{0.4\columnwidth}
\centering
\footnotesize{
A \textbf{red} bird has a \textbf{yellow head} and a \textbf{yellow belly} with a \textbf{red crown}.
}
\end{minipage}
\begin{minipage}{2\columnwidth}
\centerline{\includegraphics[width=1.05\columnwidth]{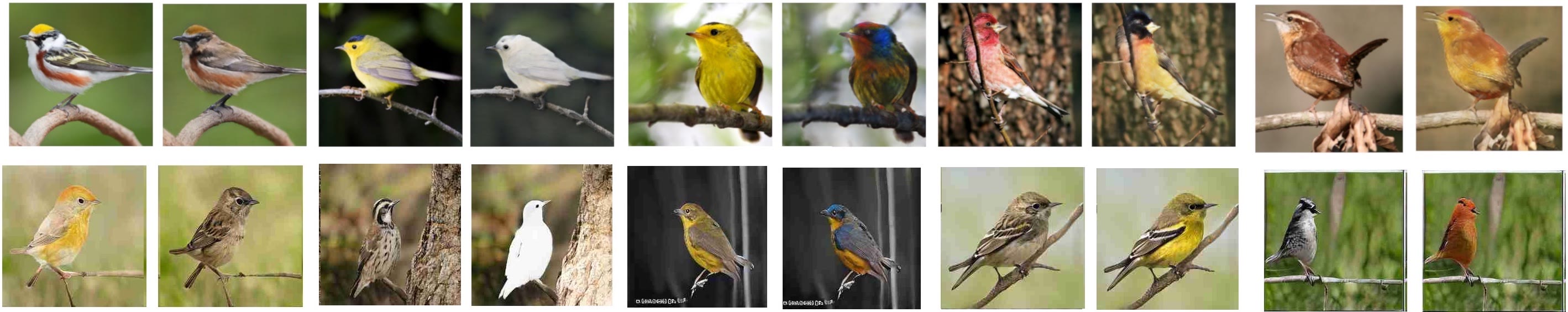}}
\hspace{0cm} Input \hspace{0.5cm}  Edited \hspace{0.6cm} Input \hspace{0.55cm}  Edited \hspace{0.65cm} Input \hspace{0.6cm}  Edited \hspace{0.7cm} Input \hspace{0.65cm}  Edited \hspace{0.65cm} Input \hspace{0.6cm}  Edited
\vspace*{0.2cm}
\end{minipage}
\captionof{figure}{Comparison with \cite{li2020lightweight} (in the first row) and our method (in the second row). The text used for the prompts are shown on top of each image. The images for the compared method are borrowed from \cite{li2020lightweight}. As can be seen from the edited images, our method achieves comparable performance to \cite{li2020lightweight} and able to manipulate bird images toward desired attributes.}
\label{fig:lightweight}
\end{strip}

\section{Additional Comparisons}
\label{appdx:additional_comparisons}

We provide additional comparisons to two text-guided image manipulation methods to demonstrate the effectiveness of our method. Figure \ref{fig:lightweight} shows the comparison between our method and \cite{li2020lightweight}. Since our method uses StyleGAN2, we trained a StyleGAN2-ADA model on the CUB Bird dataset \cite{wah2011caltech}. However, it was not possible to generate the same input images for a direct comparison\footnote{Another approach might be inverting the same input images as \cite{li2020lightweight} but it  uses a different GAN model for which a CUB Bird encoder is not available.}. Therefore, we use the same text prompt to perform manipulations on different bird images. As can be seen from Figure \ref{fig:lightweight}, our method achieves comparable performance to \cite{li2020lightweight} and is able to manipulate bird images toward desired attributes.

\begin{figure}[!t]
\hspace{0.15cm} \small{Input} \hspace{0.15cm}  \small{Edited} \hspace{0.15cm} \small{Edited} \hspace{0.3cm}  \small{Input} \hspace{0.25cm} \small{Edited} \hspace{0.3cm}  \small{Edited} 

\centerline{\includegraphics[width=\columnwidth]{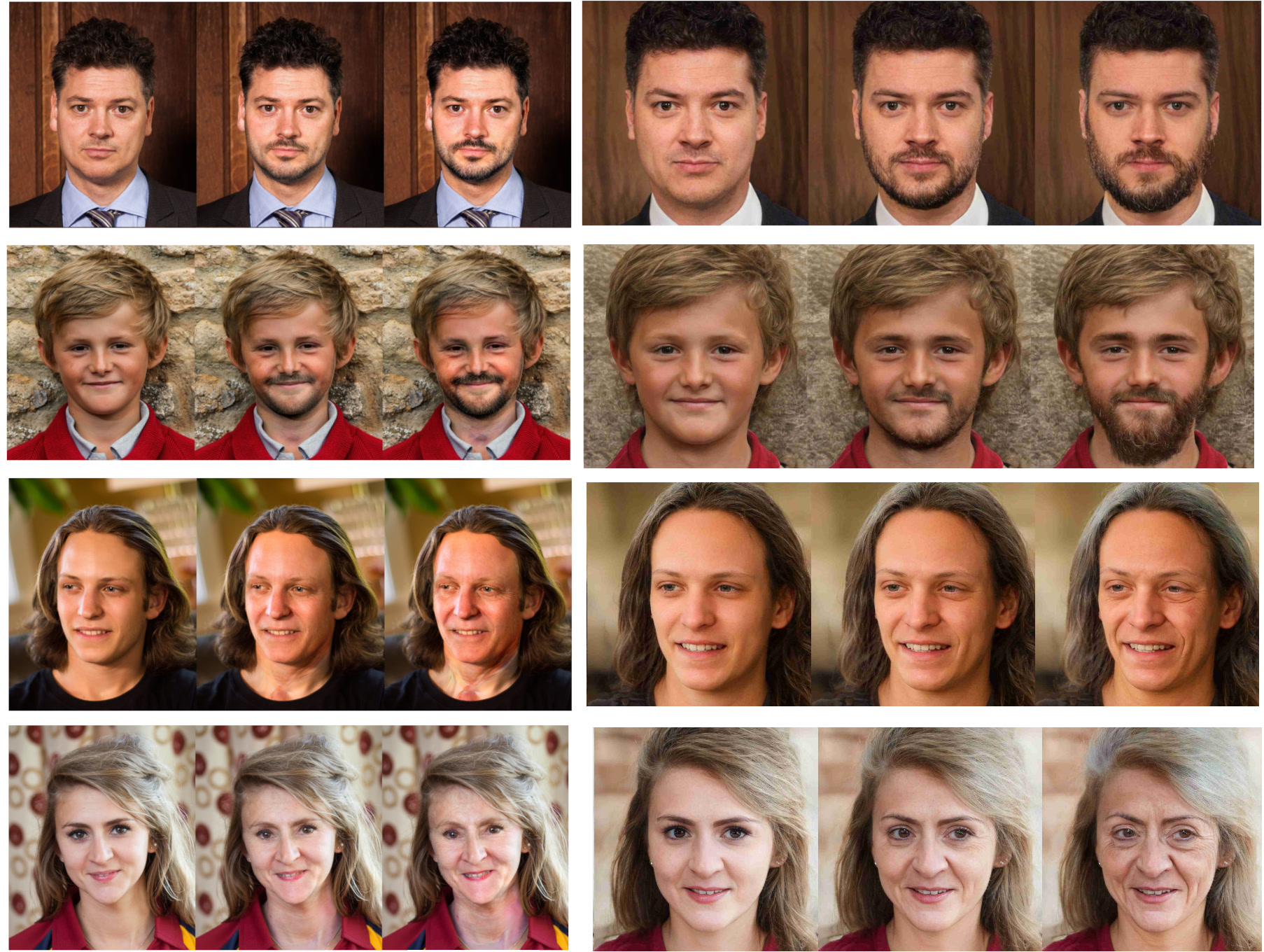}}
\centering{\hspace{-0.2cm} (a) \hspace{3cm} (b) \hspace{0.8cm}}
\vskip 0.1in
\caption{Comparison with \cite{upchurch2017deep} (denoted with (a)). The top two rows show \textit{addition of facial hair}, and bottom two rows shows \textit{aging}. As can be seen from the results, our method achieves a comparable performance to \cite{upchurch2017deep}. Original images denoted with (a) are borrowed from \cite{upchurch2017deep}.}
\label{fig:upchurch}
\end{figure}

Next, we compare our method with \cite{upchurch2017deep}. Since their method uses a different GAN model than ours, it was not possible to generate the same input images for a direct comparison. Therefore, we use the following strategy: we borrowed input images from \cite{upchurch2017deep}, and used an e4e encoder to invert the images. We perform text-based manipulation with our method and present a comparison of results in Figure \ref{fig:upchurch}. As can be seen in Figure \ref{fig:upchurch}, our method achieves a comparable performance to \cite{upchurch2017deep}. Note that some details are lost between two input images due to the nature of the inversion operation, and the inverted images are zoomed-in and cropped compared to the original images.

\section{Additional Results}
\label{appdx:additional_results}

\begin{figure}[!t]
    \vskip 0.2in
    \begin{center}
        \begin{tabular}{ C{2pt} L{0.9\columnwidth} }
            \rotatebox[origin=lc]{90}{Angry} & \includegraphics[width=.9\columnwidth]{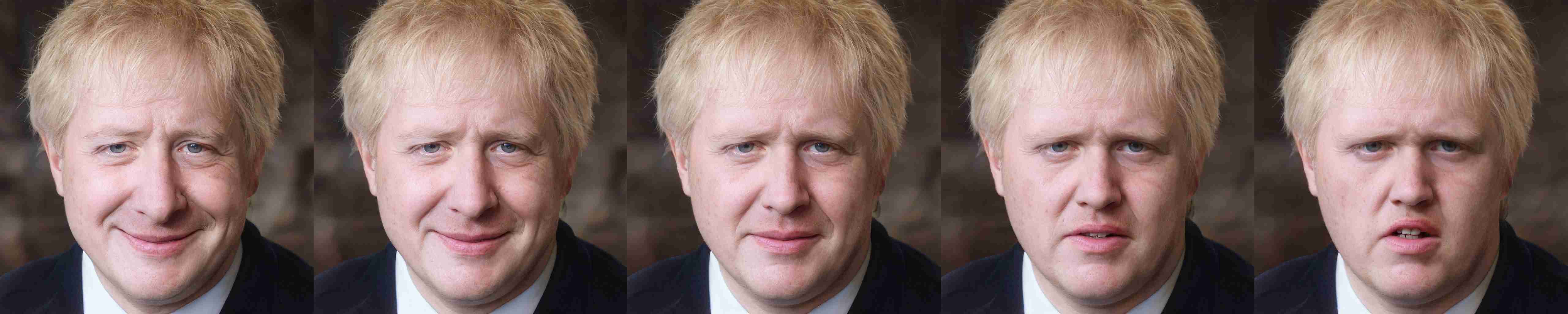} \\
            \rotatebox[origin=lc]{90}{Happy} & \includegraphics[width=.9\columnwidth]{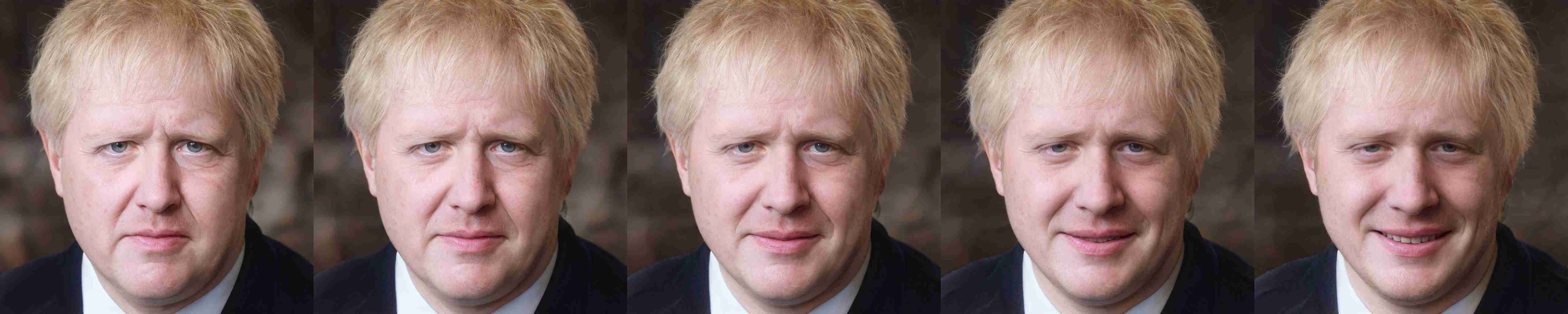} \\
            \rotatebox[origin=lc]{90}{Sad} & \includegraphics[width=.9\columnwidth]{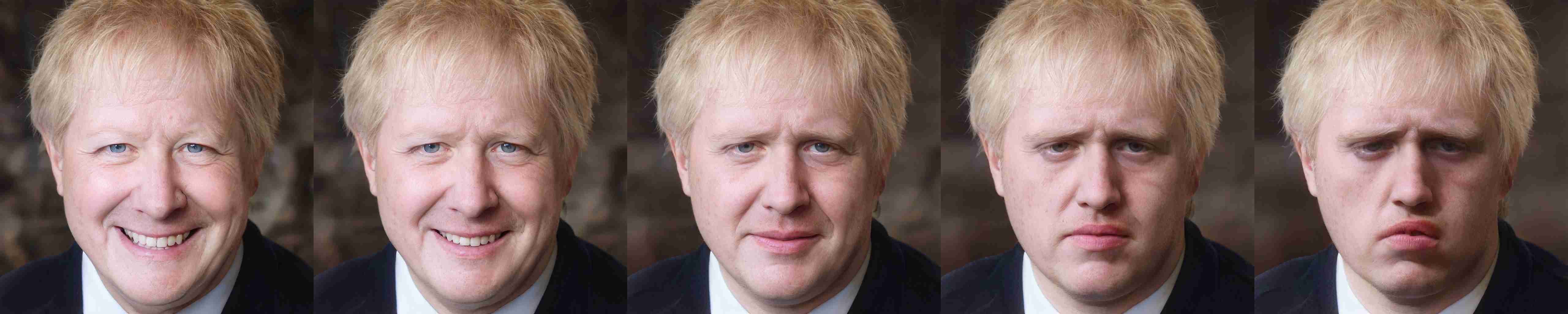} \\
            \rotatebox[origin=lc]{90}{Surprised} & \includegraphics[width=.9\columnwidth]{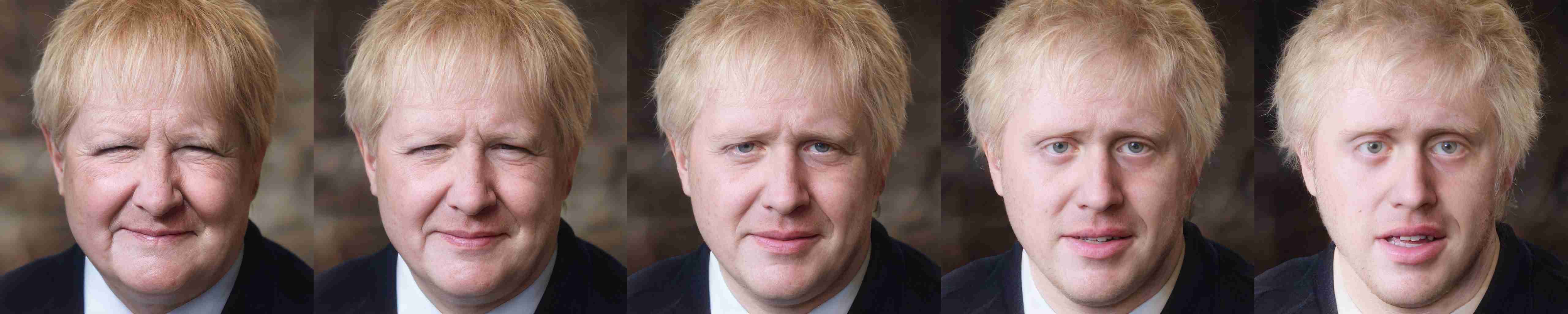} \\
            \rotatebox[origin=lc]{90}{Angry} & \includegraphics[width=.9\columnwidth]{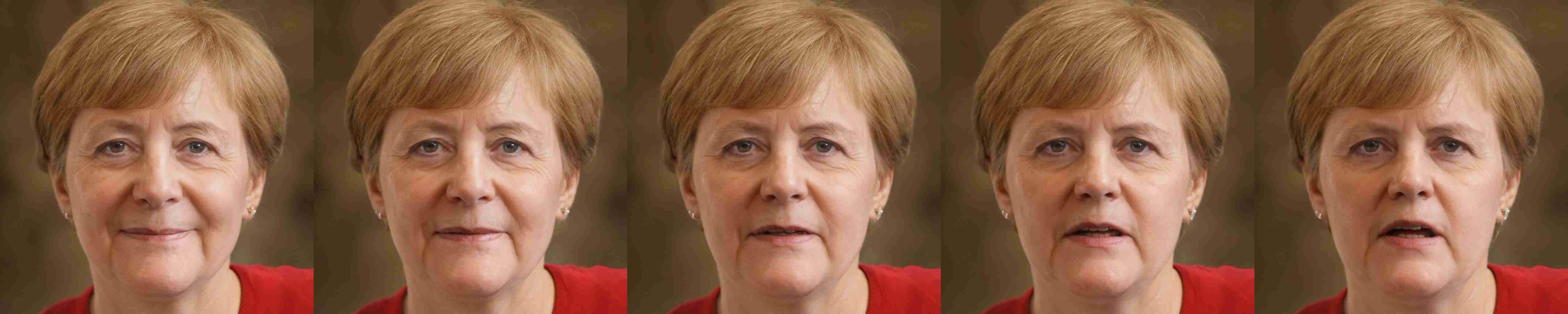} \\
            \rotatebox[origin=lc]{90}{Happy} & \includegraphics[width=.9\columnwidth]{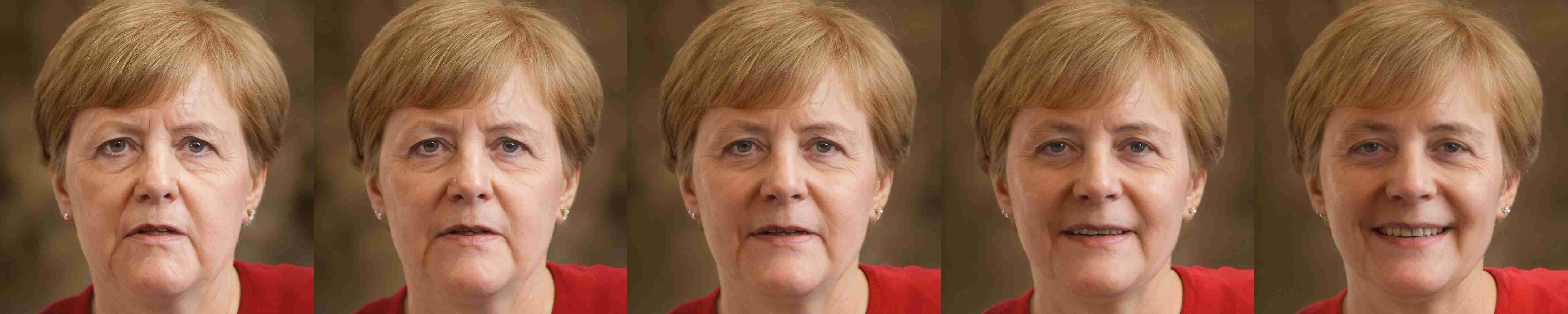} \\
            \rotatebox[origin=lc]{90}{Sad} & \includegraphics[width=.9\columnwidth]{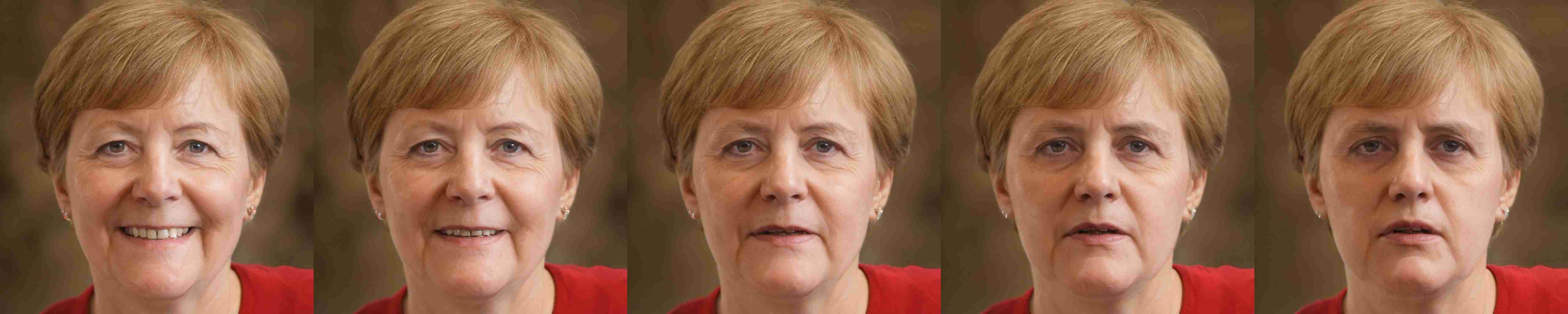} \\
            \rotatebox[origin=lc]{90}{Surprised} & \includegraphics[width=.9\columnwidth]{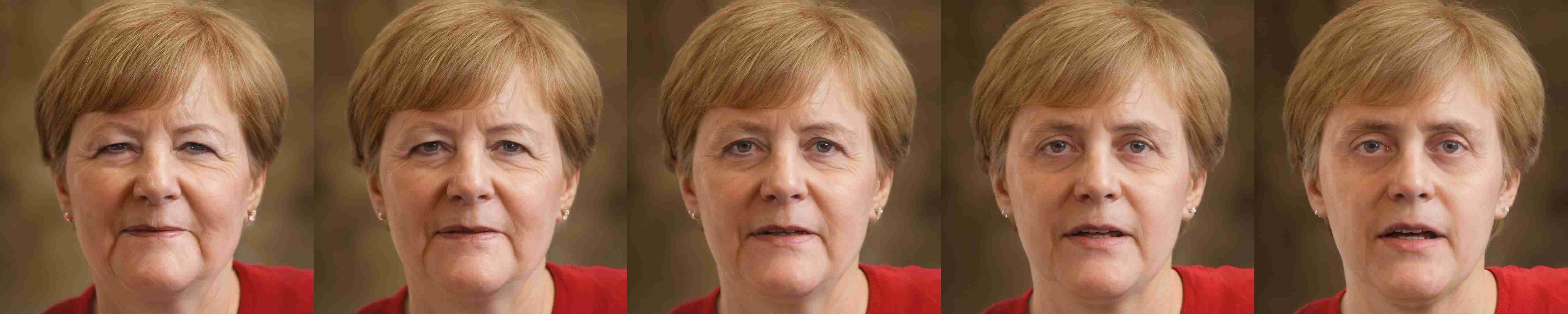} \\
        \end{tabular}
 
        \caption{Edits of inverted celebrity images with the multi-channel latent optimization method using the "angry", "happy", "sad", and "surprised" directions. }
        \label{fig:emotion_celebs}
    \end{center}
    \vskip -0.2in
\end{figure}

\begin{figure}[!t]
\begin{center}
\centerline{\includegraphics[width=\columnwidth]{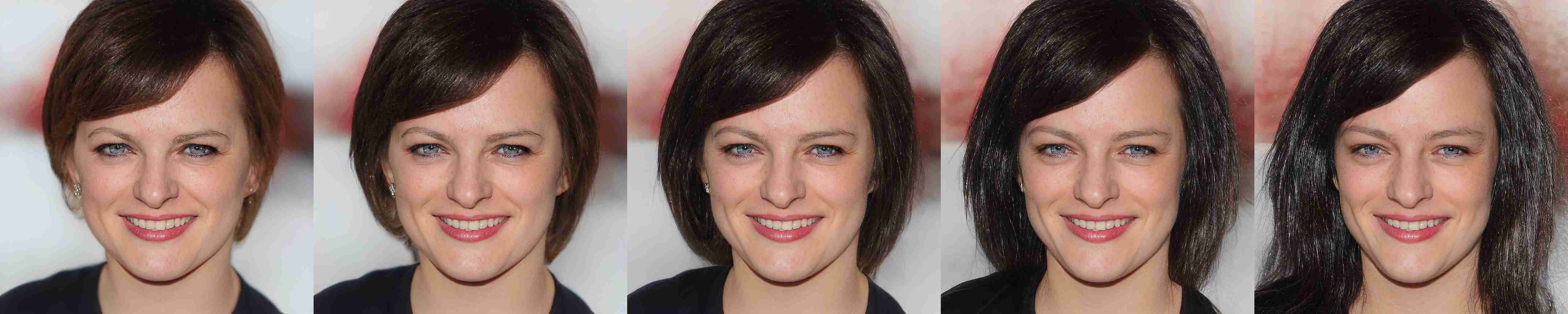}}
\centerline{\includegraphics[width=\columnwidth]{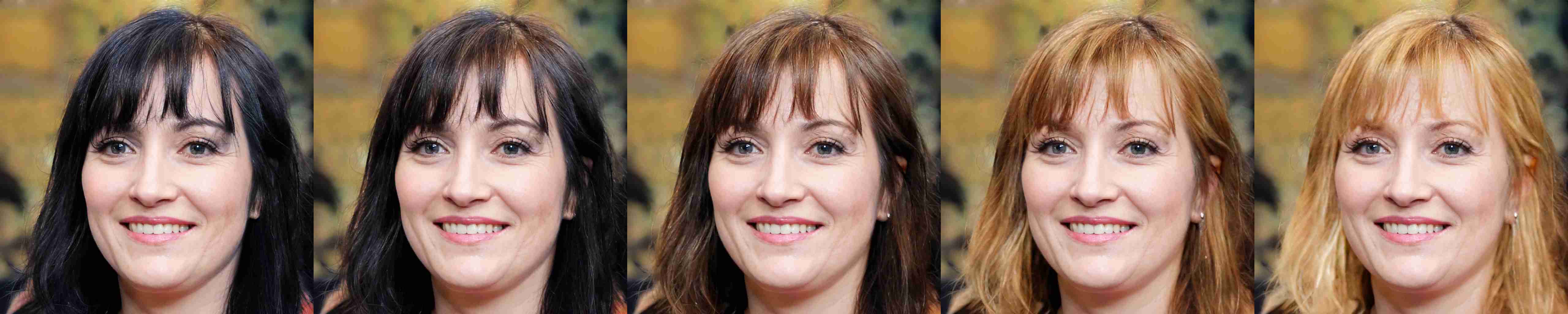}}
\centerline{\includegraphics[width=\columnwidth]{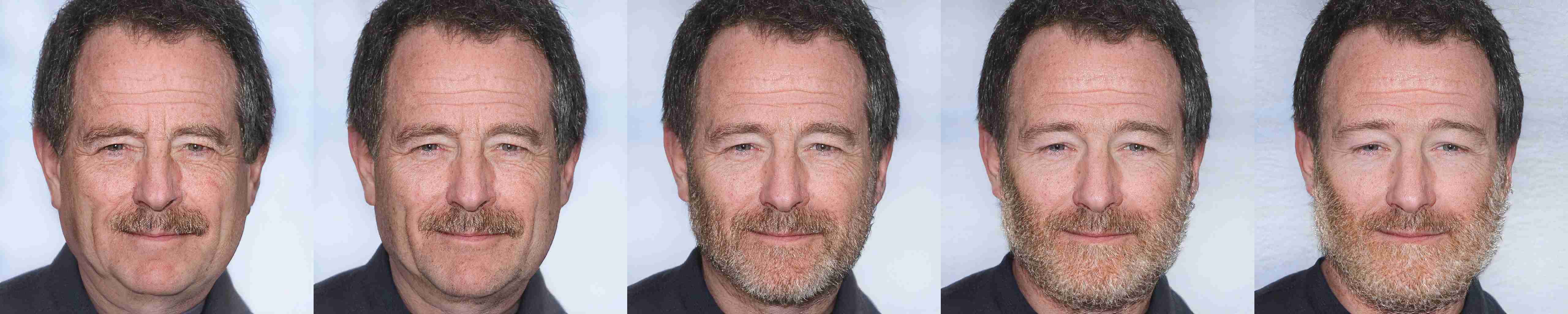}}
\centerline{\includegraphics[width=\columnwidth]{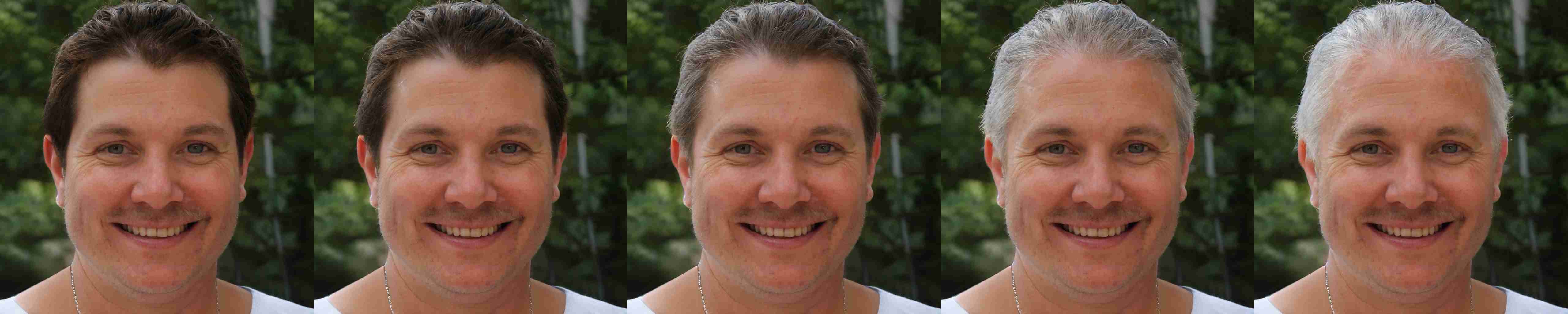}}
\centerline{\includegraphics[width=\columnwidth]{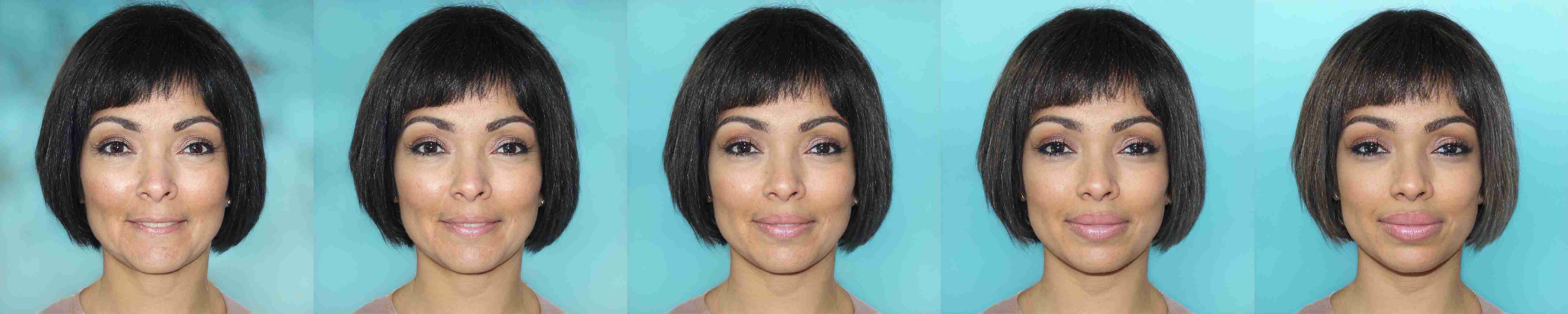}}
\centerline{\includegraphics[width=\columnwidth]{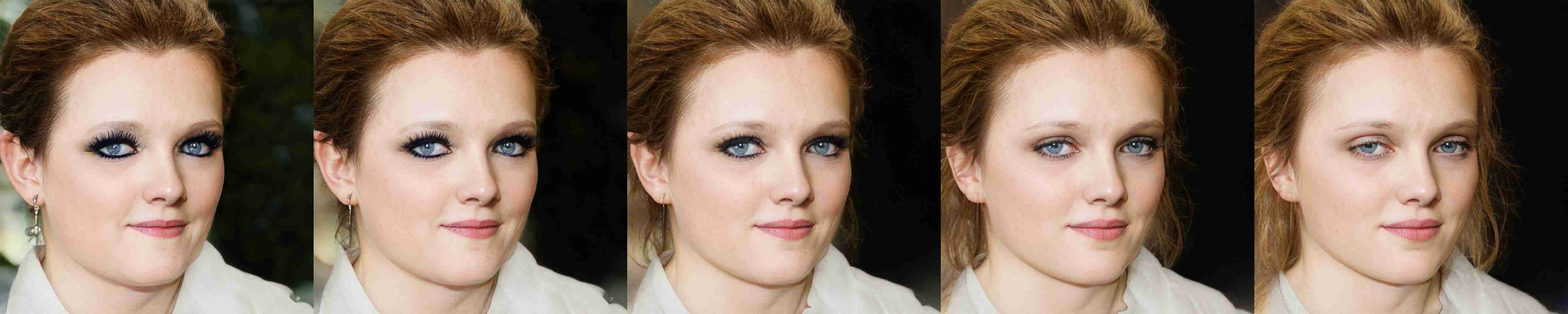}}
\centerline{\includegraphics[width=\columnwidth]{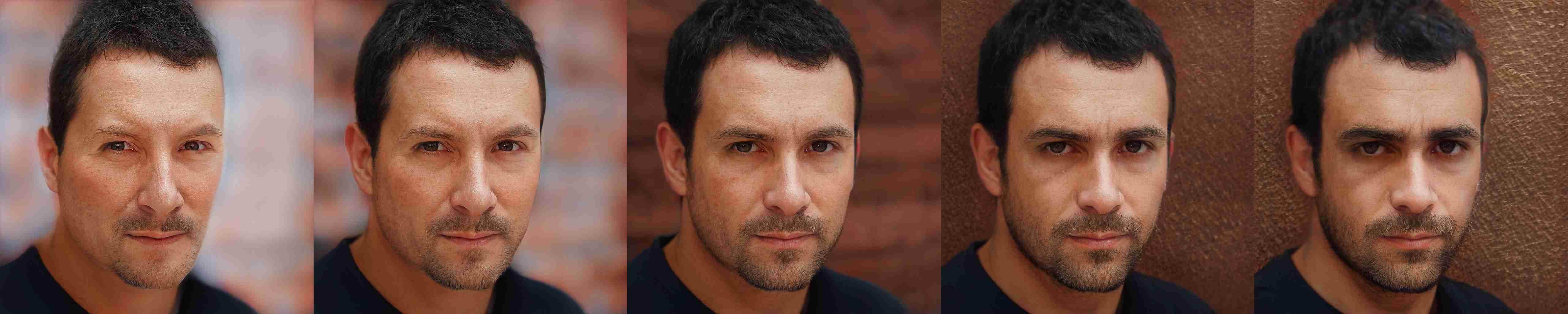}}
\centerline{\includegraphics[width=\columnwidth]{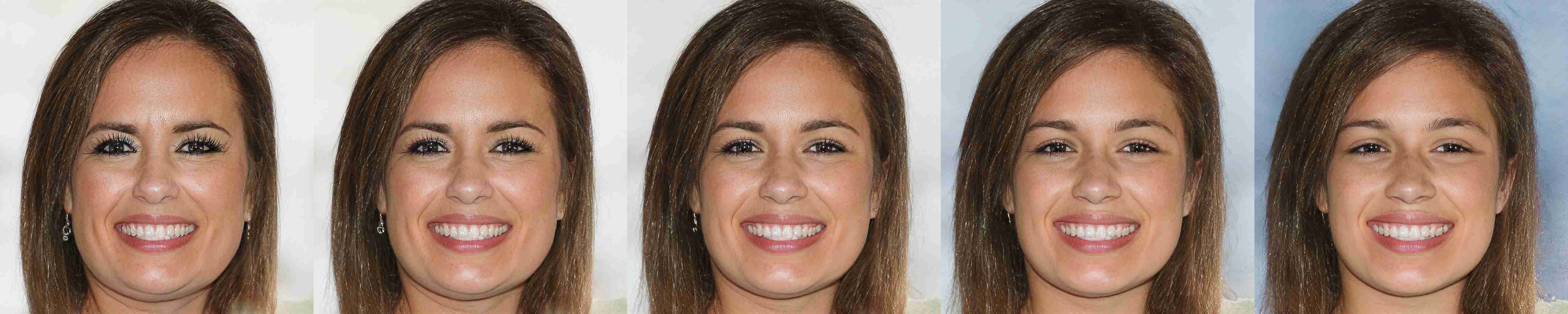}}

\caption{Manipulation results of our method on randomly generated images. Edits are performed using StyleGAN2 pre-trained on the FFHQ dataset. From top to bottom: “long hair”, “blonde”, “beard”, “gray hair”, “bigger lips”, “without makeup”, “serious” and “younger” manipulations.}
\label{fig:people}
\end{center}
\end{figure}

\begin{figure*}[]
\begin{center}
Original  \hspace{0.3cm}  Beard  \hspace{0.25cm}   Long hair \hspace{0.2cm}   Happy  \hspace{0.15cm}   Curly hair\hspace{0.15cm}     Frowning   \hspace{0.05cm}  Blonde hair   \hspace{0.05cm}  Tanned  \hspace{0.15cm}   Relieved   \hspace{0.15cm}  Excited
\includegraphics[width=2\columnwidth]{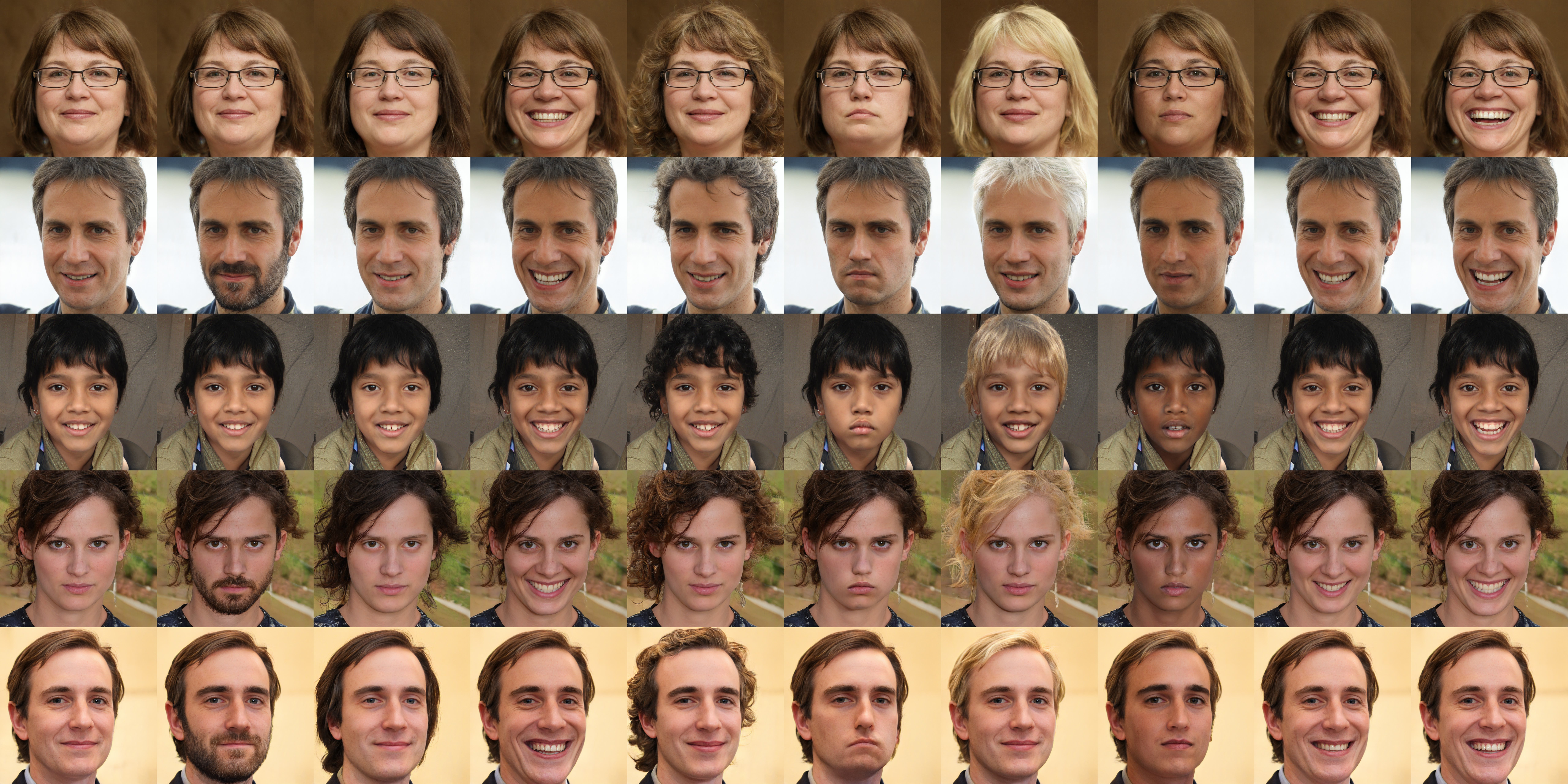}
\caption{A variety of manipulations on images randomly generated with StyleGAN2 FFHQ model. The text prompt used for the manipulation is above each column.}
\label{fig:final2}
\end{center}
\vskip -0.2in
\end{figure*}

In this section, we provide additional results to those presented in the paper. We begin with a variety of image manipulations on randomly generated images using our method and StyleGAN2 pre-trained on the FFHQ dataset. As shown in Figure \ref{fig:emotion_celebs}, our method can perform a wide variety of complex edits such as hair style and expression manipulations on real images taken from the FFHQ dataset successfully. As shown in Figure \ref{fig:emotion_celebs}, moving along the found direction introduces or emphasizes the target attribute (e.g., sad), while steps in the opposite direction yields the opposite results (e.g., happy). Next, we perform manipulations on randomly generated face images. As shown in Figure \ref{fig:people} and \ref{fig:final2}, our method can successfully perform a variety of complex hair/beard  style and expression manipulations on random images. However, we note that our method can not make extreme edits, such as adding beard to children. 

\begin{figure}[!t]
\vskip 0.2in
\begin{center}
\centerline{\includegraphics[width=\columnwidth]{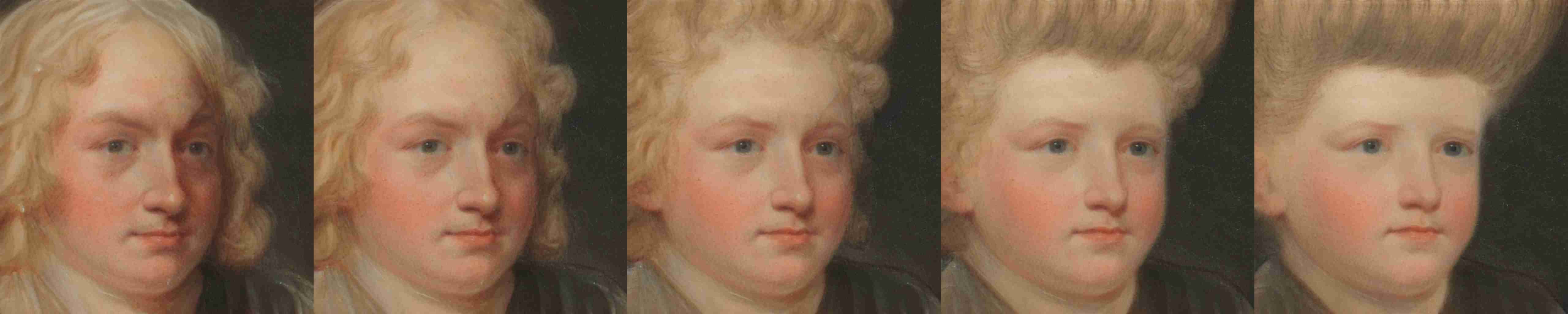}}
\centerline{\includegraphics[width=\columnwidth]{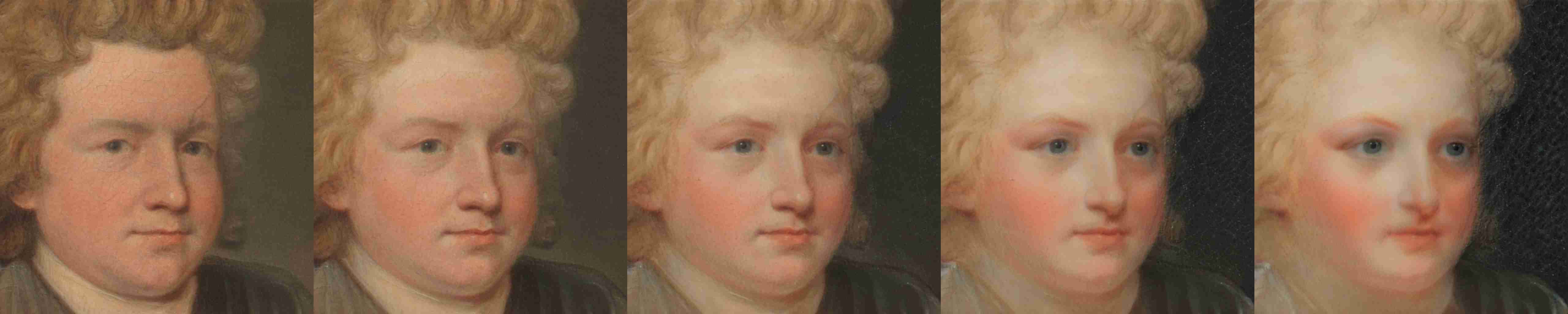}}
\centerline{\includegraphics[width=\columnwidth]{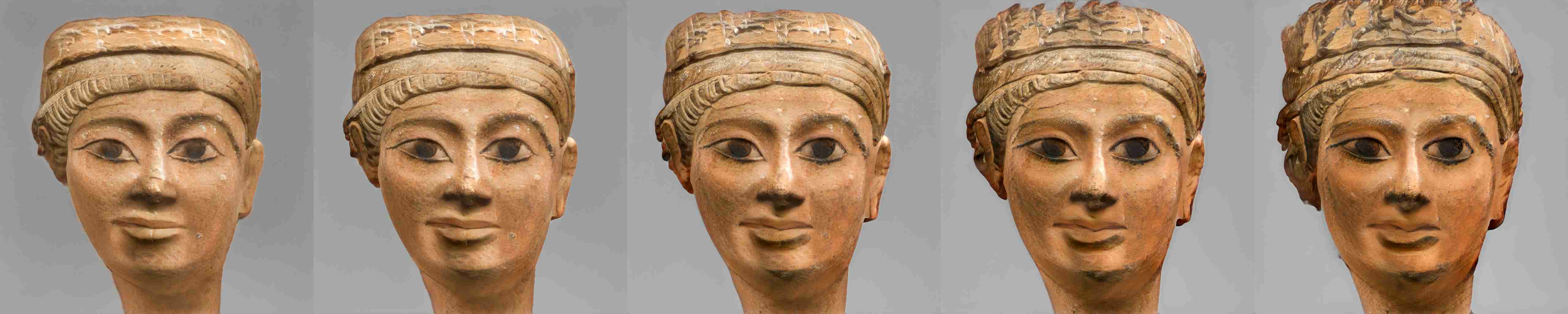}}
\centerline{\includegraphics[width=\columnwidth]{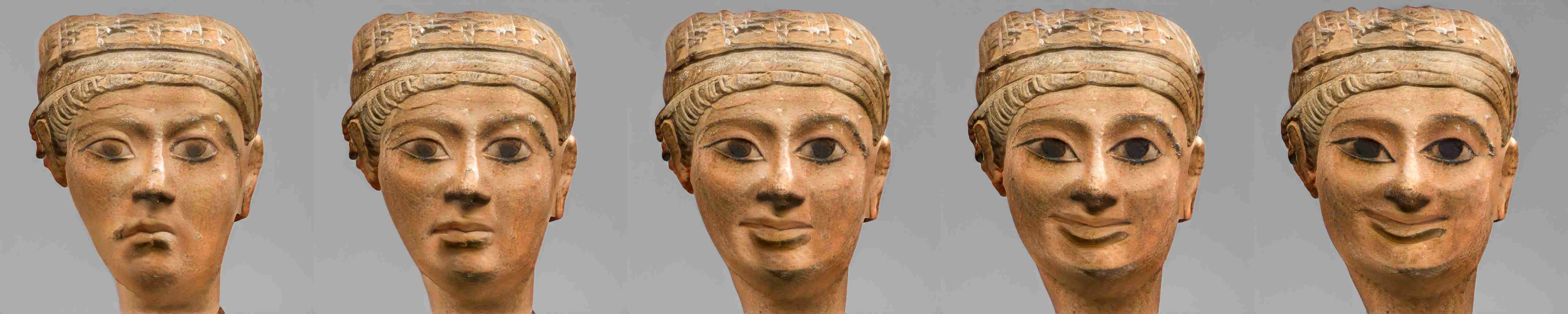}}
\centerline{\includegraphics[width=\columnwidth]{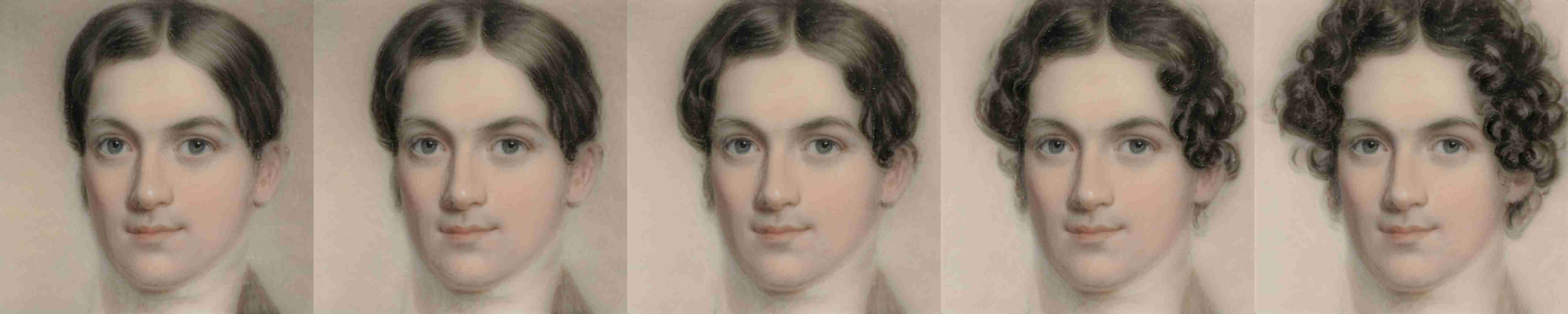}}
\centerline{\includegraphics[width=\columnwidth]{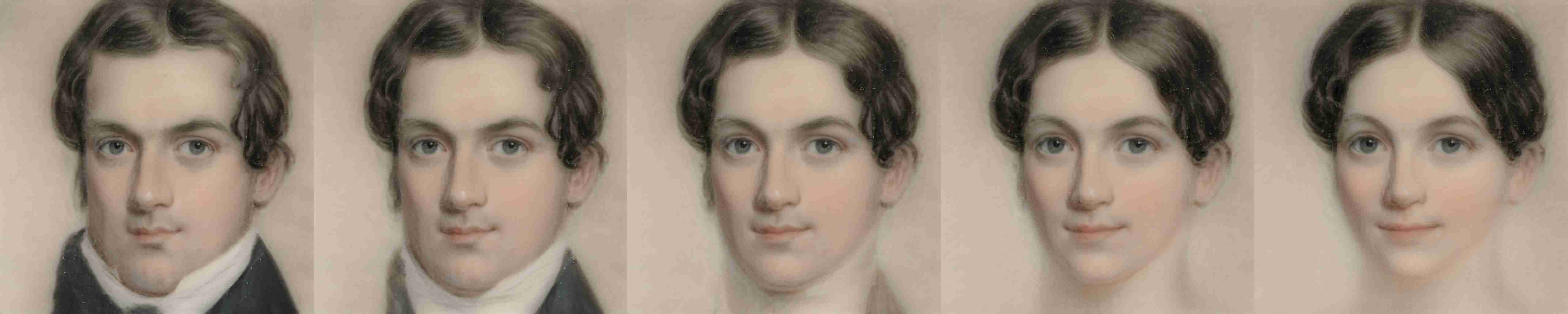}}
\centerline{\includegraphics[width=\columnwidth]{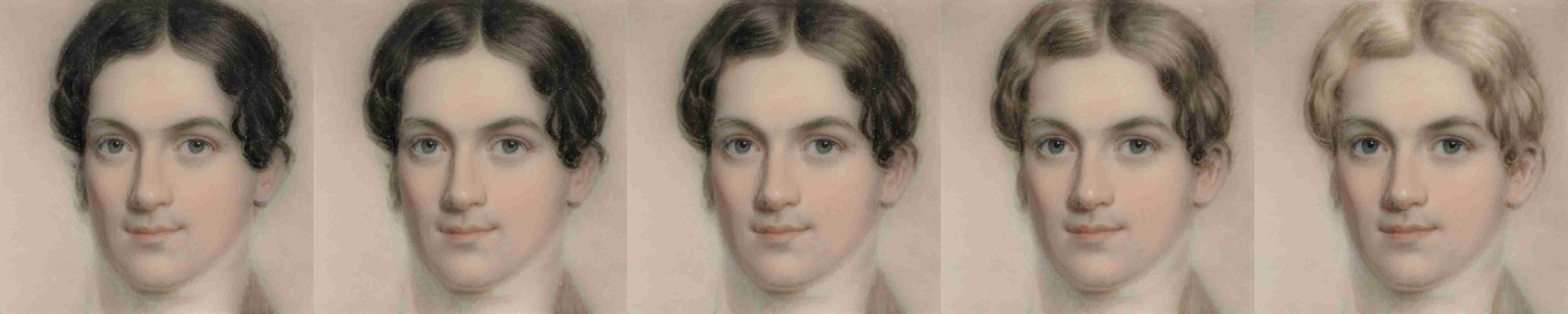}}
\centerline{\includegraphics[width=\columnwidth]{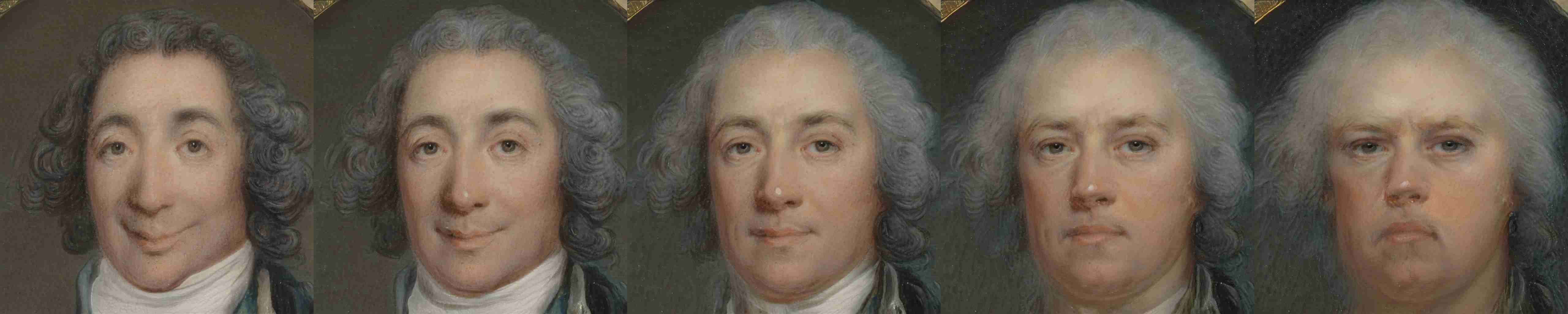}}
\centerline{\includegraphics[width=\columnwidth]{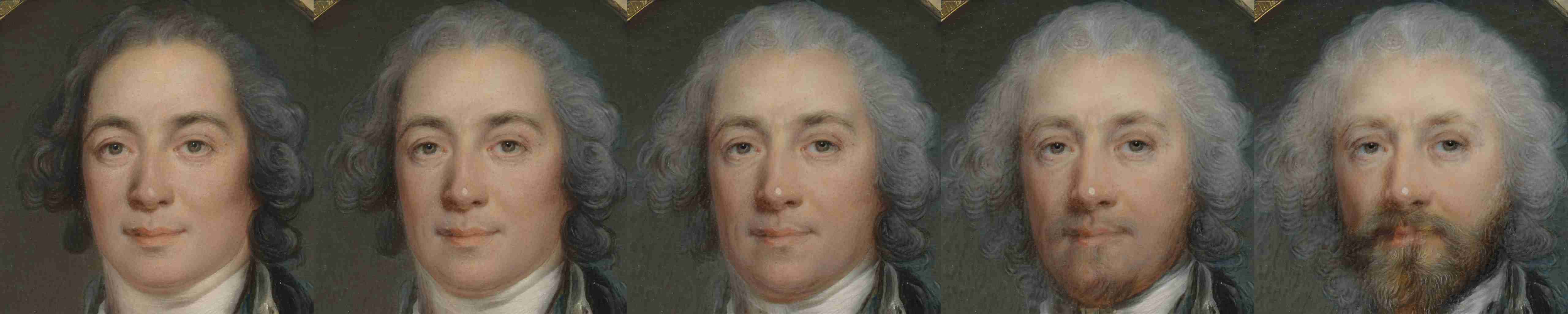}}
\centerline{\includegraphics[width=\columnwidth]{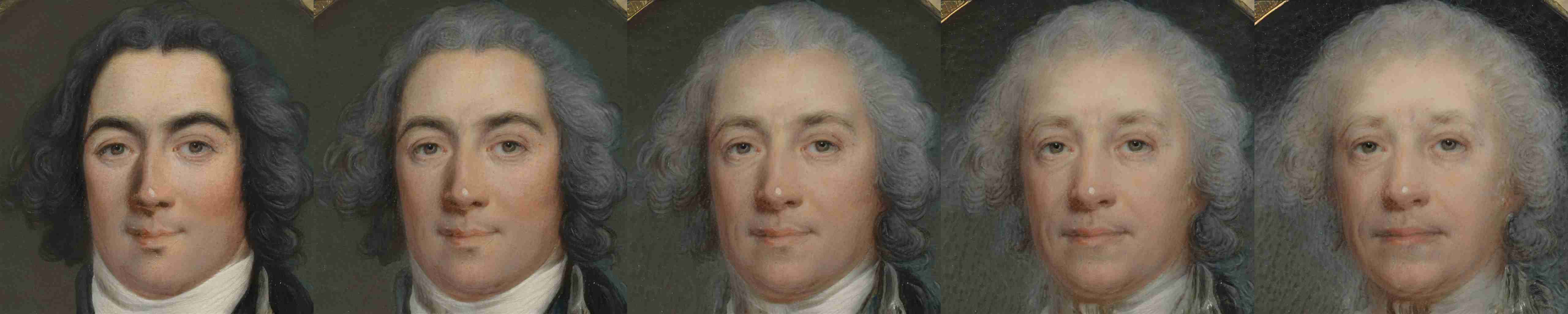}}
\caption{Manipulation results  using StyleGAN2 pre-trained on the MetFaces dataset \cite{Karras2020TrainingGA}. From top to bottom: “hitop fade”, “makeup”, “more hair”, “smile”, “curly hair”, “woman”, “blonde hair”, “angry”, “beard” and “older” manipulations.}
\label{fig:metfaces}
\end{center}
\vskip -0.2in
\end{figure}

\begin{figure}[!t]
\vskip 0.2in
\begin{center}
\centerline{\includegraphics[width=\columnwidth]{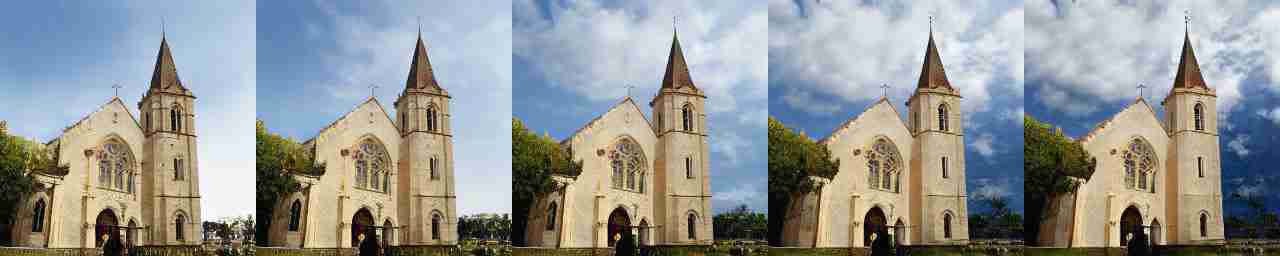}}
\centerline{\includegraphics[width=\columnwidth]{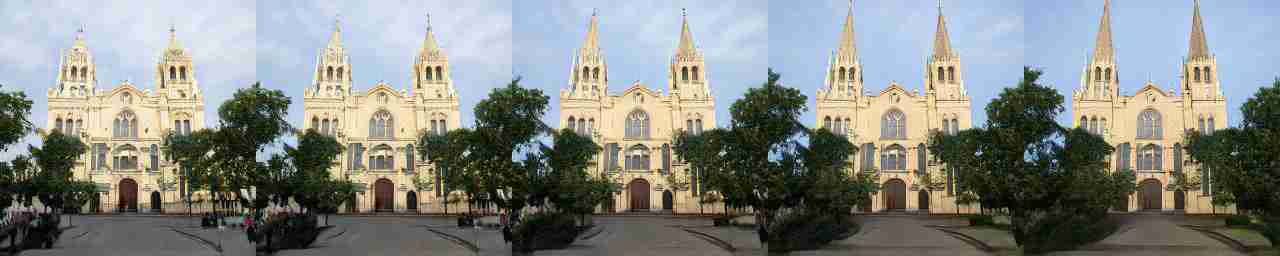}}
\centerline{\includegraphics[width=\columnwidth]{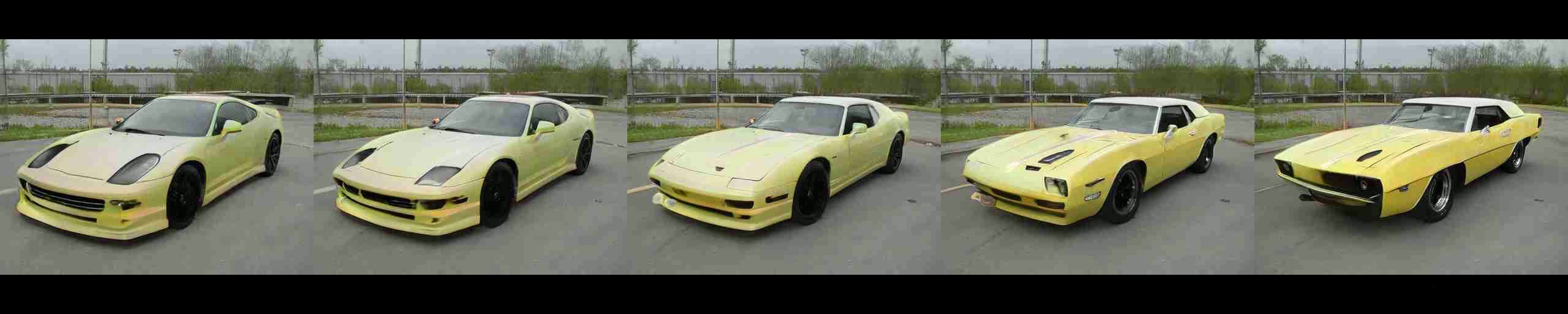}}
\centerline{\includegraphics[width=\columnwidth]{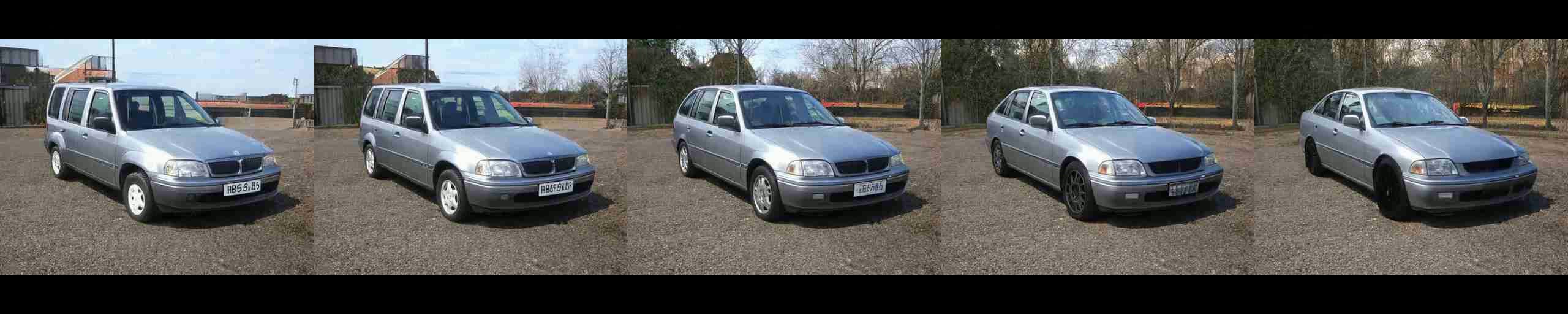}}
\centerline{\includegraphics[width=\columnwidth]{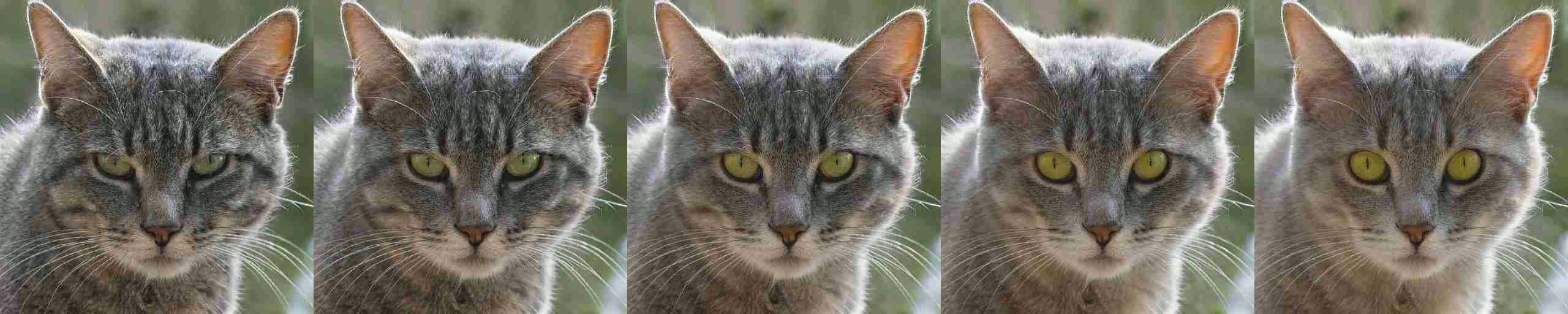}}
\centerline{\includegraphics[width=\columnwidth]{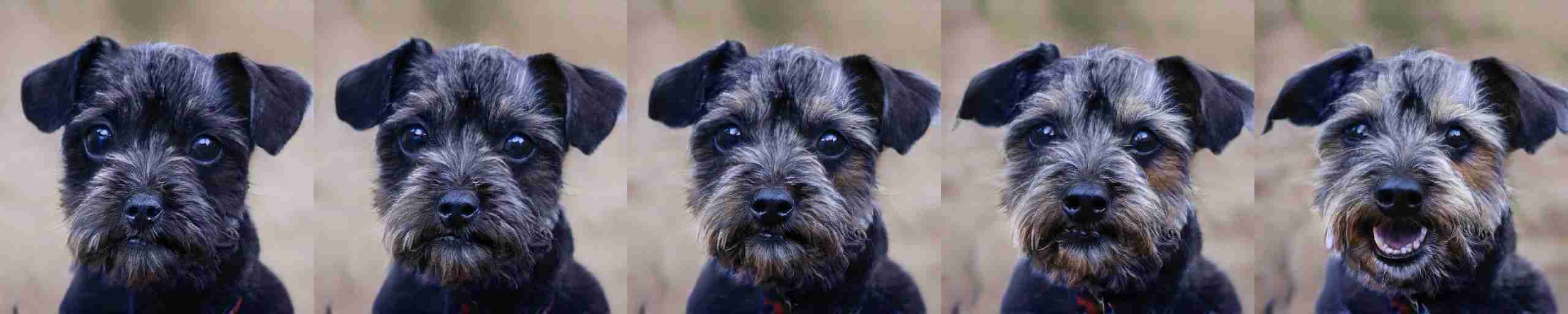}}
\centerline{\includegraphics[width=\columnwidth]{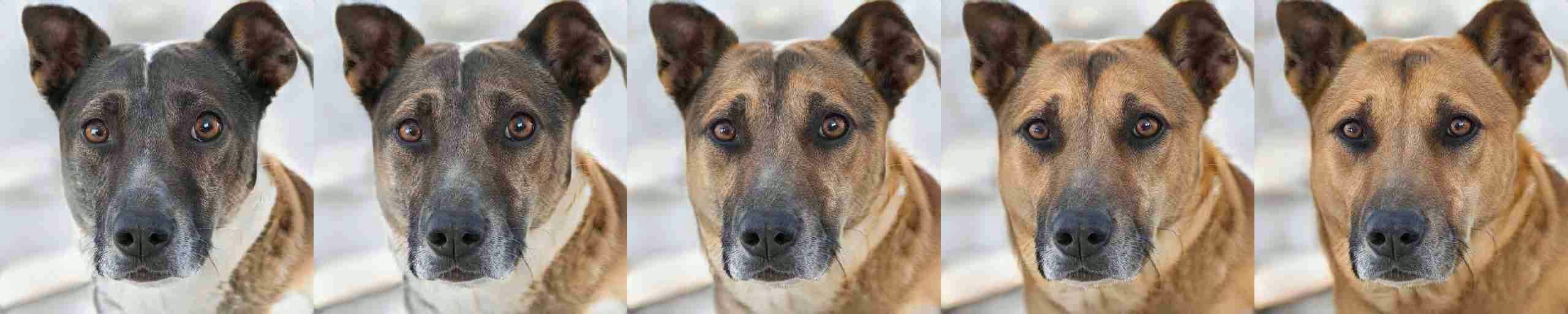}}
\caption{Manipulation results of our method on randomly generated images in various domains. Edits are performed using StyleGAN2 pre-trained on LSUN Car, Church and AFHQ Cat, Dog datasets. From top to bottom: “cloud”, “spires”, “classic car”, “sports car”, “cute”, “happy” and “golden fur” manipulations.}
\label{fig:others}
\end{center}
\vskip -0.2in
\end{figure}

Next, we perform manipulations on images from the MetFaces dataset. Figure \ref{fig:metfaces} shows that our method can perform a wide variety of simple and complex edits on images of paintings such as hair style, makeup, age and gender changes. In Figure \ref{fig:others}, we present manipulation results of random images using StyleGAN2 pre-trained on LSUN Car, Church \cite{yu2015lsun} and AFHQ Cat, Dog \cite{Choi2020StarGANVD} datasets. As can be seen in Figure \ref{fig:others}, our method can successfully handle complex edits such as adding clouds, changing the car model or altering the species and expressions of animals.

\end{appendices}
  
\end{document}